\documentclass{article}

\PassOptionsToPackage{numbers}{natbib}

\usepackage[final]{neurips_2022}

\usepackage{natbib}
\usepackage{geometry}
\usepackage{graphicx}
\usepackage[ruled,linesnumbered]{algorithm2e}
\usepackage{times}
\usepackage{soul}
\usepackage{url}
\usepackage[small]{caption}
\usepackage{graphicx}
\usepackage{amsmath}
\usepackage{amsthm}
\usepackage{booktabs}
\usepackage{amssymb}
\usepackage{mathtools}
\usepackage{bm}
\usepackage[american]{babel}
\usepackage{enumitem}
\usepackage{float,dblfloatfix}
\usepackage{multirow}
\usepackage{relsize}
\usepackage{subfigure}
\usepackage[utf8]{inputenc} 
\usepackage[T1]{fontenc}    
\usepackage{hyperref}       
\usepackage{url}            
\usepackage{booktabs}       
\usepackage{amsfonts}       
\usepackage{nicefrac}       
\usepackage{microtype}      
\usepackage{xcolor}         

\usepackage{amsmath}
\usepackage{amssymb}
\usepackage{mathtools}
\usepackage{amsthm}

\theoremstyle{plain}
\newtheorem{theorem}{Theorem}[section]

\theoremstyle{definition}
\newtheorem{definition}[theorem]{Definition}

\theoremstyle{remark}

\usepackage{array}
\usepackage{graphicx}
\usepackage{wrapfig}

\title{Multi-agent Dynamic Algorithm Configuration} 

\author{%
  Ke Xue$^1$\thanks{Equal Contribution}, \; Jiacheng Xu$^1$\footnotemark[1], \; Lei Yuan$^{1,2}$, \\ 
  \ \textbf{Miqing Li}$^3$, \ \textbf{Chao Qian}$^1$\thanks{Corresponding Author}, \ \textbf{Zongzhang Zhang}$^1$, \ \textbf{Yang Yu}$^{1,2}$ \\
  $^1$ State Key Laboratory for Novel Software Technology, Nanjing University\\
  $^2$ Polixir Technologies\\
  $^3$ CERCIA, School of Computer Science, University of Birmingham\\
  \texttt{\{xuek, xujc, yuanl\}@lamda.nju.edu.cn}, \\ \texttt{ m.li.8@bham.ac.uk}, \texttt{\{qianc, zzzhang, yuy\}@nju.edu.cn}\\
}

\begin{document}

\maketitle

\begin{abstract}
Automated algorithm configuration relieves users from tedious, trial-and-error tuning tasks. A popular algorithm configuration tuning paradigm is dynamic algorithm configuration (DAC), in which an agent learns dynamic configuration policies across instances by reinforcement learning (RL). However, in many complex algorithms, there may exist different types of configuration hyperparameters, and such heterogeneity may bring difficulties for classic DAC which uses a single-agent RL policy. In this paper, we aim to address this issue and propose multi-agent DAC (MA-DAC), with one agent working for one type of configuration hyperparameter. MA-DAC formulates the dynamic configuration of a complex algorithm with multiple types of hyperparameters as a contextual multi-agent Markov decision process and solves it by a cooperative multi-agent RL (MARL) algorithm. To instantiate, we apply MA-DAC to a well-known optimization algorithm for multi-objective optimization problems. Experimental results show the effectiveness of MA-DAC in not only achieving superior performance compared with other configuration tuning approaches based on heuristic rules, multi-armed bandits, and single-agent RL, but also being capable of generalizing to different problem classes. Furthermore, we release the environments in this paper as a benchmark for testing MARL algorithms, with the hope of facilitating the application of MARL.
\end{abstract}

\section{Introduction}
Finding right configurations of hyperparameters is critical for many learning and optimization algorithms~\citep{automl-book}. Automated methods, such as algorithm configuration (AC)~\citep{ac2009,irace}, emerge to search for right configurations, with the aim of relieving users from tedious, trial-and-error tuning tasks. However, static configuration policies obtained by AC may not necessarily yield optimal performance (compared with dynamic policies) since algorithms may require different configurations at different stages of their execution~\citep{dac-app-lhs}.

Dynamic algorithm configuration (DAC)~\citep{dac,dac-survey} is a well-known paradigm for obtaining dynamic configuration policies. Unlike AC, DAC can dynamically adjust the algorithm's configuration during the optimization process, through formulating it as a contextual Markov decision process (MDP) and then solving it by reinforcement learning (RL)~\citep{rlbook}. DAC has been found to outperform static methods on many tasks, including the learning rate tuning of deep neural networks~\citep{dac-app-dl}, step-size adaptation of evolution strategies~\citep{LTO-CMA}, and heuristic selection of AI planning~\citep{dac-app-lhs}.

The task of DAC typically focuses on a single type of configuration hyperparameter, such as tuning the step-size in CMA-ES~\citep{LTO-CMA}. However, due to the increasing complexity of real-world problem modeling, there are many algorithms whose performance rests on multiple types of hyperparameters. Tuning one type while fixing the rest may not produce promising results. Take a popular evolutionary algorithm for multi-objective optimization problems, MOEA/D~\citep{moead}, as an example. It has four types of configuration hyperparameters: weights, neighborhood size,  reproduction operator type, and parameters associated with the reproduction operator used. All of them are critical and significantly affect the performance of the algorithm~\citep{moead-survey}. Finding a near-optimal configuration combination for each part of these hyperparameters requires massive manual effort~\citep{GrecoRTN19}. However, how to jointly adjust multiple types of configuration hyperparameters is still an open problem~\cite{dac-survey}.

In this paper, we attempt to extend DAC to deal with tasks with multiple types of configuration hyperparameters. We model this as a cooperative multi-agent problem~\cite{reviewOfCoop}, with each agent handling one type of hyperparameter, for a shared goal (i.e., maximizing the team reward). Specifically, the proposed multi-agent DAC (MA-DAC) method formulates the task as a contextual multi-agent MDP (MMDP)~\citep{boutilier1996planning}, and solves it by a common pay-off cooperative multi-agent RL (MARL) algorithm. As an instantiation, we apply MA-DAC to the multi-objective evolutionary algorithm MOEA/D~\citep{moead} to learn the right configurations of its four types of hyperparameters by the classic MARL algorithm called value-decomposition networks (VDN)~\cite{sunehag2017value}.

Experiments on well-established multi-objective optimization problems show that MA-DAC outperforms other configuration tuning methods based on heuristic rules~\citep{AWA}, multi-armed bandits~\citep{FRRMAB, MAUCB} and single-agent RL~\citep{DQN}. Furthermore, we demonstrate the generalization ability of the learned MA-DAC policy to both inner classes (different instances with the same number of objectives) and outer classes (different instances with different numbers of objectives). Ablation studies also demonstrate the importance of tuning every type of hyperparameter. 

Our contributions are three-fold:
\begin{enumerate}
    \item To the best of our knowledge, the MA-DAC method is the first one to address the dynamic configuration of algorithms with multiple types of hyperparameters.
    \item The contextual MMDP formulation of MA-DAC is analyzed, and experimental results show that the presented formulation works well and has good generalization ability.
    \item The instantiation of configuring MOEA/D in this work can be used as a benchmark problem for MARL. The heterogeneity of MOEA/D's hyperparameters and the stochasticity of its search can promote the research of MARL algorithms. Besides, the learned policies are useful for a specific type of optimization task - multi-objective optimization, which will facilitate the application of MARL.
\end{enumerate} 


\section{Background}
\subsection{Dynamic algorithm configuration}

Different from the static configuration of AC, DAC aims at dynamically adjusting the configuration hyperparameters of an algorithm during its optimization process. \citet{dac} formulated DAC as a contextual MDP $ \mathcal{M}_{\mathcal I}:= {\mathcal M}_{i\sim\mathcal I}$ and applied RL to solve it. $\mathcal I$ represents the space of problem instances, and each $\mathcal M_i := \langle  \mathcal{S}, \mathcal{A}, \mathcal{T}_i,  r_i \rangle$~\citep{dac,dacbench} corresponds to one target problem instance $i \in \mathcal {I}$. The notion of context $\mathcal{I}$ allows to study generalization of policies in a principled manner~\citep{generalization}. Given a target algorithm $A$ with its configuration hyperparameters space $\Theta$, a DAC policy $\pi \in \Pi$ maps the state $s \in \mathcal S$ (e.g., history of changes in the objective value achieved by the target algorithm $A$) to the action $a \in \mathcal A$ (i.e., a hyperparameter $\theta \in \Theta$ of the target algorithm $A$). DAC aims at improving the performance of $A$ on a set of instances (e.g., optimization functions). Given a probability distribution $p$ over the space $\mathcal I$ of problem instances, the objective of DAC is to find an optimal policy $\pi^*$. That is,
\begin{equation}
    \pi^* \in \mathop{\arg\min}\limits_{\pi \in \Pi} \int_{i \in \mathcal I} p(i)c(\pi,i)\text{d}i,
\end{equation}
where $i\in \mathcal I$ is an instance to be optimized, and $c(\pi,i)\in\mathbb R$ is the cost function of the target algorithm with policy $\pi$ on the instance $i$. It has been shown that DAC outperforms static policies in learning rate adaptation in SGD~\citep{dac-app-dl}, step-size adaptation in CMA-ES~\citep{LTO-CMA}, and heuristic selection in planning~\citep{dac-app-lhs}. These applications all only involve a single type of hyperparameter. Dynamic configurations of complex algorithms with multiple types of hyperparameters have been found to be difficult for current DAC methods~\citep{dacbench,dac-survey}.

\subsection{Multi-agent reinforcement learning}\label{sec-marl}

A multi-agent system~\cite{yang2020overview} under fully observable cooperative situation can be modeled as an MMDP~\citep{boutilier1996planning}, which can be formalized as $\mathcal{M} := \langle \mathcal{N}, \mathcal{S}, \{\mathcal{A}_j\}_{j=1}^n, \mathcal{T},  r \rangle$, where  $\mathcal{N}$ is a set of $n$ agents, $\mathcal{S}$ is the state space, and $\mathcal{A}_j$ is agent $j$'s action space. At each time-step, agent $j \in \mathcal{N}$ acquires $s \in \mathcal{S}$ and then chooses the action $a^{(j)} \in \mathcal{A}_j$. The joint action $\boldsymbol{a}=\langle a^{(1)}, \dots, a^{(n)} \rangle$ leads to next state $s'\sim \mathcal{T}(\cdot \mid s, \boldsymbol{a})$ and all agents get a shared global reward $r(s, \boldsymbol{a})$.

The goal of an MMDP is to find a joint policy that maps the states to probability distributions over joint actions, $\boldsymbol{\pi} : S \rightarrow \Delta (\mathcal{A}_1\times \mathcal{A}_2\times \cdots \times \mathcal{A}_n)$, where $\Delta (\mathcal{A}_1\times \mathcal{A}_2\times \cdots \times \mathcal{A}_n)$ stands for the distribution over joint actions, with the goal of maximizing the global value function: 
\begin{equation}
 Q_{}^{\boldsymbol{\pi}}(s, \boldsymbol{a}) =\mathbb{E}_{\boldsymbol{\pi}}\left[\sum_{t=0}^\infty\gamma^tr(s_t, \boldsymbol{a}_t)\mid s_0=s, \boldsymbol{a_0}=\boldsymbol{a}\right].
\end{equation}

\section{Multi-agent DAC}
This section is devoted to the MA-DAC method, where Section~\ref{sec:madac formulation} is concerned with formulating MA-DAC as a contextual MMDP and Section~\ref{sec:madac mdp} explains components of MA-DAC.
 
\subsection{Problem formulation}\label{sec:madac formulation}
We propose MA-DAC as a new paradigm for solving the dynamic configuration of algorithms with multiple types of hyperparameters. We formulate MA-DAC as a contextual MMDP $\mathcal{M}_{\mathcal{I}}:=\{\mathcal{M}_i\}_{i \sim \mathcal{I}}$, where $\mathcal{M}_i := \langle \mathcal{N}, \mathcal{S}, \{\mathcal{A}_j\}_{j=1}^n, \mathcal{T}_i,  r_i \rangle$ is a single MMDP as defined in Section~\ref{sec-marl}. The notion of context $\mathcal{I}$ induces multiple MMDPs~\citep{dac,dacbench}. Each $\mathcal{M}_i$ stands for a specific instance $i$ sampled from a given distribution over $\mathcal{I}$, where an instance $i\in\mathcal{I}$ corresponds to a function $f$ to be optimized. Different $\mathcal{M}_i$s have shared state and action spaces, but with different transitions and reward functions. Algorithms are often tasked with solving different problem instances from the same or similar domains. Searching for well-performing parameter settings on a specific instance might achieve strong performance on that individual instance but might perform poorly in new instances. Therefore, we explicitly take the instance distribution $\mathcal{I}$ as context into account to facilitate generalization \citep{dac}. 

Given a parameterized algorithm $A$ with a configuration space $\Theta=\{\theta_j\}_{j\in \mathcal{N}}$, one agent in MA-DAC is to handle one type of configuration hyperparameter $\theta_j \in \Theta$ and all the agents work with the same goal, i.e., maximizing the team reward. The action space of each agent $j\in \mathcal{N}$ is the hyperparameter space of the corresponding $\theta_j$. Note that MA-DAC can be seen as a common-payoff cooperative MARL problem and can be solved by any cooperative MARL algorithm.

\subsection{Components of MA-DAC}\label{sec:madac mdp}

Next, we introduce several important components of MA-DAC, including the state, action, transition, reward, and instance set. 

\paragraph{State.} The state is used to describe the situation of the target algorithm, which is a key component in MA-DAC. We suggest that the state should have the following properties. 
\begin{enumerate}
    \item Accessibility. The state should be accessible in each step during the optimization process.
    \item Representation. The state should reflect the information in the optimization process.
    \item Generalization. The learned policy is expected to generalize to inner and even outer classes of instances. Thus, the state should consist of the common features across different instances.
\end{enumerate}
Besides, it is better if the state can be easily obtained, reducing the computational cost. Note that some of the above properties of state formulation have also been suggested for DAC~\citep{ACcmp, dacbench}.

\paragraph{Action.} Each agent in MA-DAC focuses on one type of configuration hyperparameter. Its action is the value of the hyperparameter that should be adjusted, which can be discrete or continuous.
MA-DAC agents are heterogeneous because they have completely different action spaces and affect different types of hyperparameters of the target algorithm.

\paragraph{Transition.} The transition function describes the dynamics of the problem. For an iterative algorithm, each iteration can be defined as a step. Given state $s_t$, each agent $j$ acts $a^{(j)}_t$, and the probability of reaching state $s_{t+1}$ can be expressed as $\mathcal{T}(s_{t+1}\mid s_t,\boldsymbol{a}_t)$. Different from the state and action spaces, the transition function depends on the given instance $f\in \mathcal{F}$. Thus, the policy should identify the characteristics of the current instance and take the optimal action under it. 

\paragraph{Reward.} The reward is used to guide the policy's learning process, whose quality has a significant impact on the policy's final performance. Many RL benchmark environments include a well-defined reward function. However, in many application scenarios, we must define the reward function manually based on a specific metric that can indicate the final performance.

Since the goal of an iterative optimization algorithm at iteration $t$ is to find a solution with the best objective function value so far rather than to just find a better solution than the solution at the $t-1$ step (since the quality of solutions in the last step may be poor), it would make sense to give the policy a reward when finding a better solution than the best solution so far (rather than finding a better solution than the solution at the last step). Another factor one may need to consider when designing a reward function is that with time, it is getting harder to find better-quality solutions. In the beginning, it may be easy to achieve rapid improvement of solution quality, but in later stages, improvement could be harder. As such, a reward function that rewards more an agent who can find a better solution in later stages can encourage the agent to find a very good solution in the end. Considering these two factors, we propose the following reward function at step $t$:
\begin{equation}\label{eq:reward}
r_t= \begin{cases}
(1/2)\cdot (p_{t+1}^2-p_{t}^2)\ & \text{if}\ f(s_{t+1})< f^*_t \\
0 & \text{otherwise} \\
\end{cases}, 
\end{equation}

where 
\begin{equation}
    \  p_{t+1}= \begin{cases} \frac{f(s_0)-f(s_{t+1})}{f(s_0)} & \text{if} \ f(s_{t+1})< f^*_t\\ 
p_{t} &  \text{otherwise} \\
\end{cases},
\end{equation}
and $f^*_t$ is the minimum metric value found until step $t$. 

\begin{wrapfigure}[14]{r}{0.41\textwidth}
\centering
\includegraphics[width=0.36\textwidth]{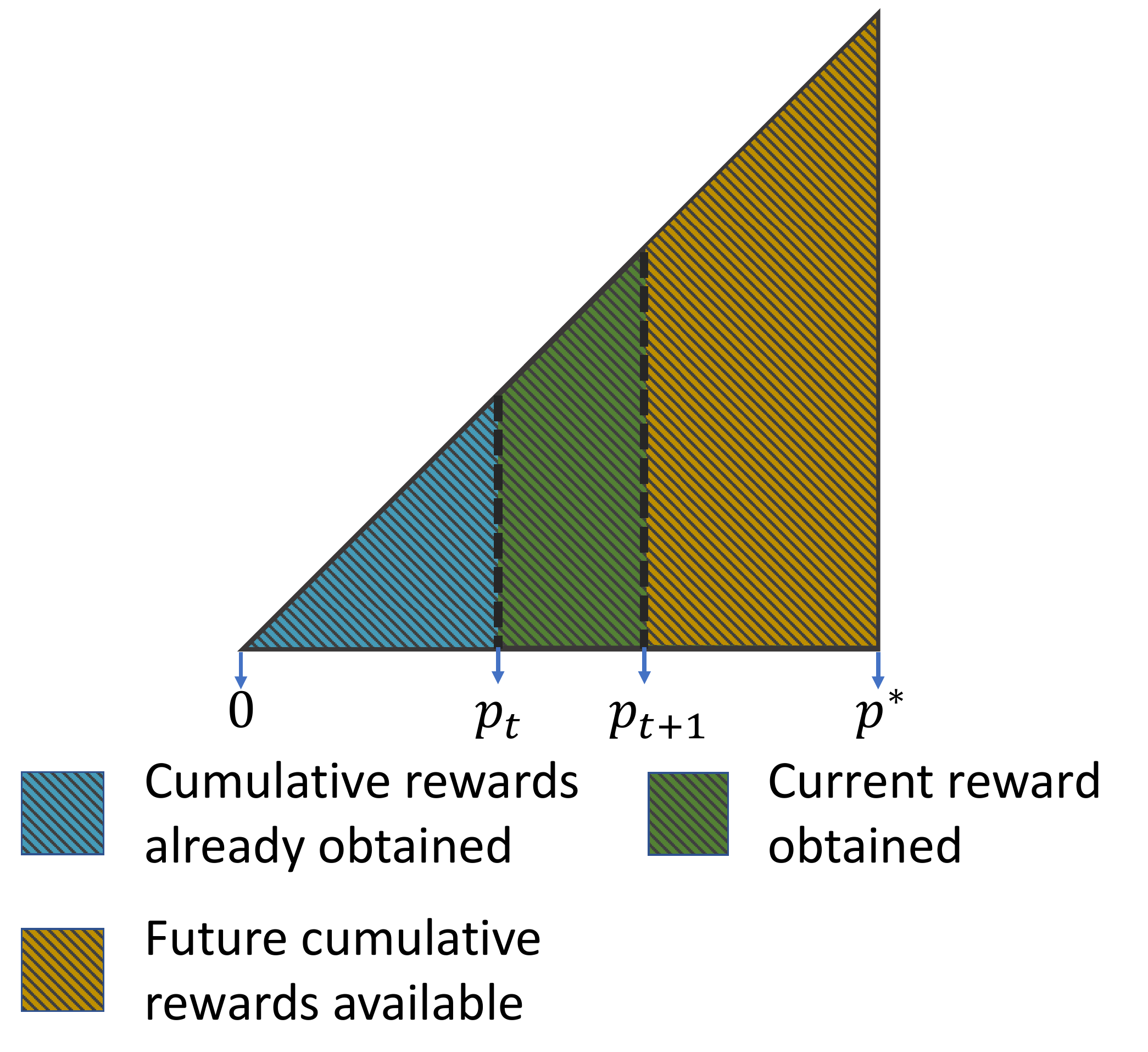} \caption{Illustration of the reward function.}
\label{fig:wrapfig}
\end{wrapfigure}

As shown in Figure~\ref{fig:wrapfig}, the isosceles right-angled triangle indicates the maximum return we can get, where $p^*$ is the largest value of $p_t$. At step $t$, the current reward we obtain is the area of the triangle with side length $p_{t+1}$ minus the area of the triangle with side length $p_{t}$. The proposed triangle-based reward function has two properties: 1) it only rewards the agent when a new solution is better than the best solution found so far and 2) the reward increases in later stages. The comparative experiment is provided in Appendix~\ref{app: analysis of reward}. This design may be applicable to a wide range of similar tasks.

\paragraph{Instance set.} Instance set defines the optimization problem instances that have to be solved by the target algorithm. Note that it is possible to learn policies even across very heterogeneous problem instances by using the context information~\citep{dac,LTO-CMA}. Admittedly, if the instances have some similar properties, the learned policy can generalize better~\citep{dacbench}.

\section{Applying MA-DAC to dynamic configuration tuning of MOEA/D}
As an instantiation, we apply MA-DAC to a well-established evolutionary algorithm for multi-objective optimization problems, multi-objective evolutionary algorithm based on decomposition (MOEA/D)~\citep{moead}. As one of the most widely used multi-objective evolutionary algorithms, MOEA/D has several heterogeneous types of configuration hyperparameters (e.g., weights, neighborhood size, and reproduction operator type), which is a good fit for the proposed MA-DAC. We first briefly introduce multi-objective optimization and MOEA/D in Section~\ref{sec:mop and moead}, followed by detailed explanations of applying MARL to MOEA/D in Section~\ref{sec:madac for moead}, which results in an MARL benchmark, denoted by MaMo. Finally, we compare MaMo with other MARL benchmarks in Section~\ref{sec: comparison of benchmarks}.  

\subsection{Multi-objective optimization and MOEA/D}\label{sec:mop and moead}

Multi-objective optimization refers to an optimization problem with more than one objective to be considered simultaneously. It is very common in the real world. For example, in deep neural networks, apart from the accuracy, one may also care about the model size, latency, and/or energy consumption~\citep{nsganet,hongwj}. 

A prominent feature of multi-objective optimization problems (MOPs) is that unlike single-objective optimization problems, in MOPs there is no single optimal solution, but a set of Pareto optimal solutions. A solution is called Pareto optimal if there is no other solution in the search space that dominates it\footnote[1]{For two solutions $x$ and $y$, $x$ is called to dominate $y$, if $x$ is not worse than $y$ on all objectives and better than $y$ on at least one objective.}. Unfortunately, the number of Pareto optimal solutions for a given MOP is typically prohibitively large or even infinite. Therefore, the goal of a multi-objective optimization algorithm is to find a good approximation that well represents all the Pareto optimal solutions.

Evolutionary algorithms have demonstrated their effectiveness to solve MOPs. Their population-based nature can approximate the problem's optimal solutions within one execution, with each solution in the population representing a unique trade-off among the objectives. MOEA/D~\citep{moead} is a representative multi-objective evolutionary algorithm. Unlike other mainstream multi-objective evolutionary algorithms like NSGA-II~\citep{nsgaii}, which compare solutions based on their Pareto dominance relation and density in the population, MOEA/D converts an MOP into a number of single-objective sub-problems through a number of weights, where neighboring solutions work cooperatively for the optimal solutions for the single-objective sub-problems. As such, MOEA/D entails different heterogeneous types of hyperparameters, e.g., weights and neighborhood size. Such hyperparameters significantly affect the performance of the algorithm. For example, different weight distributions are suitable for MOPs with different Pareto front shapes as the weights are used to control the distribution of the final population~\citep{what-weights}. A large neighborhood has high search ability in the objective space while a small neighborhood is beneficial for diversity maintenance in the decision space~\citep{neighborhood-size}. In short, finding right configurations for these hyperparameters requires massive manual effort and different configurations work well on different MOPs.

Note that some recent efforts have been made to adjust the hyperparameters of MOEA/D by heuristic rules~\citep{AWA}, multi-armed bandit~\citep{FRRMAB}, and RL~\citep{moead-dqn}. However, they all focus on a single type of hyperparameter, and also very few of them work on the dynamic algorithm configuration well. 

\subsection{MA-DAC for MOEA/D}\label{sec:madac for moead}
In this section, we show how to apply MA-DAC to learn the configuration policy of MOEA/D and also introduce the resulting environment MaMo. Each generation in the evolutionary process of MOEA/D is one step in the MaMo.

The state of MaMo can be divided into three parts. 
\begin{enumerate}
    \item  To describe the general properties of the optimization problems, the first part is the features of the problem instance, including the number of variables and objectives. 
    \item To emphasize the general information of the algorithm, the second part includes the features of the optimization process, i.e., how much computational budget has been used and how many steps of the algorithm have not made any progress. 
    \item To show the properties of the population and that how the population is evolved, we use several indicators, i.e., hypervolume\footnote[2]{Hypervolume is a quality indicator that can reflect both convergence and diversity of a solution set~\citep{indicators-survey}.}~\citep{hv}, ratio of non-dominated solutions in the population and average distance of the solutions, in the third part. For each indicator, we also use the gap between the current value and the value corresponding to the last population to reflect the immediate evolutionary progress. Besides, we use statistic metrics (the mean and standard deviation) of the indicators in several recent steps and all steps from the beginning to measure the short and long histories of the optimization, respectively. 
\end{enumerate}

To facilitate the generalization of the learned MA-DAC policy, we have pre-processed some state features. More details about state are provided in Appendix~\ref{app a: state} due to space limitation.

We consider four heterogeneous types of configuration hyperparameters in MOEA/D as the actions of different agents of MA-DAC.
\begin{enumerate}
    \item Weights. In MOEA/D, weights are used to maintain the diversity of the converted single-objective sub-problems~\citep{AWA}. The action space is discrete with two dimensions, i.e., adjusting (T) and not adjusting (N) the weights. Furthermore, we limit the frequency of adjustment, because too frequent adjustment will lead to drastic changes in the sub-problems and is detrimental to the optimization process.
    \item Neighborhood size. The neighborhood size is to control the distance between solutions in mating selection; a small size helps the search exploit the local area while a large size helps the search explore a wide objective space~\citep{AGR}. We discretize the action space into four dimensions. That is, the neighborhood size can be  $15$, $20$, $25$ or $30$.
    \item Types of the reproduction operators. We consider four types of differential evolution (DE) operators. Each type has different search ability~\citep{FRRMAB, DEDDQN}.
    \item Parameters of reproduction operators. The parameters (e.g., scaling factor) of the reproduction operators in MOEA/D significantly affect the algorithm's performance~\citep{LED}, and the action space has four discrete dimensions. 
\end{enumerate}
The detailed definitions of the actions are given in Appendix~\ref{app:action} due to space limitation.

We use the MOPs with similar properties from the well-known MOP benchmarks DTLZ~\citep{DTLZ} and WFG~\citep{WFG} as the instance set of MaMo. These MOPs can have different number of objectives, and those with the same number of objectives can be seen as inner classes because the number of objectives considerably affects the properties and difficulty level of MOPs. We use the inverted generational distance (IGD)~\citep{igd} as the metric in Eq.~(\ref{eq:reward}), resulting in the reward function.

To learn the policy, we use a classical MARL algorithm named VDN~\citep{sunehag2017value}, which is widely used in cooperative multi-agent systems. VDN learns to decompose the team value function into agent-wise value functions, alleviating the exponential growth of the action space. It follows the Individual-Global-Max (IGM) principle~\citep{son2019qtran}, i.e., the consistency between joint and local greedy action selections by the joint value function $Q_{\rm tot}(s, \boldsymbol{a})$ and individual value functions $\left[Q_j(s, a^{(j)})\right]_{j=1}^n$. All parameters in MA-DAC are updated using the standard TD loss from the global Q-value $Q_{\rm tot}$, which follows additivity to factorize the global value function, i.e., $Q_{\text{tot}}(s, \boldsymbol{a})=\sum_{j=1}^{n} Q_{j}\left(s, a^{(j)}\right)$. In the testing phase, each agent acts greedily with respect to its $Q_j$. 

\subsection{Comparison among MARL benchmarks}\label{sec: comparison of benchmarks}

There are many benchmarks~\cite{epymarl, terry2021pettingzoo} emerged in recent years. In order to clearly show the characteristics of MaMo, we compare it with some MARL benchmarks, as shown in Table~\ref{table:benchmark}. Heterogeneity refers to agents having different action spaces or skills~\cite{wakilpoor2020heterogeneous}. For example, in MaMo, there are four entirely different action spaces, i.e., weights, neighborhood size, types of the reproduction operators and their parameters. Stochasticity means that performing the same action in the same state may lead to a different next state. The high stochasticity of MaMo comes from the randomness of MOEA/D.

\begin{table}[h!]
    \centering
    \caption{Overview of MARL benchmarks and their properties.}
    \begin{tabular}{ccccc}
    \toprule
    Benchmark & Heterogeneous & \#agents & Stochastic & Application scenarios\\
    \midrule
    Matrix Games~\citep{claus1998dynamics}& $\times$  & 2 & Low & Game\\
    MPE~\citep{lowe2017multi} & $\times$ & 2-3 & Low & Game\\
    MAgent~\citep{zheng2018magent} & $\times$  & 2-1000 & Low &  Game\\
    SMAC~\citep{samvelyan2019starcraft} & $\checkmark$ & 2-30 & Low & Game\\
    Active Voltage Control~\citep{wang2021multi}& $\times$  &3-38 &Low & Control\\
    MaMo (Ours) & $\checkmark$ & 2-4 & High & Optimization\\
    \bottomrule
    \end{tabular}
    \label{table:benchmark}
\end{table}

The primary benchmarks are designed to investigate different aspects of multi-agent systems. For example, Matrix Games~\citep{claus1998dynamics} and Multi-Agent Particle Environments (MPE)~\citep{lowe2017multi} are popular and classical testbeds, to investigate cooperative or competitive behaviors in small scale settings (with 2--3 agents). MAgent~\citep{zheng2018magent} is used to test the scalability of MARL methods, where the number of agents is up to 1000. SMAC~\citep{samvelyan2019starcraft} has attracted wide attention in recent years to test the coordination ability in cooperative MARL~\cite{gorsane2022towards,ijcai2022p85,epymarl}. For the first time, Active Voltage Control~\citep{wang2021multi} creates an exciting yet challenging real-world scenario for the application of MARL. To the best of our knowledge, none of the known benchmarks has focused on highly heterogeneous and stochastic scenarios in MARL. We hope our new MaMo benchmark can offer a good supplement that could benefit the MARL community.

\section{Experiments}

To examine the effectiveness of MA-DAC, we conduct experiments using MaMo. We investigate the following research questions (RQs). RQ1: How does MA-DAC perform compared with the baseline and other tuning algorithms? RQ2: How is the generalization ability of MA-DAC? RQ3: How do the different parts of MA-DAC affect the performance? We introduce the experimental settings in Section~\ref{sec:exp settings}, and investigate the above RQs in Section~\ref{sec:exp results}.

\subsection{Experimental settings}\label{sec:exp settings}
We consider several well-established multi-objective test functions (DTLZ2, DTLZ4 and WFG4-WFG9) with 3, 5 and 7 objectives following the practice in \citet{large-scale} to examine our method. In all of the following experiments, several arbitrary functions (here DTLZ2, WFG4 and WFG6) from the function set are used as the training set, and all the other functions are considered as the testing set. Note that ``train'' and ``test'' in the tables mean that the comparison is based on the training (i.e., DTLZ2, WFG4 and WFG6) and testing (i.e., DTLZ4, WFG5, WFG7, WFG8 and WFG9) problems, respectively. IGD~\citep{igd} is considered as the metric to measure the performance of the algorithms. The mean and standard deviation of the IGD values obtained by each algorithm on each MOP for $30$ independent runs are reported. We apply the Wilcoxon rank-sum test with significance level 0.05. For a fair comparison, other parameters (such as population size and computational budget) are set to be the same across all the compared algorithms.  More details about the settings can be found in Appendix~\ref{app: details exp settings} due to space limitation. Our code is available at \url{https://github.com/lamda-bbo/madac}.

\subsection{Experimental results}\label{sec:exp results}

\begin{table}[t!]\scriptsize
\centering
\caption{IGD values obtained by MOEA/D, DQN, MA-UCB and MA-DAC on different problems. Each result consists of the mean and standard deviation of 30 runs. The best mean value on each problem is highlighted in \textbf{bold}. The symbols `$+$', `$-$' and `$\approx$' indicate that the result is significantly superior to, inferior to, and almost equivalent to MA-DAC, respectively, according to the Wilcoxon rank-sum test with significance level 0.05.}\vspace{0.5em}
\resizebox{\linewidth}{!}{
\begin{tabular}{c c|c c c c c c}
\toprule
     Problem     &   $M$   & MOEA/D  & DQN & MA-UCB & MA-DAC \\\midrule
\multirow{3}{*}{DTLZ2} & 3 & 4.605E-02 (3.54E-04) $-$  & 4.628E-02 (2.96E-04) $-$  & 4.671E-02 (3.70E-04) $-$  & \textbf{3.807E-02 }(5.05E-04)  \\
 & 5 & 3.006E-01 (1.55E-03) $-$  & 3.016E-01 (1.34E-03) $-$  & 3.041E-01 (1.69E-03) $-$  & \textbf{2.442E-01 }(1.26E-02)  \\
 & 7 & 4.455E-01 (1.41E-02) $-$  & 4.671E-01 (1.15E-02) $-$  & 4.826E-01 (9.59E-03) $-$  & \textbf{3.944E-01 }(1.17E-02)  \\\midrule

\multirow{3}{*}{WFG4} & 3 & 5.761E-02 (5.41E-04) $-$  & 6.920E-02 (1.20E-03) $-$  & 7.165E-02 (1.83E-03) $-$  & \textbf{5.200E-02 }(1.19E-03)  \\
 & 5 & 3.442E-01 (1.21E-02) $-$  & 2.810E-01 (6.86E-03) $-$  & 2.859E-01 (6.77E-03) $-$  & \textbf{1.868E-01 }(2.81E-03)  \\
 & 7 & 4.529E-01 (1.79E-02) $-$  & 3.725E-01 (1.14E-02) $-$  & 3.868E-01 (1.54E-02) $-$  & \textbf{3.033E-01 }(3.66E-03)  \\\midrule

\multirow{3}{*}{WFG6} & 3 & 6.938E-02 (5.50E-03) $-$  & 6.834E-02 (1.78E-02) $-$  & 6.601E-02 (1.00E-02) $-$  & \textbf{4.831E-02 }(8.95E-03)  \\
 & 5 & 3.518E-01 (2.82E-03) $-$  & 3.160E-01 (2.40E-02) $-$  & 3.359E-01 (1.47E-02) $-$  & \textbf{1.942E-01 }(6.90E-03)  \\
 & 7 & 4.869E-01 (3.03E-02) $-$  & 4.322E-01 (2.95E-02) $-$  & 4.389E-01 (3.41E-02) $-$  & \textbf{3.112E-01 }(4.93E-03)  \\\midrule
 
\multicolumn{2}{c|}{Train: $+$/$-$/$\approx$}  & 0/9/0  & 0/9/0  & 0/9/0  & \\\midrule

\multirow{3}{*}{DTLZ4} & 3 & 6.231E-02 (8.85E-02) $\approx$  & \textbf{5.590E-02 }(5.77E-03) $-$  & 6.011E-02 (5.08E-03) $-$  & 6.700E-02 (6.14E-02)  \\
 & 5 & 3.133E-01 (4.45E-02) $\approx$  & 3.457E-01 (1.61E-02) $-$  & 3.492E-01 (1.69E-02) $-$  & \textbf{2.995E-01 }(2.10E-02)  \\
 & 7 & 4.374E-01 (2.57E-02) $-$  & 4.552E-01 (1.47E-02) $-$  & 4.756E-01 (2.01E-02) $-$  & \textbf{4.182E-01 }(1.21E-02)  \\\midrule
 
 \multirow{3}{*}{WFG5} & 3 & 6.327E-02 (1.10E-03) $-$  & 6.212E-02 (5.54E-04) $-$  & 6.118E-02 (7.03E-04) $-$  & \textbf{4.730E-02 }(7.89E-04)  \\
 & 5 & 3.350E-01 (9.77E-03) $-$  & 3.077E-01 (6.36E-03) $-$  & 3.036E-01 (8.83E-03) $-$  & \textbf{1.811E-01 }(3.02E-03)  \\
 & 7 & 4.101E-01 (2.08E-02) $-$  & 4.996E-01 (1.32E-02) $-$  & 5.024E-01 (1.38E-02) $-$  & \textbf{3.206E-01 }(8.04E-03)  \\\midrule
 
\multirow{3}{*}{WFG7} & 3 & 5.811E-02 (6.31E-04) $-$  & 5.930E-02 (7.32E-04) $-$  & 6.014E-02 (7.11E-04) $-$  & \textbf{4.066E-02 }(5.31E-04)  \\
 & 5 & 3.572E-01 (5.47E-03) $-$  & 2.993E-01 (1.43E-02) $-$  & 3.207E-01 (1.71E-02) $-$  & \textbf{1.858E-01 }(2.12E-03)  \\
 & 7 & 5.236E-01 (2.19E-02) $-$  & 4.576E-01 (2.38E-02) $-$  & 4.879E-01 (2.75E-02) $-$  & \textbf{3.258E-01 }(1.25E-02)  \\\midrule
\multirow{3}{*}{WFG8} & 3 & 8.646E-02 (3.44E-03) $-$  & 9.280E-02 (1.06E-03) $-$  & 9.612E-02 (1.48E-03) $-$  & \textbf{7.901E-02 }(1.19E-03)  \\
 & 5 & 4.258E-01 (8.42E-03) $-$  & 3.969E-01 (1.26E-02) $-$  & 3.956E-01 (1.32E-02) $-$  & \textbf{2.479E-01 }(7.20E-03)  \\
 & 7 & 5.816E-01 (1.30E-02) $-$  & 5.575E-01 (1.39E-02) $-$  & 5.642E-01 (1.38E-02) $-$  & \textbf{4.127E-01 }(5.93E-03)  \\\midrule
\multirow{3}{*}{WFG9} & 3 & 5.817E-02 (1.24E-03) $-$  & 5.628E-02 (7.29E-04) $-$  & 7.953E-02 (2.45E-02) $-$  & \textbf{4.159E-02 }(6.10E-04)  \\
 & 5 & 3.633E-01 (1.20E-02) $-$  & 3.258E-01 (1.61E-02) $-$  & 3.396E-01 (1.55E-02) $-$  & \textbf{1.832E-01 }(7.10E-03)  \\
 & 7 & 5.538E-01 (2.63E-02) $-$  & 5.115E-01 (2.15E-02) $-$  & 5.227E-01 (1.79E-02) $-$  & \textbf{3.278E-01 }(7.21E-03)  \\\midrule

\multicolumn{2}{c|}{Test: $+$/$-$/$\approx$}  & 0/13/2  & 0/15/0  & 0/15/0  & \\\bottomrule
\end{tabular}}
\label{table:1 baseline}
\end{table}

\begin{table}[t!]\scriptsize
\centering
\caption{IGD values obtained by MA-DAC (M), MA-DAC (3), MA-DAC (5) and MA-DAC (7) on different problems. Each result consists of the mean and standard deviation of 30 runs. The best mean value on each problem is highlighted in \textbf{bold}. The symbols `$+$', `$-$' and `$\approx$' indicate that the result is significantly superior to, inferior to, and almost equivalent to the best value except for MA-DAC in Table~\ref{table:1 baseline}, respectively, according to the Wilcoxon rank-sum test with significance level 0.05.}\vspace{0.5em}
\resizebox{\linewidth}{!}{
\begin{tabular}{c c|c c c c c c }
\toprule
     Problem     &   $M$  & MA-DAC (M) &  MA-DAC (3) &  MA-DAC (5)  & MA-DAC (7)    \\\midrule
\multirow{3}{*}{DTLZ2} & 3 & 3.839E-02 (5.35E-04) $+$  & \textbf{3.807E-02 }(5.05E-04) $+$  & 3.830E-02 (7.24E-04) $+$  & 3.837E-02 (5.69E-04) $+$ \\
 & 5 & 2.468E-01 (7.55E-03) $+$  & 2.472E-01 (1.56E-02) $+$  & \textbf{2.442E-01 }(1.26E-02) $+$  & 2.569E-01 (1.39E-02) $+$ \\
 & 7 & 3.921E-01 (8.84E-03) $+$  & \textbf{3.880E-01 }(1.02E-02) $+$  & 4.081E-01 (1.52E-02) $+$  & 3.944E-01 (1.17E-02) $+$ \\\midrule
\multirow{3}{*}{WFG4} & 3 & 5.220E-02 (9.83E-04) $+$  & \textbf{5.200E-02 }(1.19E-03) $+$  & 5.236E-02 (1.10E-03) $+$  & 5.302E-02 (9.78E-04) $+$ \\
 & 5 & \textbf{1.850E-01 }(3.14E-03) $+$  & 1.867E-01 (3.01E-03) $+$  & 1.868E-01 (2.81E-03) $+$  & 1.853E-01 (2.67E-03) $+$ \\
 & 7 & 3.091E-01 (5.80E-03) $+$  & 3.104E-01 (7.14E-03) $+$  & 3.100E-01 (5.89E-03) $+$  & \textbf{3.033E-01 }(3.66E-03) $+$ \\\midrule
 
\multirow{3}{*}{WFG6} & 3 & 5.078E-02 (1.20E-02) $+$  & 4.831E-02 (8.95E-03) $+$  & \textbf{4.599E-02 }(9.48E-03) $+$  & 5.206E-02 (1.64E-02) $+$ \\
 & 5 & 1.971E-01 (6.40E-03) $+$  & 2.003E-01 (6.26E-03) $+$  & \textbf{1.942E-01 }(6.90E-03) $+$  & 1.957E-01 (6.67E-03) $+$ \\
 & 7 & 3.114E-01 (5.08E-03) $+$  & 3.242E-01 (9.24E-03) $+$  & 3.129E-01 (5.71E-03) $+$  & \textbf{3.112E-01 }(4.93E-03) $+$ \\\midrule
 
 \multirow{3}{*}{DTLZ4} & 3 & \textbf{6.171E-02 }(3.67E-02) $+$  & 6.700E-02 (6.14E-02) $+$  & 6.618E-02 (4.62E-02) $+$  & 8.088E-02 (7.12E-02) $+$ \\
 & 5 & 3.044E-01 (1.66E-02) $\approx$  & \textbf{2.974E-01 }(1.94E-02) $+$  & 2.995E-01 (2.10E-02) $\approx$  & 3.036E-01 (1.69E-02) $\approx$ \\
 & 7 & 4.271E-01 (1.45E-02) $+$  & 4.313E-01 (1.39E-02) $\approx$  & 4.327E-01 (2.15E-02) $\approx$  & \textbf{4.182E-01 }(1.21E-02) $+$ \\\midrule
 
 \multirow{3}{*}{WFG5} & 3 & \textbf{4.721E-02 }(7.15E-04) $+$  & 4.730E-02 (7.89E-04) $+$  & 4.733E-02 (8.10E-04) $+$  & 4.746E-02 (5.90E-04) $+$ \\
 & 5 & 1.811E-01 (2.59E-03) $+$  & 1.817E-01 (2.96E-03) $+$  & 1.811E-01 (3.02E-03) $+$  & \textbf{1.808E-01 }(2.83E-03) $+$ \\
 & 7 & 3.256E-01 (5.49E-03) $+$  & 3.266E-01 (8.98E-03) $+$  & 3.263E-01 (9.73E-03) $+$  & \textbf{3.206E-01 }(8.04E-03) $+$ \\\midrule
 
\multirow{3}{*}{WFG7} & 3 & 4.076E-02 (5.33E-04) $+$  & \textbf{4.066E-02 }(5.31E-04) $+$  & 4.077E-02 (5.12E-04) $+$  & 4.124E-02 (4.98E-04) $+$ \\
 & 5 & 1.839E-01 (2.38E-03) $+$  & 1.881E-01 (3.70E-03) $+$  & 1.858E-01 (2.12E-03) $+$  & \textbf{1.836E-01 }(2.21E-03) $+$ \\
 & 7 & 3.368E-01 (1.54E-02) $+$  & 3.461E-01 (1.97E-02) $+$  & 3.390E-01 (1.38E-02) $+$  & \textbf{3.258E-01 }(1.25E-02) $+$ \\\midrule
\multirow{3}{*}{WFG8} & 3 & \textbf{7.828E-02 }(1.46E-03) $+$  & 7.901E-02 (1.19E-03) $+$  & 7.921E-02 (1.36E-03) $+$  & 7.944E-02 (1.30E-03) $+$ \\
 & 5 & 2.506E-01 (1.11E-02) $+$  & 2.653E-01 (1.51E-02) $+$  & \textbf{2.479E-01 }(7.20E-03) $+$  & 2.532E-01 (9.28E-03) $+$ \\
 & 7 & 4.303E-01 (1.49E-02) $+$  & 4.364E-01 (1.38E-02) $+$  & 4.242E-01 (9.08E-03) $+$  & \textbf{4.127E-01 }(5.93E-03) $+$ \\\midrule
\multirow{3}{*}{WFG9} & 3 & 4.324E-02 (7.07E-04) $+$  & \textbf{4.159E-02 }(6.10E-04) $+$  & 4.359E-02 (1.00E-02) $+$  & 6.415E-02 (2.64E-02) $+$ \\
 & 5 & 1.858E-01 (7.63E-03) $+$  & \textbf{1.814E-01 }(4.59E-03) $+$  & 1.832E-01 (7.10E-03) $+$  & 1.918E-01 (1.13E-02) $+$ \\
 & 7 & 3.328E-01 (1.02E-02) $+$  & 3.298E-01 (1.03E-02) $+$  & 3.307E-01 (1.37E-02) $+$  & \textbf{3.278E-01 }(7.21E-03) $+$ \\\midrule
\multirow{3}{*}{Test: average rank} &3 & 1.4 & 1.8 & 2.8 & 4.0\\
&5 & 2.6 & 2.8 & 2.2 & 2.4\\
&7 & 2.6 & 3.4 & 3.0 & 1.0\\ \midrule
\multirow{3}{*}{Test: $+$/$-$/$\approx$}& 3 & 5/0/0 & 5/0/0 & 5/0/0 & 5/0/0\\
&5 & 4/0/1 & 5/0/0 & 4/0/1 & 4/0/1\\
&7 & 5/0/0 & 4/0/1 & 4/0/1 & 5/0/0\\
\bottomrule
\end{tabular}
}
\label{table:2 mix}
\end{table}

\paragraph{RQ1: How does MA-DAC perform compared with the baseline and other tuning algorithms?}

We compare the MA-DAC policy with the original MOEA/D~\citep{moead}, DQN~\citep{DQN} and MA-UCB~\citep{MAUCB}. DQN is a single-agent RL algorithm that shares the same state, reward, and transition as our MA-DAC but has a different action space that is the concatenation of the four types of hyperparameters. DQN can be seen as an instantiation of DAC in MaMo. MA-UCB is a simplified algorithm that uses four UCB agents to adjust the four types of hyperparameters. For MA-DAC and DQN, we perform the training and testing on the problems with the number of objectives.

The results are shown in Table~\ref{table:1 baseline}. As can be seen in the table, MA-DAC is significantly superior to the other algorithms on almost all the 24 problems. Furthermore, MA-DAC (which is trained only on DTLZ2, WFG4 and WFG6) shows excellent performance on previously unseen problems during the training process, demonstrating its good generalization ability. DQN policy is significantly worse than MA-DAC on all the problems, which is consistent with the previous observation that DQN performs poorly in MARL tasks~\citep{marlsurvey}. DQN suffers from the exponentially increasing action space with the number of agents~\citep{reviewOfCoop,selectiveOverview}, while MA-DAC effectively decomposes the action space, making each agent much easier to find their own near-optimal policy and then leading to a good joint policy. The superiority of MA-DAC over MA-UCB also demonstrates the necessity of learning configuration policy by an MARL algorithm.   

We also compare our algorithm with two adaptive MOEA/D algorithms, i.e., MOEA/D-FRRMAB~\citep{FRRMAB} and MOEA/D-AWA~\citep{AWA}, which adjust reproduction operator types and weights based on MAB and heuristic rules, respectively. The results clearly show the superior performance of MA-DAC, and are provided in Appendix~\ref{analysis of reproduction operators} and Appendix~\ref{analysis of adaptive weights} due to space limitation.

\paragraph{RQ2: How is the generalization ability of MA-DAC?}

We compare MA-DAC policies trained from different training sets to test the generalization ability of MA-DAC policies. MA-DAC (M) denotes that the policy is trained on the problems (i.e., DTLZ2, WFG4 and WFG6) with all 3, 5 and 7 objectives. Meanwhile, MA-DAC (3), (5) and (7) denote that the policies are trained on the problems (i.e., DTLZ2, WFG4 and WFG6)
with 3, 5 and 7 objectives, respectively. 

As shown in Table~\ref{table:2 mix}, it is unsurprising that the results of average rank show that the MA-DAC (3), (5) and (7) policies have excellent performance on the problems with 3, 5 and 7 objectives, respectively. For example, MA-DAC (7) has the best average rank (i.e., 1.0) on the test problems with 7 objectives, but has the worst average rank (i.e., 4.0) on the problems with 3 objectives. To obtain a more robust policy, we mix the problems with
different number of objectives as the training set, resulting in the MA-DAC (M). As can be seen in Table~\ref{table:2 mix}, MA-DAC (M) demonstrates its robustness. Among all the policies, MA-DAC (M) takes the first, third, and second places in terms of its performance in the three types of problems, respectively. Lastly, the last row of the table shows the outperformance of these MA-DAC policies compared with other policies -- the IGD values of all the four MA-DAC policies perform significantly better than the best result obtained by the three peer algorithms MOEA/D, DQN and MA-UCB in Table~\ref{table:1 baseline}. 
\vspace{-0.5em}
\paragraph{RQ3: How do the different parts of MA-DAC affect the performance?}
\label{rq3}

\begin{table}[t!]\scriptsize
\centering
\caption{IGD values obtained by MA-DAC (M) w/o 1, MA-DAC (M) w/o 2, MA-DAC (M) w/o 3 and MA-DAC (M) w/o 4 on different problems. Each result consists of the mean and standard deviation of 30 runs. The best mean value on each problem is highlighted in \textbf{bold}. The symbols `$+$', `$-$' and `$\approx$' indicate that the result is significantly superior to, inferior to, and almost equivalent to MA-DAC (M), respectively, according to the Wilcoxon rank-sum test with significance level 0.05.}\vspace{0.5em}
\resizebox{\linewidth}{!}{
\begin{tabular}{p{0.8cm} c|c c c c c }
\toprule
     Problem     &   $M$  & MA-DAC (M) w/o 1  & MA-DAC (M) w/o 2  & MA-DAC (M) w/o 3 & MA-DAC (M) w/o 4 & MA-DAC (M)   \\\midrule
\multirow{3}{*}{DTLZ2} & 3 & 4.656E-02 (3.80E-04) $-$  & 3.914E-02 (8.43E-04) $-$  & 3.935E-02 (6.72E-04) $-$  & 3.919E-02 (5.91E-04) $-$  & \textbf{3.839E-02 }(5.35E-04)  \\
 & 5 & 3.086E-01 (7.24E-03) $-$  & 2.619E-01 (8.99E-03) $-$  & 2.503E-01 (1.30E-02) $-$  & \textbf{2.433E-01 }(1.59E-02) $\approx$  & 2.468E-01 (7.55E-03)  \\
 & 7 & 4.970E-01 (1.26E-02) $-$  & 4.067E-01 (1.20E-02) $-$  & 4.003E-01 (1.19E-02) $-$  & 4.228E-01 (1.25E-02) $-$  & \textbf{3.921E-01 }(8.84E-03)  \\\midrule

\multirow{3}{*}{WFG4} & 3 & 7.222E-02 (1.93E-03) $-$  & 5.484E-02 (1.01E-03) $-$  & 5.410E-02 (8.85E-04) $-$  & 5.410E-02 (8.85E-04) $-$  & \textbf{5.220E-02 }(9.83E-04)  \\
 & 5 & 2.868E-01 (1.01E-02) $-$  & 1.879E-01 (3.76E-03) $\approx$  & \textbf{1.845E-01 }(2.17E-03) $+$  & 1.846E-01 (2.39E-03) $+$  & 1.850E-01 (3.14E-03)  \\
 & 7 & 3.758E-01 (1.33E-02) $-$  & 3.102E-01 (6.34E-03) $-$  & \textbf{3.020E-01 }(3.99E-03) $\approx$  & 3.032E-01 (4.32E-03) $\approx$  & 3.091E-01 (5.80E-03)  \\\midrule

\multirow{3}{*}{WFG6} & 3 & 6.864E-02 (8.14E-03) $-$  & 5.338E-02 (1.37E-02) $-$  & 6.543E-02 (1.69E-02) $-$  & 6.067E-02 (2.11E-02) $\approx$  & \textbf{5.078E-02 }(1.20E-02)  \\
 & 5 & 3.480E-01 (1.34E-02) $-$  & 2.005E-01 (5.21E-03) $-$  & 1.996E-01 (6.51E-03) $-$  & 1.979E-01 (6.76E-03) $-$  & \textbf{1.971E-01 }(6.40E-03)  \\
 & 7 & 4.784E-01 (3.37E-02) $-$  & 3.147E-01 (5.85E-03) $-$  & 3.162E-01 (6.15E-03) $-$  & 3.147E-01 (5.73E-03) $-$  & \textbf{3.114E-01 }(5.08E-03)  \\\midrule

\multicolumn{2}{c|}{Train: $+$/$-$/$\approx$}  & 0/9/0  & 0/8/1  & 1/7/1  & 1/5/3 & \\\midrule

\multirow{3}{*}{DTLZ4} & 3 & 6.463E-02 (3.85E-02) $-$  & 6.242E-02 (4.07E-02) $-$  & \textbf{4.496E-02 }(2.45E-03) $+$  & 4.496E-02 (2.45E-03) $+$  & 6.171E-02 (3.67E-02)  \\
 & 5 & 3.497E-01 (1.41E-02) $-$  & 3.061E-01 (2.12E-02) $\approx$  & 3.054E-01 (1.55E-02) $\approx$  & 3.130E-01 (1.81E-02) $-$  & \textbf{3.044E-01 }(1.66E-02)  \\
 & 7 & 4.853E-01 (1.82E-02) $-$  & 4.275E-01 (1.90E-02) $-$  & \textbf{4.208E-01 }(1.52E-02) $\approx$  & 4.289E-01 (2.07E-02) $-$  & 4.271E-01 (1.45E-02)  \\\midrule

\multirow{3}{*}{WFG5} & 3 & 6.189E-02 (7.40E-04) $-$  & 4.827E-02 (6.31E-04) $-$  & 4.766E-02 (5.72E-04) $\approx$  & 4.766E-02 (5.72E-04) $\approx$  & \textbf{4.721E-02 }(7.15E-04)  \\
 & 5 & 3.202E-01 (9.77E-03) $-$  & 1.821E-01 (2.73E-03) $\approx$  & 1.835E-01 (2.90E-03) $-$  & 1.822E-01 (2.31E-03) $-$  & \textbf{1.811E-01 }(2.59E-03)  \\
 & 7 & 4.948E-01 (1.47E-02) $-$  & 3.290E-01 (8.87E-03) $-$  & 3.310E-01 (7.70E-03) $-$  & 3.261E-01 (9.26E-03) $-$  & \textbf{3.256E-01 }(5.49E-03)  \\\midrule

\multirow{3}{*}{WFG7} & 3 & 6.004E-02 (9.45E-04) $-$  & 4.250E-02 (5.82E-04) $-$  & 4.150E-02 (6.60E-04) $-$  & 4.150E-02 (6.60E-04) $-$  & \textbf{4.076E-02 }(5.33E-04)  \\
 & 5 & 3.402E-01 (2.49E-02) $-$  & 1.873E-01 (3.67E-03) $\approx$  & \textbf{1.826E-01 }(2.60E-03) $+$  & 1.847E-01 (2.80E-03) $\approx$  & 1.839E-01 (2.38E-03)  \\
 & 7 & 4.877E-01 (5.70E-02) $-$  & 3.393E-01 (1.23E-02) $-$  & 3.373E-01 (1.16E-02) $-$  & 3.377E-01 (1.59E-02) $-$  & \textbf{3.368E-01 }(1.54E-02)  \\\midrule
\multirow{3}{*}{WFG8} & 3 & 9.661E-02 (1.87E-03) $-$  & 8.374E-02 (1.70E-03) $-$  & 8.029E-02 (1.29E-03) $-$  & 8.029E-02 (1.29E-03) $-$  & \textbf{7.828E-02 }(1.46E-03)  \\
 & 5 & 4.119E-01 (1.30E-02) $-$  & 2.695E-01 (1.49E-02) $-$  & 2.571E-01 (1.09E-02) $-$  & 2.632E-01 (9.96E-03) $-$  & \textbf{2.506E-01 }(1.11E-02)  \\
 & 7 & 5.830E-01 (1.59E-02) $-$  & 4.345E-01 (9.27E-03) $-$  & \textbf{4.260E-01 }(8.71E-03) $-$  & 4.322E-01 (1.12E-02) $-$  & 4.303E-01 (1.49E-02)  \\\midrule
\multirow{3}{*}{WFG9} & 3 & 5.894E-02 (9.24E-04) $-$  & \textbf{4.321E-02 }(7.50E-04) $-$  & 4.650E-02 (1.40E-02) $-$  & 4.650E-02 (1.40E-02) $-$  & 4.324E-02 (7.07E-04)  \\
 & 5 & 3.148E-01 (2.06E-02) $-$  & 1.875E-01 (5.14E-03) $-$  & 1.941E-01 (6.45E-03) $-$  & 1.865E-01 (9.02E-03) $-$  & \textbf{1.858E-01 }(7.63E-03)  \\
 & 7 & 5.069E-01 (2.53E-02) $-$  & 3.433E-01 (1.35E-02) $-$  & 3.402E-01 (1.00E-02) $-$  & 3.368E-01 (1.05E-02) $-$  & \textbf{3.328E-01 }(1.02E-02)  \\\midrule

\multicolumn{2}{c|}{Test: $+$/$-$/$\approx$}   & 0/15/0  & 0/12/3  & 2/10/3  & 1/12/2 &  \\
\bottomrule
\end{tabular}}
\label{table:3 ablation}
\end{table}

We conduct ablation studies to show the importance of different types of hyperparameters, and the results are given in Table~\ref{table:3 ablation}. We use MA-DAC (M) w/o $i$ to denote the MA-DAC policy that does not include the $i$-th agent, and we use a reasonable setting as the default for each ablated agent; the detailed settings are provided in Appendix~\ref{app: details exp settings}. That is, MA-DAC (M) w/o $1$, $2$, $3$ and $4$ represent MA-DAC (M) without tuning weights, neighborhood size, types of reproduction operators, and parameters of reproduction operators, respectively. 

As can be seen in Table~\ref{table:3 ablation}. MA-DAC (M) outperforms all the ablations, demonstrating the importance of tuning every type of hyperparameter. Note that the column of MA-DAC (M) in Table~\ref{table:3 ablation} is just as same as that in Table~\ref{table:2 mix}. On the other hand, we notice that the importance of hyperparameters varies. For example, adaptive weights are in general more important as the performance of MA-DAC (M) w/o 1 drops significantly. 
\paragraph{Further studies}
Due to the space limitation, more experiments on MaMo and DACBench~\cite{dacbench} are provided in Appendix~\ref{app: additional results on mamo} and Appendix~\ref{app: exp on dacbench}, respectively. 
\begin{itemize}
    \item To show the effectiveness of the proposed triangle-based reward function, we compare it with different reward functions in Appendix~\ref{app: analysis of reward}.
    \item We give a detailed analysis of the reproduction operators and the adaptive weights in Appendix~\ref{analysis of reproduction operators} and Appendix~\ref{analysis of adaptive weights}, respectively.
    \item To show the optimization process of each method, we plot the curves of IGD value of different methods (i.e., MOEA/D, MOEA/D-FRRMAB, MOEA/D-AWA, DQN, MA-UCB and MA-DAC) in Appendix~\ref{app: IGD values during the optimization process}.
    \item To improve the compared baseline, we use DQN to dynamically adjust each type of hyperparameter of MOEA/D in Appendix~\ref{app: Comparison with different DQN variants}.
    \item We use two more MARL algorithms as the implementations of policy networks, i.e., Independent Q Learning (IQL)~\citep{iql} and QMIX~\citep{qmix} in Appendix~\ref{app: different marl algorithms}.
    \item We also conduct experiments on Sigmoid from DACBench~\cite{dacbench} in Appendix~\ref{app: exp on dacbench}, which can generate instance sets with a wide range of difficulties. We conduct experiments on the $5D$-Sigmoid and $10D$-Sigmoid (i.e., there are $5$ and $10$ agents in MA-DAC, respectively) with action space size $3$ (i.e., each agent has a $3$-dimensional discrete action space). The experimental results show the versatility and scalability of MA-DAC.
\end{itemize}
%

\section{Conclusion}
This paper considers the dynamic configuration of algorithms with multiple types of hyperparameters. We propose MA-DAC to solve it, where one agent works to handle one type of configuration hyperparameter. Experimental results show that MA-DAC works well and has good generalization ability. The instantiation of configuring MOEA/D forms the benchmark MaMo for MARL, with the hope of facilitating the application of MARL. 

Considering the superior performance of MA-DAC in empirical studies, an interesting future work is to perform theoretical analysis, to better understand why MA-DAC can work. Particularly, many MARL algorithms follow the IGM principle that assumes the global Q is factorizable, while it is not yet clear whether DAC problems are (approximately) factorizable. In addition, we will try to propose better contextual MMDP formulation as well as better cooperative MARL algorithms, based on the heterogeneity and stochasticity of MA-DAC. It is also interesting to include more real-world optimization problems into MaMo.

\section*{Acknowledgements}
We thank the reviewers for their insightful and valuable comments. We thank Chenghe Wang and Hao Yin for providing helpful
comments. This work was supported by the National Key Research and Development Program of China
(2020AAA0107200), the NSFC (62022039, 62276124, 61876119), the Fundamental
Research Funds for the Central Universities (0221-14380009, 0221-14380014), and the program B for Outstanding Ph.D. candidate of Nanjing University.

\bibliographystyle{plainnat}
\bibliography{madac}
\section*{Checklist}


\begin{enumerate}

\item For all authors...
\begin{enumerate}
  \item Do the main claims made in the abstract and introduction accurately reflect the paper's contributions and scope? 
    \answerYes{} See Abstract and Section1. 
  \item Did you describe the limitations of your work?
    \answerYes{} We describe the limitations in Section 6.
  \item Did you discuss any potential negative societal impacts of your work?
    \answerYes{} The authors do not see obvious negative societal impacts of the proposed method
  \item Have you read the ethics review guidelines and ensured that your paper conforms to them? 
    \answerYes{} Our paper conforms to them.
\end{enumerate}

\item If you are including theoretical results...
\begin{enumerate}
  \item Did you state the full set of assumptions of all theoretical results?
    \answerNA{}
        \item Did you include complete proofs of all theoretical results?
    \answerNA{}
\end{enumerate}

\item If you ran experiments...
\begin{enumerate}
  \item Did you include the code, data, and instructions needed to reproduce the main experimental results (either in the supplemental material or as a URL)?
    \answerYes{} Yes, see Section 5, Appendix and Supplementary material.
  \item Did you specify all the training details (e.g., data splits, hyperparameters, how they were chosen)?
    \answerYes{} Yes, see Section 5 and Appendix.
        \item Did you report error bars (e.g., with respect to the random seed after running experiments multiple times)?
    \answerYes{} Yes, see Section 5 and Appendix.
        \item Did you include the total amount of compute and the type of resources used (e.g., type of GPUs, internal cluster, or cloud provider)? 
    \answerYes{} Yes, see Appendix.
\end{enumerate}

\item If you are using existing assets (e.g., code, data, models) or curating/releasing new assets...
\begin{enumerate}
  \item If your work uses existing assets, did you cite the creators?
    \answerYes{} See References.
  \item Did you mention the license of the assets?
    \answerNA{}
  \item Did you include any new assets either in the supplemental material or as a URL?
    \answerYes{} See supplemental material.
  \item Did you discuss whether and how consent was obtained from people whose data you're using/curating?
    \answerNA{}
  \item Did you discuss whether the data you are using/curating contains personally identifiable information or offensive content?
    \answerNA{}
\end{enumerate}

\item If you used crowdsourcing or conducted research with human subjects...
\begin{enumerate}
  \item Did you include the full text of instructions given to participants and screenshots, if applicable?
    \answerNA{}
  \item Did you describe any potential participant risks, with links to Institutional Review Board (IRB) approvals, if applicable?
    \answerNA{}
  \item Did you include the estimated hourly wage paid to participants and the total amount spent on participant compensation?
    \answerNA{}
\end{enumerate}

\end{enumerate}

\newpage
\appendix

\section{Details of MA-DAC}\label{app: details of MA-DAC}
\subsection{Details of MOEA/D}
First, we give a brief introduction to multi-objective optimization problems (MOPs), which can be defined as
\begin{equation}\label{eq:mop}
    \min\; \bm F(\bm x) = (f_1(\bm x), \dots, f_m(\bm x)) \quad \text{s.t.}\quad \bm x\in \Omega,
\end{equation}
where $\bm{x} = (x_1,\dots,x_D)$ is a solution, $\bm F: \Omega \to \mathbb R^m$ constitutes $m$ objective functions, $\Omega=[x_i^L,x_i^U]^D\subseteq \mathbb R^D$ is the solution space, and $\mathbb R^m$ is the objective space.
\begin{definition}
A solution $\bm x^*$ is Pareto-optimal with respect to Eq.~(\ref{eq:mop}), if $\nexists \bm x\in \Omega$ such that $\forall i: f_i(\bm{x})\leq f_i(\bm{x}^*)$ and $\exists i: f_i(\bm{x})< f_i(\bm{x}^*)$. The set of all Pareto-optimal solutions is called Pareto-optimal set (PS). The set of the corresponding objective vectors of PS, i.e., $\{\bm{F(\bm{x})\mid\bm{x}\in \text{PS}}\}$, is called Pareto front (PF).
\end{definition}
Instead of focusing on a single optimal solution in single-objective optimization, the goal of MOP is to find at least one Pareto-optimal solution for each objective vector in the PF. However, as the size of PF can be prohibitively large or even infinite, it is often to find a set of solutions that can approximate the PS well, i.e., the set of their objective vectors can approximate the PF well.

Evolutionary algorithms have demonstrated their effectiveness in solving MOPs. Their population-based nature can approximate the Pareto optimal solutions within one execution, with each solution in the population representing a unique trade-off among the objectives. MOEA/D~\citep{moead} is a representative multi-objective evolutionary algorithm. MOEA/D converts an MOP into a number of single-objective sub-problems through a number of weights, where neighboring solutions work cooperatively for the optimal solutions for the single-objective sub-problems. Note that an optimal solution for a single-objective sub-problem must be Pareto optimal for the MOP.

MOEA/D consists of two major processes, i.e., \emph{decomposition} and \emph{collaboration}~\citep{moead, moead-survey, moead-survey2}. In decomposition, MOEA/D transforms the task of approximating the PF into a number of sub-problems through a number of weights and an aggregation function. There have been several aggregation functions for MOEA/D. Here, we introduce the common Tchebycheff approach (TCH) that is also used in this paper. Given a weight vector $\bm{w}=(w_1,\dots,w_m)$ where $w_i\geq 0, \forall i \in \{1,\dots,m\}$ and $\sum^m_{i=1}w_i=1$, the sub-problem by TCH is formulated as
\begin{equation}\label{TCH}
    \min_{\bm{x}\in\Omega}\; g(\bm{x}\mid\bm{w},\bm{z}^*) = \max_{1\leq i \leq m} \{w_i\cdot |f_i(\bm{x}) - z_i^*|\},
\end{equation}
where $\bm{z}^*=(z_1^*,\dots,z_m^*)$ is the ideal point consisting of the best objective values obtained so far.

The basic idea of collaboration is that neighboring sub-problems are more likely to share similar properties, e.g., similar objective functions and/or optimal solutions~\citep{moead-survey2}. In particular, the neighborhood of a sub-problem is determined by the Euclidean distance of its corresponding weight vector with respect to the others and the hyperparameter \emph{neighborhood size}: if the distance between two sub-problems is smaller than the neighborhood size, they are the neighborhood of each other. In the mating selection of a sub-problem, the parent solutions are randomly selected from its neighborhood, and the newly generated offspring solution is used to update the solutions of sub-problems within the same neighborhood.

\begin{algorithm}[t!]
\SetAlgoLined
\caption{MOEA/D}\label{moead}
\SetKwInput{KWInput}{Parameters}
\SetKwInput{KWOutput}{Output}
\KWInput{Population size $N$, number $T$ of iterations}
Initialize a population $\{\bm x^{(i)}\}^N_{i=1}$ of solutions, and a corresponding set $W=\{\bm{w}^{(i)}\}^N_{i=1}$ of weight vectors \;
$t=0$ \;
\While{$t < T$}{
\For{$i=1:N$}{Randomly select parent solutions from the neighborhood of $\bm{w}^{(i)}$ , denoted as $\Theta^{\bm{w}^{(i)}}$ \;
Use crossover and mutation operators to generate an offspring solution $\bm {x}'^{(i)}$\;
Evaluate the offspring solution to obtain $\bm{F}(\bm {x}'^{(i)})$\;
Update the ideal point $\bm {z}^*$. That is, for any $j \in \{1,2,\ldots,m\}$, if $f_j(\bm {x}'^{(i)})<\bm {z}^*_j$, then $\bm {z}^*_j=f_j(\bm {x}'^{(i)})$\;
Update the corresponding solution of each sub-problem within $\Theta^{\bm{w}^{(i)}}$ by $\bm {x}'^{(i)}$. That is, for each $\bm{w}^{(j)}\in \Theta^{\bm{w}^{(i)}}$, if $g(\bm {x}'^{(i)}\mid \bm{w}^{(j)},\bm {z}^*)<g(\bm{x}^{(j)}\mid \bm{w}^{(j)},\bm {z}^*)$, then $\bm{x}^{(j)}=\bm {x}'^{(i)}$
}
$t=t+1$
}
\end{algorithm}

The procedure of MOEA/D used in this paper is described in Algorithm~\ref{moead}. Firstly, it generates a population $\{\bm x^{(i)}\}^N_{i=1}$ of solutions with size $N$, associated with $N$ weight vectors $\{\bm{w}^{(i)}\}^N_{i=1}$ in line~1. A weight vector $\bm{w}^{(i)}$ corresponds to a single-objective sub-problem, and $\bm x^{(i)}$ is the current best solution associated with this sub-problem. Then, in each iteration (i.e., lines~4--10) of MOEA/D, for each sub-problem, it selects parent solutions from the neighborhood, generates an offspring solution by reproduction operators, and updates the solutions of the sub-problem and its neighboring sub-problem(s). After obtaining parent solutions in line~5, it uses crossover operators (e.g., simulated binary crossover (SBX) operator or differential evolution (DE) operator) and polynomial mutation (PM) operator~\citep{moead}, to generate an offspring solution $\bm {x}'^{(i)}$ in line~6. Then, it evaluates the offspring solution and obtain the objective vector $\bm{F}(\bm {x}'^{(i)})$ in line~7. Finally, it uses the offspring solution to update the ideal point $\bm{z}^*$ in line~8 and the solutions of the sub-problems within the neighborhood $\Theta^{\bm{w}^{(i)}}$ of the current sub-problem $\bm{w}^{(i)}$ in line~9. For each $j \in \{1,2,\ldots,m\}$, $z^*_j$ is the best found value of the $j$-th objective $f_j$, and thus if $f_j(\bm {x}'^{(i)})$ is better, i.e., $f_j(\bm {x}'^{(i)})<z^*_j$, then $z^*_j$ will be udpated accordingly. For each sub-problem $\bm{w}^{(j)}$ within $\Theta^{\bm{w}^{(i)}}$, if $\bm {x}'^{(i)}$ is better than its corresponding solution $\bm {x}^{(j)}$, i.e., $g(\bm {x}'^{(i)}\mid \bm{w}^{(j)},\bm {z}^*)<g(\bm{x}^{(j)}\mid \bm{w}^{(j)},\bm {z}^*)$, then $\bm {x}^{(j)}$ will be udpated accordingly.

\subsection{State formulation}\label{app a: state}
The state of our proposed benchmark, i.e., MaMo, can be divided into three parts.
\begin{enumerate}
    \item To describe the general properties of the optimization problems, the first part (i.e., indexes~0--1 in Table~\ref{table:state}) contains the features of the problem instance, i.e., the numbers of objectives and variables.  
    \item To emphasize the general information of the algorithm, the second part (i.e., indexes~2--3 in Table~\ref{table:state}) contains the features of the optimization process, i.e., how much computational budget has been used and how many steps of the algorithm have not made any progress, i.e., stagnant count ratio. 
    \item To show the properties of the population and that how the population is evolved, we use several indicators, i.e., hypervolume~\citep{hv}, ratio of non-dominated solutions in the population, and average distance of the solutions~\citep{indicators-survey} in the third part (i.e., indexes~4--21 in Table~\ref{table:state}). For each indicator, we also use the gap between the current value and the value corresponding to the last population to reflect the immediate evolutionary progress. Besides, we use statistic metrics (i.e., the mean and standard deviation) of these indicators in the last five steps and all steps from the beginning to characterize the short and long histories of the optimization, respectively. 
\end{enumerate}

The detailed state features at step $t$ are shown in Table~\ref{table:state}. 
We use $\operatorname{List}(I,t,l)$ to denote a list of indicator values from step $t-l+1$ to $t$, i.e., $[I_{t-l+1},\ldots,I_t]$, where $I$ denotes a specific indicator and $I_i$ denotes the indicator value at step $i$. Note that $\forall i<0, I_{i}=0$ as default. For the considered three indicators, i.e., hypervolume, ratio of non-dominated solutions in the population and average distance of the solutions, they are denoted as $\operatorname{HV}$, $\operatorname{NDRatio}$ and $\operatorname{Dist}$, respectively. 

To facilitate the generalization of the learned MA-DAC policy, we pre-process some state features to make them in $[-1,1]$.
For state features 0 and 1, we use $1/m$ and $1/D$, respectively, where $m$ is the number of objectives and $D$ is the number of variables. They apparently belong to $[0,1]$. For state features 2, 3, 4 and 5, they are defined in $[0,1]$. State feature 6 is the average distance between the solutions in the population. We sample a sufficient number of solutions before optimization, and calculate the maximum distance between them, which is used as a scaling denominator when calculating state feature 6. State features 7-9 are the difference of two values in $[0,1]$, and thus belong to $[-1,1]$. For state features 10-21, they are statistical values (i.e., mean and standard deviation) of state features 4-6.

\begin{table}[t!]
    \centering\small
    \caption{State at step $t$ in MaMo.}
    \begin{tabular}{c c c c}
    \toprule
    Index & Parts of state &Feature & Notes \\ \toprule
    0 & 1 & $1/m$ & $m$: Number of objectives \\
    1 & 1 &$1/D$ & $D$: Number of variables \\ \toprule
    2 & 2 & $t/T$ & Computational budget that has been used \\
    3 & 2 & $N_{\text{stag}}/T$ & Stagnant count ratio \\ \toprule
    4 & 3 & $\operatorname{HV}_t$ & Hypervolume value \\
    5 & 3 & $\operatorname{NDRatio}_t$ & Ratio of non-dominated solutions  \\
    6 & 3 & $\operatorname{Dist}_t$ & Average distance \\ \midrule
    7 & 3 & $\operatorname{HV}_t - \operatorname{HV}_{t-1}$ & Change of HV between steps $t$ and $t-1$ \\
    8 & 3 & $\operatorname{NDRatio}_t - \operatorname{NDRatio}_{t-1}$ & Change of NDRatio between steps $t$ and $t-1$ \\
    9 & 3 &$\operatorname{Dist}_t - \operatorname{Dist}_{t-1}$ & Change of Dist between steps $t$ and $t-1$    \\ \midrule
    10 & 3 &$\operatorname{Mean}(\operatorname{List}(\operatorname{HV}, t, 5))$ & Mean of HV in the last 5 steps \\
    11 & 3 &$\operatorname{Mean}(\operatorname{List}(\operatorname{NDRatio}, t, 5))$ & Mean of NDRatio in the last 5 steps \\
    12 & 3 &$\operatorname{Mean}(\operatorname{List}(\operatorname{Dist}, t, 5))$ & Mean of Dist in the last 5 steps\\
    13 & 3 &$\operatorname{Std}(\operatorname{List}(\operatorname{HV}, t, 5))$ & Standard deviation of HV in the last 5 steps\\
    14 & 3 &$\operatorname{Std}(\operatorname{List}(\operatorname{NDRatio}, t, 5))$ & Standard deviation of NDRatio in the last 5 steps\\
    15 & 3 &$\operatorname{Std}(\operatorname{List}(\operatorname{Dist}, t, 5))$ & Standard deviation of Dist in the last 5 steps\\ \midrule
    16 & 3 &$\operatorname{Mean}(\operatorname{List}(\operatorname{HV}, t, t))$ & Mean of HV in all the steps so far\\
    17 & 3 &$\operatorname{Mean}(\operatorname{List}(\operatorname{NDRatio}, t, t))$ & Mean of NDRatio in all the steps so far \\
    18 & 3 &$\operatorname{Mean}(\operatorname{List}(\operatorname{Dist}, t, t))$ & Mean of Dist in all the steps so far\\
    19 & 3 &$\operatorname{Std}(\operatorname{List}(\operatorname{HV}, t, t))$ & Standard deviation of HV in all the steps so far \\
    20 & 3 &$\operatorname{Std}(\operatorname{List}(\operatorname{NDRatio}, t, t))$ & Standard deviation of NDRatio in all the steps so far\\
    21 & 3 &$\operatorname{Std}(\operatorname{List}(\operatorname{Dist}, t, t))$ & Standard deviation of Dist in all the steps so far\\
    \bottomrule
    \end{tabular}
    \label{table:state}
\end{table}

\subsection{Action formulation}\label{app:action}
We consider four heterogeneous types of configuration hyperparameters in MOEA/D as the actions of four different agents of MA-DAC.

\paragraph{Weights.} In MOEA/D, weights are used to transform an MOP into multiple single-objective sub-problems, which should be as diverse as possible~\citep{moead}. Inspired by MOEA/D-AWA~\citep{AWA}, the action space for weights is discrete with two dimensions, i.e., adjusting (T) and not adjusting (N) the weights. Furthermore, we limit the frequency of adjustment because too frequent adjustment will lead to drastic changes in the sub-problems and is detrimental to the optimization process~\citep{AWA}. If the action is T, weights will be updated before selecting the parent solutions. The weights adaptation mechanism is as follows.

We first calculate the sparsity level of each solution $\bm{x}^{(i)}$ based on vicinity distance~\citep{vicinity}:
\begin{equation}\label{TCH}
S L\left(\bm{x}^{(i)},\{\bm{x}^{(p)}\}^N_{p=1}\right)=\prod_{j=1}^{m} l(\bm{x}^{(i)},j),
\end{equation}
where $l(\bm{x}^{(i)},j)$ is the Euclidean distance between $\bm{x}^{(i)}$ and its $j$-th nearest neighbor in the population $\{\bm{x}^{(p)}\}^N_{p=1}$. The $m$ closest neighbors in the population are used for calculation, where $m$ is the number of objectives. After calculating the sparsity level of each solution, the sub-problems corresponding to the solutions whose sparsity levels are ranked bottom 5\%, i.e., the overcrowded solutions, will be removed. 

To ensure that there are still $N$ sub-problems in total, we should add $0.05N$ new sub-problems and their corresponding solutions. The newly added solutions are from an elite population, which stores all historical non-dominated solutions with a capacity of $1.5N$. If the size of the elite population exceeds the capacity, the solutions with the lowest sparsity level will be removed. For each solution $\bm{x}'$ in the elite population, we calculate its sparsity level with respect to the current population, i.e., $SL(\bm{x}', \text{Pop})$, where $\text{Pop}$ denotes the set of $0.95N$ solutions in the current population after removing the overcrowded solutions. Then, we select the solution from the elite population, which has the highest sparsity level with respect to the current population, and add it to the current population; this process is repeated for $0.05N$ times. For each newly added solution, the corresponding sub-problem (i.e., weight vector) is generated in a specific way, whose details can refer to Algorithm~3 in~\citep{AWA}.

\paragraph{Neighborhood size.} The neighborhood size is to control the distance between solutions in mating selection. A small size helps the search exploit the local area, while a large size helps the search explore a wide objective space~\citep{AGR}. We discretize the action space into four dimensions, i.e., $15$, $20$, $25$ and $30$, where $20$ is the default value.

\paragraph{Types of reproduction operators.} We consider four types of DE operators with different search abilities introduced in~\citep{FRRMAB}. Assuming that we are reproducing an offspring solution for the $i$-th sub-problem. Let $\bm{x}^{(i)}$ and $\bm {x}'^{(i)}$ denote its current solution and the generated offspring solution, respectively. The equations of four types of DE operators are shown as follows:
\begin{itemize}
    \item OP1: 
    $\bm {x}'^{(i)}=\bm{x}^{(i)}+F \times\left(\bm{x}^{(r_{1})}-\bm{x}^{(r_{2})}\right)$,
    
    \item OP2:
    $
    \bm {x}'^{(i)}=\bm{x}^{(i)}+F \times\left(\bm{x}^{(r_{1})}-\bm{x}^{(r_{2})}\right)+F \times\left(\bm{x}^{(r_{3})}-\bm{x}^{(r_{4})}\right)
    $,
    \item OP3:
    $
    \bm {x}'^{(i)}=\bm{x}^{(i)}+K \times\left(\bm{x}^{(i)}-\bm{x}^{(r_{1})}\right)+F \times\left(\bm{x}^{(r_{2})}-\bm{x}^{(r_{3})}\right)+F \times\left(\bm{x}^{(r_{4})}-\bm{x}^{(r_{5})}\right)
    $,
    \item OP4:
    $
    \bm {x}'^{(i)}=\bm{x}^{(i)}+K \times\left(\bm{x}^{(i)}-\bm{x}^{(r_{1})}\right)+F \times\left(\bm{x}^{(r_{2})}-\bm{x}^{(r_{3})}\right).
    $
\end{itemize}
Here, $\bm{x}^{(r_{1})}, \bm{x}^{(r_{2})}, \bm{x}^{(r_{3})}, \bm{x}^{(r_{4})}$, and $\bm{x}^{(r_{5})}$ are different parent solutions randomly selected from the neighborhood of $\bm{x}^{(i)}$. The scaling factor $F>0$ controls the impact of the vector differences on the mutant vector, and $K \in[0,1]$ plays a similar role to $F$.

\paragraph{Parameters of reproduction operators.} The parameters (e.g., scaling factor) of the reproduction operators in MOEA/D significantly affect the algorithm's performance~\citep{LED}. We set the scaling factor $K$ to a fixed value of 0.5 as recommended~\citep{FRRMAB}, and dynamically adjust the scaling factor $F$. The action space has four discrete dimensions, i.e., $0.4$, $0.5$, $0.6$ and $0.7$, where $0.5$ is the default value.

\section{Additional results on MaMo} 
\label{app: additional results on mamo}
\subsection{Details of experimental settings} 
\label{app: details exp settings}
\paragraph{Common settings of MOEA/D} We implement MOEA/D with \texttt{Platypus}.\footnote{\url{https://github.com/Project-Platypus/Platypus}} All algorithms mentioned in this paper use the same common settings~\citep{moead,moead-dqn}, as shown in Table~\ref{table:moead settings}.

\begin{table}[h!]
    \centering
    \caption{Common Settings of MOEA/D.}
    \begin{tabular}{c|c}
    \toprule
    \multicolumn{2}{c}{General settings}\\ \midrule
    Population size $N$ & $210$ \\ 
    
    Number $T$ of iterations & $100\times m$ \\
    
    \midrule \multicolumn{2}{c}{Reproduction operators}\\ \midrule
    Crossover operator & Simulated binary crossover (SBX)  \\
    Distribution index of SBX & $20$ \\
    Mutation operator & Polynomial mutation (PM) \\
    Probability of PM & $1/D$ \\
    Distribution index of PM & $20$ \\
    \midrule \multicolumn{2}{c}{Aggregation function}\\ \midrule
    Aggregation function & Tchebycheff approach \\
    Neighborhood size & $20$ \\
    \bottomrule
    \end{tabular}
    \label{table:moead settings}
\end{table}

\paragraph{DQN} We implement DQN with the~\texttt{tianshou}\footnote{\url{https://github.com/thu-ml/tianshou}}~\citep{tianshou} framework and adjust some of the hyperparameters to fit this new task. The network structure is:
\begin{align*}
        \text{state} &\to \textbf{MLP}(128)\to \textbf{relu}
        \to \textbf{MLP}(128)\to \textbf{relu} 
        \to \textbf{MLP}(128) \\
        &\to \textbf{relu}\to \textbf{MLP}(\text{number of actions})
\end{align*}
where $\textbf{MLP}(n)$ means a fully-connected layer with output size of $n$, and $\textbf{relu}$ means Rectified Linear Units. Here, the action apace is the concatenation of the four types of configuration hyperparameters, with a dimension of $128$ (i.e., $4\times 4\times 4\times 2$). Some key hyperparameters of DQN are as follows:
\begin{itemize}
    \item The learning rate is $3e$-$4$.
    \item The discounting factor $\gamma$ is $0.99$.
    \item The buffer size is $50,000$ (where unit is transition).
    \item The number of training steps is $400,000$. 
\end{itemize}

\paragraph{MA-UCB} MA-UCB uses four upper confidence bound (UCB)~\citep{UCB} agents to adjust the four types of hyperparameters~\citep{MAUCB}. Each agent follows the UCB action selection rule, i.e., the action taken by agent $i$ at step $t$ is
\begin{equation}
    a_{t}^{(i)} \doteq \underset{a^{(i)}}{\arg \max }\left[Q_{t}(a^{(i)})+c \sqrt{\frac{\ln t}{N_{t}(a^{(i)})}}\right],
\end{equation}
where $Q_t(a^{(i)})$ denotes the estimated value of $a^{(i)}$ at step $t$, $N_t(a^{(i)})$ denotes the number of times that action $a^{(i)}$ has been selected at step $t$, and the number $c>0$ (the value is $1.0$ here) controls the degree of exploration.

\paragraph{MOEA/D-FRRMAB} We modify the implementation of FRRMAB from the~\texttt{PlatEMO}\footnote{\url{https://github.com/BIMK/PlatEMO}}~\citep{platemo} framework to make a fair comparison. That is, the original adaptive operator selection mechanism and related hyperparameters are retained, except that it uses the same settings of MOEA/D as MA-DAC. MOEA/D-FRRMAB adjusts the four types of DE operators by MAB.
In particular, we searched for some sensitive hyperparameters according to the suggestions in~\citep{FRRMAB}, and the best performing combination is shown as follows:

\begin{itemize}
    \item Scaling factor is $2.0$.
    \item Size of the sliding window is $0.5\times N$.
    \item Decaying factor is $0.3$.
\end{itemize}

\paragraph{MOEA/D-AWA} We modify the implementation of MOEA/D-AWA from the \texttt{PlatEMO} framework to make a fair comparison. That is, the original adaptive weight vector adjustment strategy and related hyperparameters are retained, except that it uses the same settings of MOEA/D as MA-DAC. 

\paragraph{MA-DAC} We use the default VDN policy network without parameter sharing in the~\texttt{EPyMARL}\footnote{\url{https://github.com/uoe-agents/epymarl}}~\citep{epymarl} framework. MA-DAC and all its variants use the same hyperparameters. Some key hyperparameters are as follows:
\begin{itemize}
    \item The learning rate is $1e$-$4$.
    \item The discounting factor $\gamma$ is $0.99$.
    \item The buffer size is $5,000$ (where unit is episode).
    \item The number of training steps is $400,000$. 
\end{itemize}
All hyperparameters of the above algorithms can be found in the code.

\paragraph{Computing resources} The experiments are conducted on six PCs with an AMD Ryzen 9 3950X 16-Core Processor and an NVIDIA GeForce RTX 3090 GPU.

\subsection{Analysis of the reward function} \label{app: analysis of reward}
In this subsection, we compare our proposed reward function with the three types of reward functions proposed by~\citep{DEDDQN}, as shown in the following:
\begin{align}
    r_t^1= &  \max\{f(s_t)-f(s_{t+1}),0\},\\
    r_t^2= &  \begin{cases}
    10 & 
    \text { if } 
    f(s_{t+1})<f_t^* \\ 
    1 & 
    \text { else if } 
    f(s_{t+1})<f(s_{t}) \\ 
    0 & 
    \text { otherwise }\end{cases},\\
    r_t^3= &  \max \left\{
    \frac{f(s_t)-f(s_{t+1})}{f(s_{t+1})-f_\text{opt}}
    ,0 \right\},
\end{align}
where $f(s_t)$ is the metric value at step $t$, $f_t^*$ is the minimum metric value achieved until step $t$, and $f_\text{opt}$ is the optimal metric value, i.e., the global minimum value. Here, we use IGD~\citep{igd} as the metric $f(\cdot)$, and thus $f_\text{opt}=0$. We train MA-DAC policy with these three reward functions $r_t^1$, $r_t^2$ and $r_t^3$, which are denoted as MA-DAC-R1, MA-DAC-R2 and MA-DAC-R3, respectively. 

The experimental results are shown in Table~\ref{table:4 reward}. We can see that MA-DAC has the best average rank, indicating the effectiveness of our proposed reward function. For the other three methods, MA-DAC-R2 and MA-DAC-R3 are better than MA-DAC-R1, which is consistent with the observation in~\citep{DEDDQN}.

\begin{table}[htbp!]\scriptsize
\centering
\caption{IGD values obtained by MA-DAC-R1, MA-DAC-R2, MA-DAC-R3 and MA-DAC on different problems. Each result consists of the mean and standard deviation of 30 runs. The best mean value on each problem is highlighted in \textbf{bold}. The symbols `$+$', `$-$' and `$\approx$' indicate that the result is significantly superior to, inferior to, and almost equivalent to MA-DAC, respectively, according to the Wilcoxon rank-sum test with significance level 0.05.}\vspace{0.5em}
\resizebox{\linewidth}{!}{
\begin{tabular}{c c|c c c c c }
\toprule
     Problem     &   $M$  & MA-DAC-R1  & MA-DAC-R2 & MA-DAC-R3 & MA-DAC \\\midrule
\multirow{3}{*}{DTLZ2} & 3 & 4.223E-02 (2.50E-03) $-$  & 3.853E-02 (5.58E-04) $-$  & 3.809E-02 (4.64E-04) $\approx$  & \textbf{3.807E-02 }(5.05E-04)  \\
 & 5 & 2.401E-01 (8.27E-03) $-$  & 2.726E-01 (1.51E-02) $\approx$  & \textbf{2.364E-01 }(1.04E-02) $+$  & 2.442E-01 (1.26E-02)  \\
 & 7 & 4.142E-01 (1.12E-02) $-$  & 4.248E-01 (1.30E-02) $-$  & 4.215E-01 (9.03E-03) $-$  & \textbf{3.944E-01 }(1.17E-02)  \\\midrule

\multirow{3}{*}{WFG4} & 3 & 5.989E-02 (5.60E-03) $\approx$  & 5.255E-02 (1.14E-03) $-$  & 5.309E-02 (8.02E-04) $-$  & \textbf{5.200E-02 }(1.19E-03)  \\
 & 5 & 1.848E-01 (2.61E-03) $+$  & 1.851E-01 (2.43E-03) $+$  & \textbf{1.846E-01 }(2.20E-03) $+$  & 1.868E-01 (2.81E-03)  \\
 & 7 & 3.028E-01 (3.19E-03) $+$  & \textbf{3.008E-01 }(3.51E-03) $\approx$  & 3.029E-01 (3.36E-03) $\approx$  & 3.033E-01 (3.66E-03)  \\\midrule

\multirow{3}{*}{WFG6} & 3 & 7.920E-02 (1.81E-02) $+$  & 4.909E-02 (1.50E-02) $-$  & \textbf{4.814E-02 }(1.22E-02) $\approx$  & 4.831E-02 (8.95E-03)  \\
 & 5 & 1.977E-01 (6.17E-03) $-$  & 2.037E-01 (4.49E-03) $-$  & 1.975E-01 (5.78E-03) $-$  & \textbf{1.942E-01 }(6.90E-03)  \\
 & 7 & \textbf{3.110E-01 }(4.86E-03) $-$  & 3.151E-01 (5.01E-03) $\approx$  & 3.148E-01 (4.05E-03) $-$  & 3.112E-01 (4.93E-03)  \\\midrule
 
 \multicolumn{2}{c|}{Train: average rank}  & 2.67 & 3.11 & 2.11 & 2.11\\
\midrule 
 \multicolumn{2}{c|}{Train: $+$/$-$/$\approx$}  & 3/5/1  & 1/5/3  & 2/4/3  &  \\
\midrule  
 
\multirow{3}{*}{DTLZ4} & 3 & \textbf{5.567E-02 }(7.33E-03) $-$  & 7.236E-02 (6.19E-02) $-$  & 6.144E-02 (5.10E-02) $\approx$  & 6.700E-02 (6.14E-02)  \\
 & 5 & 3.119E-01 (1.91E-02) $-$  & 3.221E-01 (2.12E-02) $-$  & 3.119E-01 (1.58E-02) $-$  & \textbf{2.995E-01 }(2.10E-02)  \\
 & 7 & 4.354E-01 (1.29E-02) $-$  & 4.385E-01 (1.23E-02) $-$  & 4.275E-01 (1.60E-02) $-$  & \textbf{4.182E-01 }(1.21E-02)  \\\midrule
 
 \multirow{3}{*}{WFG5} & 3 & 4.841E-02 (7.78E-04) $-$  & 4.763E-02 (7.73E-04) $-$  & 4.773E-02 (6.58E-04) $-$  & \textbf{4.730E-02 }(7.89E-04)  \\
 & 5 & 1.823E-01 (2.49E-03) $\approx$  & 1.818E-01 (2.90E-03) $-$  & 1.812E-01 (3.06E-03) $\approx$  & \textbf{1.811E-01 }(3.02E-03)  \\
 & 7 & 3.212E-01 (6.60E-03) $\approx$  & \textbf{3.174E-01 }(6.43E-03) $\approx$  & 3.196E-01 (5.99E-03) $\approx$  & 3.206E-01 (8.04E-03)  \\\midrule

\multirow{3}{*}{WFG7} & 3 & 4.555E-02 (1.26E-03) $\approx$  & 4.076E-02 (5.41E-04) $-$  & 4.168E-02 (6.40E-04) $-$  & \textbf{4.066E-02 }(5.31E-04)  \\
 & 5 & 1.842E-01 (3.28E-03) $\approx$  & 1.865E-01 (2.93E-03) $+$  & \textbf{1.841E-01 }(3.95E-03) $+$  & 1.858E-01 (2.12E-03)  \\
 & 7 & 3.335E-01 (1.09E-02) $+$  & \textbf{3.199E-01 }(9.86E-03) $-$  & 3.271E-01 (9.65E-03) $\approx$  & 3.258E-01 (1.25E-02)  \\\midrule
\multirow{3}{*}{WFG8} & 3 & 8.914E-02 (2.96E-03) $\approx$  & 7.911E-02 (1.06E-03) $-$  & 8.199E-02 (1.96E-03) $-$  & \textbf{7.901E-02 }(1.19E-03)  \\
 & 5 & 2.551E-01 (1.02E-02) $-$  & 2.628E-01 (1.22E-02) $-$  & 2.541E-01 (9.08E-03) $-$  & \textbf{2.479E-01 }(7.20E-03)  \\
 & 7 & 4.163E-01 (9.54E-03) $\approx$  & \textbf{4.115E-01 }(9.80E-03) $\approx$  & 4.197E-01 (7.52E-03) $-$  & 4.127E-01 (5.93E-03)  \\\midrule
\multirow{3}{*}{WFG9} & 3 & 5.003E-02 (9.00E-03) $-$  & 4.208E-02 (6.56E-04) $-$  & 4.428E-02 (9.97E-03) $-$  & \textbf{4.159E-02 }(6.10E-04)  \\
 & 5 & 1.929E-01 (8.84E-03) $\approx$  & \textbf{1.819E-01 }(5.73E-03) $-$  & 1.951E-01 (9.83E-03) $-$  & 1.832E-01 (7.10E-03)  \\
 & 7 & 3.342E-01 (8.56E-03) $-$  & 3.322E-01 (8.89E-03) $-$  & 3.327E-01 (8.02E-03) $-$  & \textbf{3.278E-01 }(7.21E-03)  \\\midrule

\multicolumn{2}{c|}{Test: average rank}  & 3.27 & 2.47 & 2.67 & 1.60\\\midrule  

\multicolumn{2}{c|}{Test: $+$/$-$/$\approx$}  & 1/7/7  & 1/12/2  & 1/10/4  &  \\

\bottomrule
\end{tabular}}
\label{table:4 reward}
\end{table}

\subsection{Analysis of the reproduction operators}
\label{analysis of reproduction operators}
In this subsection, we give a detailed analysis  of the reproduction operators, including the four types of DE operators introduced in Appendix~\ref{app:action}, and also further compare MA-DAC with MOEA/D-FRRMAB, which applies the MAB-based adaptive tuning method FRRMAB~\citep{FRRMAB} to dynamically adjust the types of DE operators used in MOEA/D. 

First, we examine the performance of MOEA/D equipped with each type of DE operator, where the DE operator is used as the crossover operator with a default scaling factor $F=0.5$. The results are shown in Table~\ref{table:5 op}. Compared with the original MOEA/D using the SBX operator, these methods using the DE operator achieve a similar performance, as the numbers of `$+$' and `$-$' are close. Among the methods using the DE operator, MOEA/D-OP2 has the best average rank, which has thus also been used as the default DE operator in MA-DAC (M) w/o 3. Note that MA-DAC (M) w/o 3 denotes MA-DAC (M) without tuning the types of reproduction operators, which is used to validate the effectiveness of adjusting all configuration hyperparameters simultaneously in RQ3 of the main paper.

\begin{table}[htbp]\scriptsize
\centering
\caption{IGD values obtained by MOEA/D-OP1, MOEA/D-OP2, MOEA/D-OP3 and MOEA/D-OP4 on different problems. Each result consists of the mean and standard deviation of 30 runs. The best mean value on each problem is highlighted in \textbf{bold}. The symbols `$+$', `$-$' and `$\approx$' indicate that the result is significantly superior to, inferior to, and almost equivalent to the original MOEA/D (i.e., the column MOEA/D in Table~2 of the main paper or Table~\ref{table:6 frrmab}), respectively, according to the Wilcoxon rank-sum test with significance level 0.05.}\vspace{0.5em}
\resizebox{\linewidth}{!}{
\begin{tabular}{c c|c c c c c }
\toprule
     Problem     &   $M$  & MOEA/D-OP1  & MOEA/D-OP2 & MOEA/D-OP3 & MOEA/D-OP4 \\\midrule
\multirow{3}{*}{DTLZ2} & 3 & \textbf{4.681E-02 }(2.95E-04) $-$  & 4.691E-02 (3.97E-04) $-$  & 6.050E-02 (2.64E-03) $-$  & 5.033E-02 (1.07E-03) $-$ \\
 & 5 & 3.037E-01 (9.85E-04) $-$  & \textbf{3.012E-01 }(1.51E-03) $\approx$  & 3.391E-01 (1.06E-02) $-$  & 3.083E-01 (2.69E-03) $-$ \\
 & 7 & 4.735E-01 (9.68E-03) $-$  & \textbf{4.551E-01 }(4.43E-03) $-$  & 4.988E-01 (1.01E-02) $-$  & 4.887E-01 (9.51E-03) $-$ \\\midrule

\multirow{3}{*}{WFG4} & 3 & \textbf{6.934E-02 }(1.54E-03) $-$  & 7.293E-02 (1.43E-03) $-$  & 9.046E-02 (4.01E-03) $-$  & 7.998E-02 (2.25E-03) $-$ \\
 & 5 & 2.930E-01 (1.03E-02) $+$  & 2.761E-01 (6.39E-03) $+$  & 2.762E-01 (5.86E-03) $+$  & \textbf{2.761E-01 }(7.63E-03) $+$ \\
 & 7 & 4.057E-01 (1.45E-02) $+$  & 3.711E-01 (9.79E-03) $+$  & \textbf{3.617E-01 }(6.46E-03) $+$  & 3.696E-01 (1.06E-02) $+$ \\\midrule

\multirow{3}{*}{WFG6} & 3 & 7.470E-02 (2.22E-02) $+$  & \textbf{6.714E-02 }(1.59E-02) $+$  & 7.665E-02 (9.41E-03) $-$  & 9.557E-02 (1.71E-02) $-$ \\
 & 5 & 3.513E-01 (1.46E-02) $+$  & 3.285E-01 (2.33E-02) $+$  & \textbf{3.254E-01 }(1.39E-02) $+$  & 3.421E-01 (1.30E-02) $+$ \\
 & 7 & 4.918E-01 (3.31E-02) $\approx$  & 4.797E-01 (3.04E-02) $\approx$  & \textbf{4.328E-01 }(2.81E-02) $+$  & 4.478E-01 (3.08E-02) $+$ \\\midrule
 
\multirow{3}{*}{DTLZ4} & 3 & 7.897E-02 (6.36E-02) $-$  & \textbf{6.226E-02 }(4.05E-03) $-$  & 1.296E-01 (1.23E-02) $-$  & 7.890E-02 (9.62E-03) $-$ \\
 & 5 & 3.504E-01 (2.77E-02) $-$  & \textbf{3.413E-01 }(1.48E-02) $-$  & 3.631E-01 (7.24E-03) $-$  & 3.521E-01 (1.23E-02) $-$ \\
 & 7 & 4.923E-01 (1.89E-02) $-$  & \textbf{4.519E-01 }(1.15E-02) $-$  & 4.766E-01 (1.35E-02) $-$  & 4.975E-01 (2.23E-02) $-$ \\\midrule
 
 \multirow{3}{*}{WFG5} & 3 & 6.181E-02 (5.85E-04) $+$  & 6.177E-02 (8.01E-04) $+$  & 6.128E-02 (5.59E-04) $+$  & \textbf{6.113E-02 }(5.30E-04) $+$ \\
 & 5 & 3.138E-01 (6.20E-03) $+$  & 3.052E-01 (7.19E-03) $+$  & \textbf{3.031E-01 }(7.37E-03) $+$  & 3.116E-01 (8.26E-03) $+$ \\
 & 7 & \textbf{4.945E-01 }(1.24E-02) $-$  & 4.988E-01 (1.04E-02) $-$  & 5.197E-01 (1.01E-02) $-$  & 5.189E-01 (1.22E-02) $-$ \\\midrule
 
\multirow{3}{*}{WFG7} & 3 & \textbf{5.929E-02 }(6.35E-04) $-$  & 6.033E-02 (8.84E-04) $-$  & 8.382E-02 (4.86E-03) $-$  & 6.699E-02 (1.75E-03) $-$ \\
 & 5 & 3.286E-01 (1.55E-02) $+$  & 2.941E-01 (9.66E-03) $+$  & \textbf{2.924E-01 }(1.12E-02) $+$  & 3.148E-01 (1.58E-02) $+$ \\
 & 7 & 5.062E-01 (2.46E-02) $+$  & 4.739E-01 (2.51E-02) $+$  & \textbf{4.479E-01 }(2.28E-02) $+$  & 4.859E-01 (2.72E-02) $+$ \\\midrule
\multirow{3}{*}{WFG8} & 3 & \textbf{9.314E-02 }(9.12E-04) $-$  & 9.598E-02 (1.22E-03) $-$  & 1.213E-01 (3.36E-03) $-$  & 1.070E-01 (2.16E-03) $-$ \\
 & 5 & 4.112E-01 (1.14E-02) $+$  & 3.884E-01 (1.19E-02) $+$  & \textbf{3.808E-01 }(7.26E-03) $+$  & 3.925E-01 (1.33E-02) $+$ \\
 & 7 & 5.743E-01 (1.09E-02) $+$  & 5.587E-01 (1.56E-02) $+$  & \textbf{5.564E-01 }(1.13E-02) $+$  & 5.570E-01 (1.22E-02) $+$ \\\midrule
\multirow{3}{*}{WFG9} & 3 & \textbf{5.993E-02 }(1.32E-02) $+$  & 8.122E-02 (2.54E-02) $-$  & 8.912E-02 (1.83E-02) $-$  & 8.652E-02 (2.15E-02) $-$ \\
 & 5 & \textbf{3.246E-01 }(1.54E-02) $+$  & 3.300E-01 (1.47E-02) $+$  & 3.325E-01 (1.63E-02) $+$  & 3.389E-01 (1.18E-02) $+$ \\
 & 7 & 5.179E-01 (2.68E-02) $+$  & \textbf{5.001E-01 }(2.59E-02) $+$  & 5.252E-01 (2.03E-02) $+$  & 5.472E-01 (2.12E-02) $\approx$ \\\midrule
 
\multicolumn{2}{c|}{Average rank}  & 2.63 & 1.88 & 2.69 & 2.80\\\midrule  
\multicolumn{2}{c|}{$+$/$-$/$\approx$}  & 13/10/1  & 12/10/2  & 12/12/0  & 11/12/1 \\

\bottomrule
\end{tabular}}
\label{table:5 op}
\end{table}

Then, we examine the performance of MOEA/D, MOEA/D-OP2, MOEA/D-FRRMAB and MA-DAC on different problems. The operator pool of FRRMAB is just the four types of DE operators. The results in Table~\ref{table:6 frrmab} show that MOEA/D-FRRMAB is better than MOEA/D and MOEA/D-OP2, disclosing the effectiveness of adjusting the type of reproduction operators. We can also observe that the proposed MA-DAC clearly performs the best.

\begin{table}[htbp]\scriptsize
\centering
\caption{IGD values obtained by MOEA/D, MOEA/D-OP2, MOEA/D-FRRMAB and MA-DAC on different problems. Each result consists of the mean and standard deviation of 30 runs. The best mean value on each problem is highlighted in \textbf{bold}. The symbols `$+$', `$-$' and `$\approx$' indicate that the result is significantly superior to, inferior to, and almost equivalent to MA-DAC, respectively, according to the Wilcoxon rank-sum test with significance level 0.05.}\vspace{0.5em}
\resizebox{\linewidth}{!}{
\begin{tabular}{c c|c c c c c }
\toprule
     Problem     &   $M$  & MOEA/D  & MOEA/D-OP2 & MOEA/D-FRRMAB & MA-DAC \\\midrule
\multirow{3}{*}{DTLZ2} & 3 & 4.605E-02 (3.54E-04) $-$  & 4.691E-02 (3.97E-04) $-$  & 4.668E-02 (2.50E-04) $-$  & \textbf{3.807E-02 }(5.05E-04)  \\
 & 5 & 3.006E-01 (1.55E-03) $-$  & 3.012E-01 (1.51E-03) $-$  & 3.031E-01 (1.29E-03) $-$  & \textbf{2.442E-01 }(1.26E-02)  \\
 & 7 & 4.455E-01 (1.41E-02) $-$  & 4.551E-01 (4.43E-03) $-$  & 4.724E-01 (7.80E-03) $-$  & \textbf{3.944E-01 }(1.17E-02)  \\\midrule

\multirow{3}{*}{WFG4} & 3 & 5.761E-02 (5.41E-04) $-$  & 7.293E-02 (1.43E-03) $-$  & 7.097E-02 (1.63E-03) $-$  & \textbf{5.200E-02 }(1.19E-03)  \\
 & 5 & 3.442E-01 (1.21E-02) $-$  & 2.761E-01 (6.39E-03) $-$  & 2.799E-01 (9.44E-03) $-$  & \textbf{1.868E-01 }(2.81E-03)  \\
 & 7 & 4.529E-01 (1.79E-02) $-$  & 3.711E-01 (9.79E-03) $-$  & 3.778E-01 (1.01E-02) $-$  & \textbf{3.033E-01 }(3.66E-03)  \\\midrule

\multirow{3}{*}{WFG6} & 3 & 6.938E-02 (5.50E-03) $-$  & 6.714E-02 (1.59E-02) $-$  & 6.266E-02 (8.47E-03) $-$  & \textbf{4.831E-02 }(8.95E-03)  \\
 & 5 & 3.518E-01 (2.82E-03) $-$  & 3.285E-01 (2.33E-02) $-$  & 3.272E-01 (1.61E-02) $-$  & \textbf{1.942E-01 }(6.90E-03)  \\
 & 7 & 4.869E-01 (3.03E-02) $-$  & 4.797E-01 (3.04E-02) $-$  & 4.417E-01 (3.29E-02) $-$  & \textbf{3.112E-01 }(4.93E-03)  \\\midrule

\multicolumn{2}{c|}{Train: average rank} & 3.11 & 3.00 & 2.89 & 1.00\\\midrule 
\multicolumn{2}{c|}{Train: $+$/$-$/$\approx$} & 0/9/0  & 0/9/0  & 0/9/0  &  \\\midrule

\multirow{3}{*}{DTLZ4} & 3 & 6.231E-02 (8.85E-02) $\approx$  & 6.226E-02 (4.05E-03) $-$  & \textbf{5.782E-02 }(3.48E-03) $-$  & 6.700E-02 (6.14E-02)  \\
 & 5 & 3.133E-01 (4.45E-02) $\approx$  & 3.413E-01 (1.48E-02) $-$  & 3.373E-01 (1.70E-02) $-$  & \textbf{2.995E-01 }(2.10E-02)  \\
 & 7 & 4.374E-01 (2.57E-02) $-$  & 4.519E-01 (1.15E-02) $-$  & 4.681E-01 (1.87E-02) $-$  & \textbf{4.182E-01 }(1.21E-02)  \\\midrule

\multirow{3}{*}{WFG5} & 3 & 6.327E-02 (1.10E-03) $-$  & 6.177E-02 (8.01E-04) $-$  & 6.120E-02 (7.38E-04) $-$  & \textbf{4.730E-02 }(7.89E-04)  \\
 & 5 & 3.350E-01 (9.77E-03) $-$  & 3.052E-01 (7.19E-03) $-$  & 3.033E-01 (8.69E-03) $-$  & \textbf{1.811E-01 }(3.02E-03)  \\
 & 7 & 4.101E-01 (2.08E-02) $-$  & 4.988E-01 (1.04E-02) $-$  & 5.045E-01 (9.70E-03) $-$  & \textbf{3.206E-01 }(8.04E-03)  \\\midrule

\multirow{3}{*}{WFG7} & 3 & 5.811E-02 (6.31E-04) $-$  & 6.033E-02 (8.84E-04) $-$  & 5.976E-02 (7.44E-04) $-$  & \textbf{4.066E-02 }(5.31E-04)  \\
 & 5 & 3.572E-01 (5.47E-03) $-$  & 2.941E-01 (9.66E-03) $-$  & 3.042E-01 (1.52E-02) $-$  & \textbf{1.858E-01 }(2.12E-03)  \\
 & 7 & 5.236E-01 (2.19E-02) $-$  & 4.739E-01 (2.51E-02) $-$  & 4.762E-01 (2.74E-02) $-$  & \textbf{3.258E-01 }(1.25E-02)  \\\midrule
\multirow{3}{*}{WFG8} & 3 & 8.646E-02 (3.44E-03) $-$  & 9.598E-02 (1.22E-03) $-$  & 9.536E-02 (1.14E-03) $-$  & \textbf{7.901E-02 }(1.19E-03)  \\
 & 5 & 4.258E-01 (8.42E-03) $-$  & 3.884E-01 (1.19E-02) $-$  & 3.917E-01 (9.00E-03) $-$  & \textbf{2.479E-01 }(7.20E-03)  \\
 & 7 & 5.816E-01 (1.30E-02) $-$  & 5.587E-01 (1.56E-02) $-$  & 5.570E-01 (1.60E-02) $-$  & \textbf{4.127E-01 }(5.93E-03)  \\\midrule
\multirow{3}{*}{WFG9} & 3 & 5.817E-02 (1.24E-03) $-$  & 8.122E-02 (2.54E-02) $-$  & 6.445E-02 (1.72E-02) $-$  & \textbf{4.159E-02 }(6.10E-04)  \\
 & 5 & 3.633E-01 (1.20E-02) $-$  & 3.300E-01 (1.47E-02) $-$  & 3.312E-01 (1.70E-02) $-$  & \textbf{1.832E-01 }(7.10E-03)  \\
 & 7 & 5.538E-01 (2.63E-02) $-$  & 5.001E-01 (2.59E-02) $-$  & 5.145E-01 (2.82E-02) $-$  & \textbf{3.278E-01 }(7.21E-03)  \\\midrule

\multicolumn{2}{c|}{Test: average rank} & 3.13 & 2.87 & 2.80 & 1.20\\\midrule 
\multicolumn{2}{c|}{Test: $+$/$-$/$\approx$} & 0/13/2  & 0/15/0  & 0/15/0  & \\

\bottomrule
\end{tabular}}
\label{table:6 frrmab}
\end{table}

\begin{table}[htbp]\scriptsize
\centering
\caption{IGD values obtained by MOEA/D, MOEA/D-AWA, MOEA/D-OP2-AWA and MA-DAC on different problems. Each result consists of the mean and standard deviation of 30 runs. The best mean value on each problem is highlighted in \textbf{bold}. The symbols `$+$', `$-$' and `$\approx$' indicate that the result is significantly superior to, inferior to, and almost equivalent to MA-DAC, respectively, according to the Wilcoxon rank-sum test with significance level 0.05.}\vspace{0.5em}
\resizebox{\linewidth}{!}{
\begin{tabular}{c c|c c c c c }
\toprule
     Problem     &   $M$  & MOEA/D  & MOEA/D-AWA & MOEA/D-OP2-AWA & MA-DAC \\\midrule
\multirow{3}{*}{DTLZ2} & 3 & 4.605E-02 (3.54E-04) $-$  & 4.596E-02 (3.54E-04) $-$  & 4.670E-02 (3.30E-04) $-$  & \textbf{3.807E-02 }(5.05E-04)  \\
 & 5 & 3.006E-01 (1.55E-03) $-$  & 2.900E-01 (2.73E-03) $-$  & 2.764E-01 (3.40E-03) $-$  & \textbf{2.442E-01 }(1.26E-02)  \\
 & 7 & 4.455E-01 (1.41E-02) $-$  & 4.167E-01 (2.37E-02) $-$  & 4.436E-01 (8.67E-03) $-$  & \textbf{3.944E-01 }(1.17E-02)  \\\midrule

\multirow{3}{*}{WFG4} & 3 & 5.761E-02 (5.41E-04) $-$  & 5.748E-02 (7.11E-04) $-$  & 7.280E-02 (1.33E-03) $-$  & \textbf{5.200E-02 }(1.19E-03)  \\
 & 5 & 3.442E-01 (1.21E-02) $-$  & 3.168E-01 (5.37E-03) $-$  & 2.648E-01 (8.15E-03) $-$  & \textbf{1.868E-01 }(2.81E-03)  \\
 & 7 & 4.529E-01 (1.79E-02) $-$  & 4.285E-01 (1.55E-02) $-$  & 3.676E-01 (1.06E-02) $-$  & \textbf{3.033E-01 }(3.66E-03)  \\\midrule

\multirow{3}{*}{WFG6} & 3 & 6.938E-02 (5.50E-03) $-$  & 6.846E-02 (4.70E-03) $-$  & 6.078E-02 (1.16E-03) $-$  & \textbf{4.831E-02 }(8.95E-03)  \\
 & 5 & 3.518E-01 (2.82E-03) $-$  & 3.190E-01 (3.93E-03) $-$  & 3.143E-01 (2.52E-02) $-$  & \textbf{1.942E-01 }(6.90E-03)  \\
 & 7 & 4.869E-01 (3.03E-02) $-$  & 4.727E-01 (3.05E-02) $-$  & 4.770E-01 (3.24E-02) $-$  & \textbf{3.112E-01 }(4.93E-03)  \\\midrule
 
 \multicolumn{2}{c|}{Train: average rank}  & 3.78 & 2.56 & 2.67 & 1.00\\\midrule 
 \multicolumn{2}{c|}{Train: $+$/$-$/$\approx$}  & 0/9/0  & 0/9/0  & 0/9/0  &  \\\midrule  
 
 \multirow{3}{*}{DTLZ4} & 3 & 6.231E-02 (8.85E-02) $\approx$  & \textbf{4.597E-02 }(3.66E-04) $\approx$  & 6.219E-02 (3.90E-03) $-$  & 6.700E-02 (6.14E-02)  \\
 & 5 & 3.133E-01 (4.45E-02) $\approx$  & \textbf{2.816E-01 }(3.24E-03) $+$  & 3.283E-01 (1.08E-02) $-$  & 2.995E-01 (2.10E-02)  \\
 & 7 & 4.374E-01 (2.57E-02) $-$  & \textbf{3.696E-01 }(1.32E-02) $+$  & 4.437E-01 (9.46E-03) $-$  & 4.182E-01 (1.21E-02)  \\\midrule
 
 \multirow{3}{*}{WFG5} & 3 & 6.327E-02 (1.10E-03) $-$  & 6.376E-02 (9.85E-04) $-$  & 6.168E-02 (4.61E-04) $-$  & \textbf{4.730E-02 }(7.89E-04)  \\
 & 5 & 3.350E-01 (9.77E-03) $-$  & 3.173E-01 (5.33E-03) $-$  & 3.024E-01 (6.02E-03) $-$  & \textbf{1.811E-01 }(3.02E-03)  \\
 & 7 & 4.101E-01 (2.08E-02) $-$  & 4.095E-01 (1.94E-02) $-$  & 4.865E-01 (1.28E-02) $-$  & \textbf{3.206E-01 }(8.04E-03)  \\\midrule
 
\multirow{3}{*}{WFG7} & 3 & 5.811E-02 (6.31E-04) $-$  & 5.837E-02 (6.25E-04) $-$  & 6.017E-02 (6.74E-04) $-$  & \textbf{4.066E-02 }(5.31E-04)  \\
 & 5 & 3.572E-01 (5.47E-03) $-$  & 3.227E-01 (4.19E-03) $-$  & 2.885E-01 (1.25E-02) $-$  & \textbf{1.858E-01 }(2.12E-03)  \\
 & 7 & 5.236E-01 (2.19E-02) $-$  & 5.004E-01 (3.80E-02) $-$  & 4.560E-01 (2.56E-02) $-$  & \textbf{3.258E-01 }(1.25E-02)  \\\midrule
\multirow{3}{*}{WFG8} & 3 & 8.646E-02 (3.44E-03) $-$  & 8.742E-02 (6.36E-04) $-$  & 9.572E-02 (8.39E-04) $-$  & \textbf{7.901E-02 }(1.19E-03)  \\
 & 5 & 4.258E-01 (8.42E-03) $-$  & 4.216E-01 (1.18E-02) $-$  & 3.824E-01 (9.74E-03) $-$  & \textbf{2.479E-01 }(7.20E-03)  \\
 & 7 & 5.816E-01 (1.30E-02) $-$  & 5.790E-01 (1.06E-02) $-$  & 5.632E-01 (1.27E-02) $-$  & \textbf{4.127E-01 }(5.93E-03)  \\\midrule
\multirow{3}{*}{WFG9} & 3 & 5.817E-02 (1.24E-03) $-$  & 5.809E-02 (1.45E-03) $-$  & 6.470E-02 (1.75E-02) $-$  & \textbf{4.159E-02 }(6.10E-04)  \\
 & 5 & 3.633E-01 (1.20E-02) $-$  & 3.517E-01 (2.19E-02) $-$  & 3.024E-01 (1.36E-02) $-$  & \textbf{1.832E-01 }(7.10E-03)  \\
 & 7 & 5.538E-01 (2.63E-02) $-$  & 5.108E-01 (2.65E-02) $-$  & 4.861E-01 (2.78E-02) $-$  & \textbf{3.278E-01 }(7.21E-03)  \\\midrule

\multicolumn{2}{c|}{Test: average rank}  & 3.33 & 2.53 & 2.80 & 1.33 \\\midrule  
\multicolumn{2}{c|}{Test: $+$/$-$/$\approx$}  & 0/13/2  & 2/12/1  & 0/15/0  & \\
\bottomrule
\end{tabular}}
\label{table:7 awa}
\end{table}

\subsection{Analysis of the adaptive weights}
\label{analysis of adaptive weights}
In this subsection, we compare MA-DAC with MOEA/D-AWA~\citep{AWA}, which dynamically adjusts the weights of MOEA/D based on predefined heuristic intervals. The concrete way of adjusting the weights of MOEA/D-AWA and MA-DAC are the same, as described in Appendix~\ref{app:action}. Table~\ref{table:7 awa} shows the results, where MOEA/D-OP2-AWA refers to MOEA/D-AWA using the DE operator OP2 (which has been shown to be the best among the four investigated DE operators in the last subsection) instead of the SBX operator. We can observe that MA-DAC performs the best in all problems except DTLZ4, where MOEA/D-AWA is better. Note that DTLZ4 is not used for training MA-DAC, and the worse performance than MOEA/D-AWA on this problem also implies that MA-DAC can be further improved in the future.

\subsection{IGD values during the optimization process}\label{app: IGD values during the optimization process}

\begin{figure*}[htbp!]
    \centering
    \vspace{-1em}
    \centering
    \subfigure[DTLZ2\_3]{
	    \includegraphics[width=0.23\textwidth]{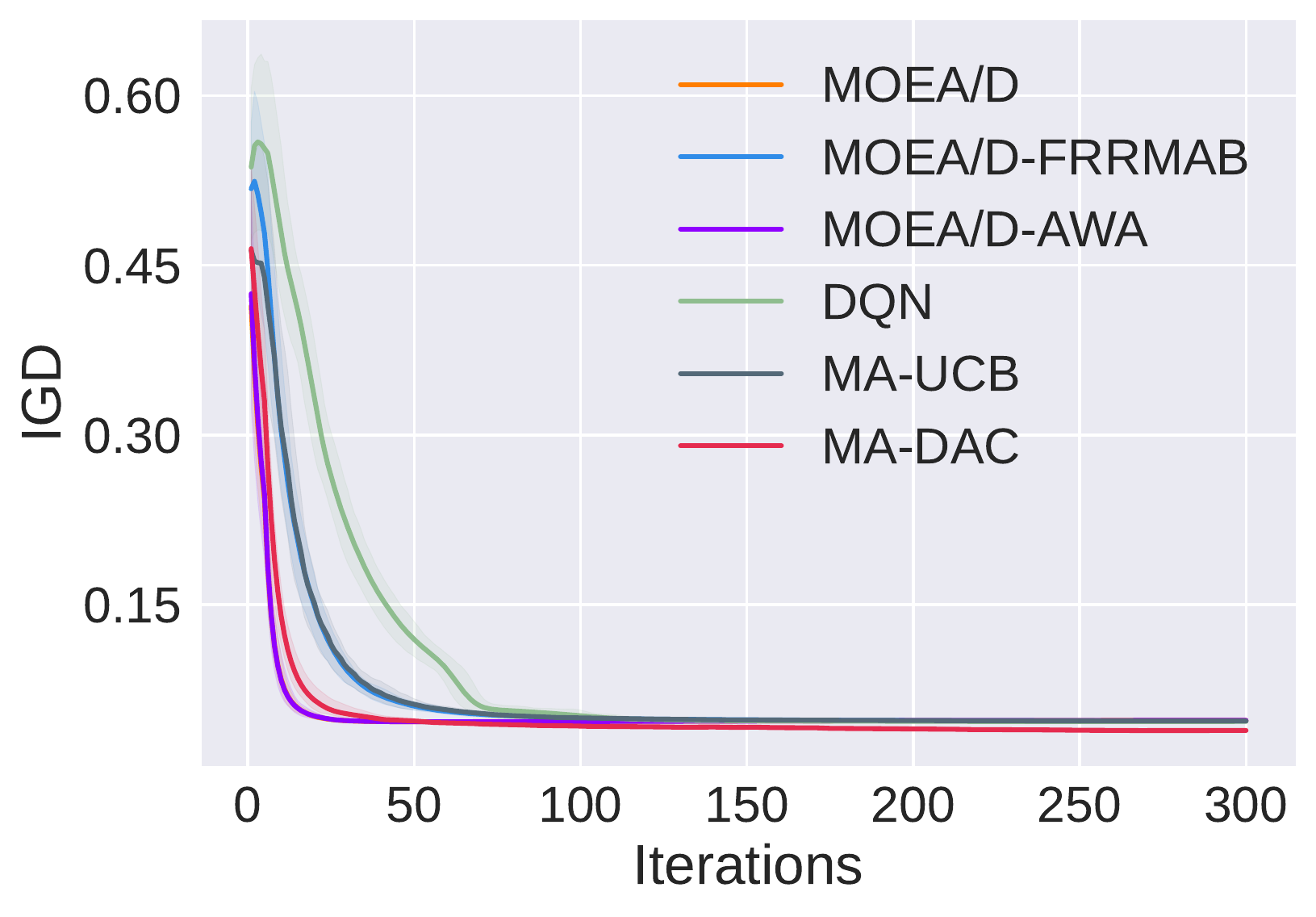}
    }
    \subfigure[DTLZ4\_3]{
        \includegraphics[width=0.23\textwidth]{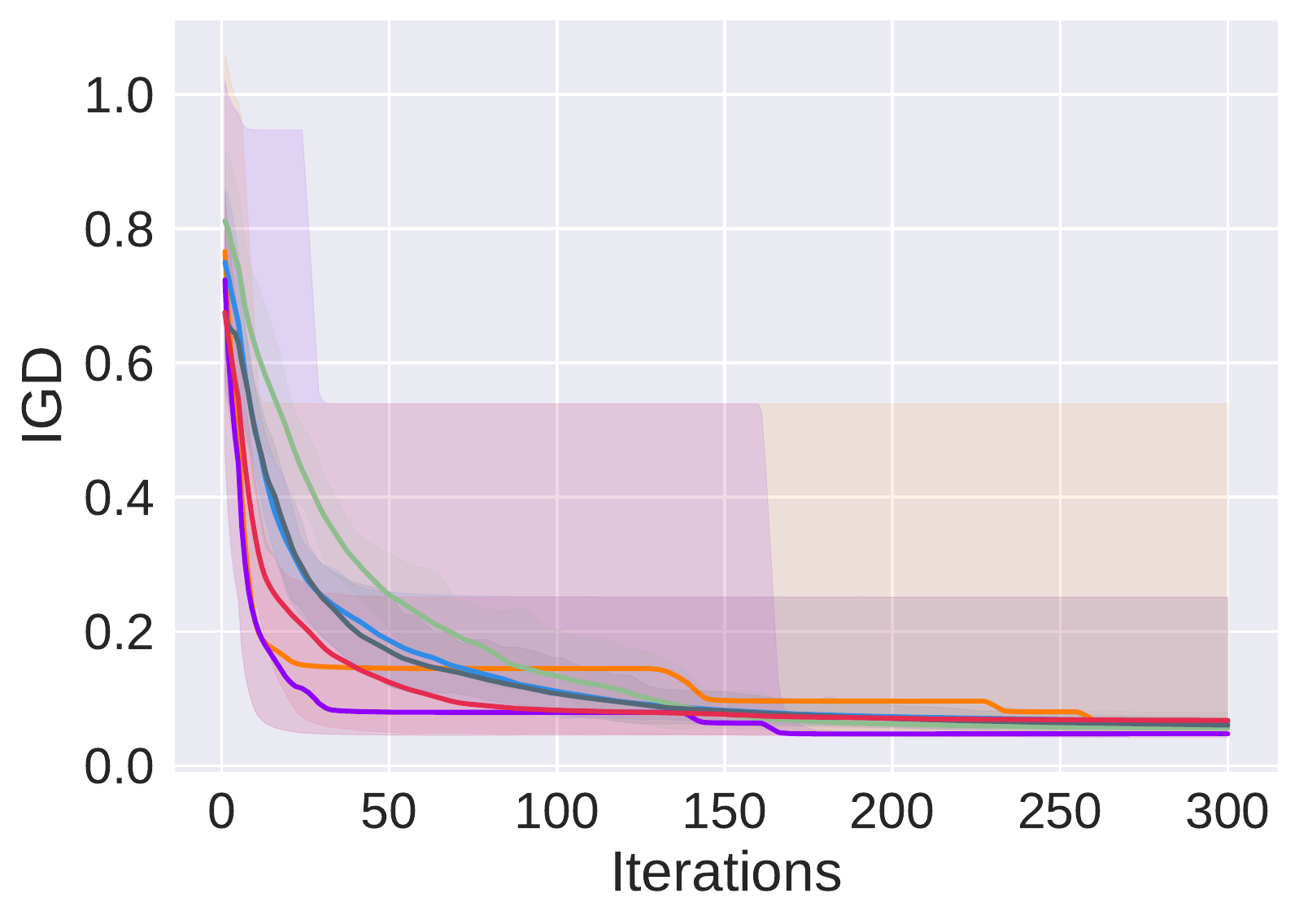}
    }
    \subfigure[WFG4\_3]{
        \includegraphics[width=0.23\textwidth]{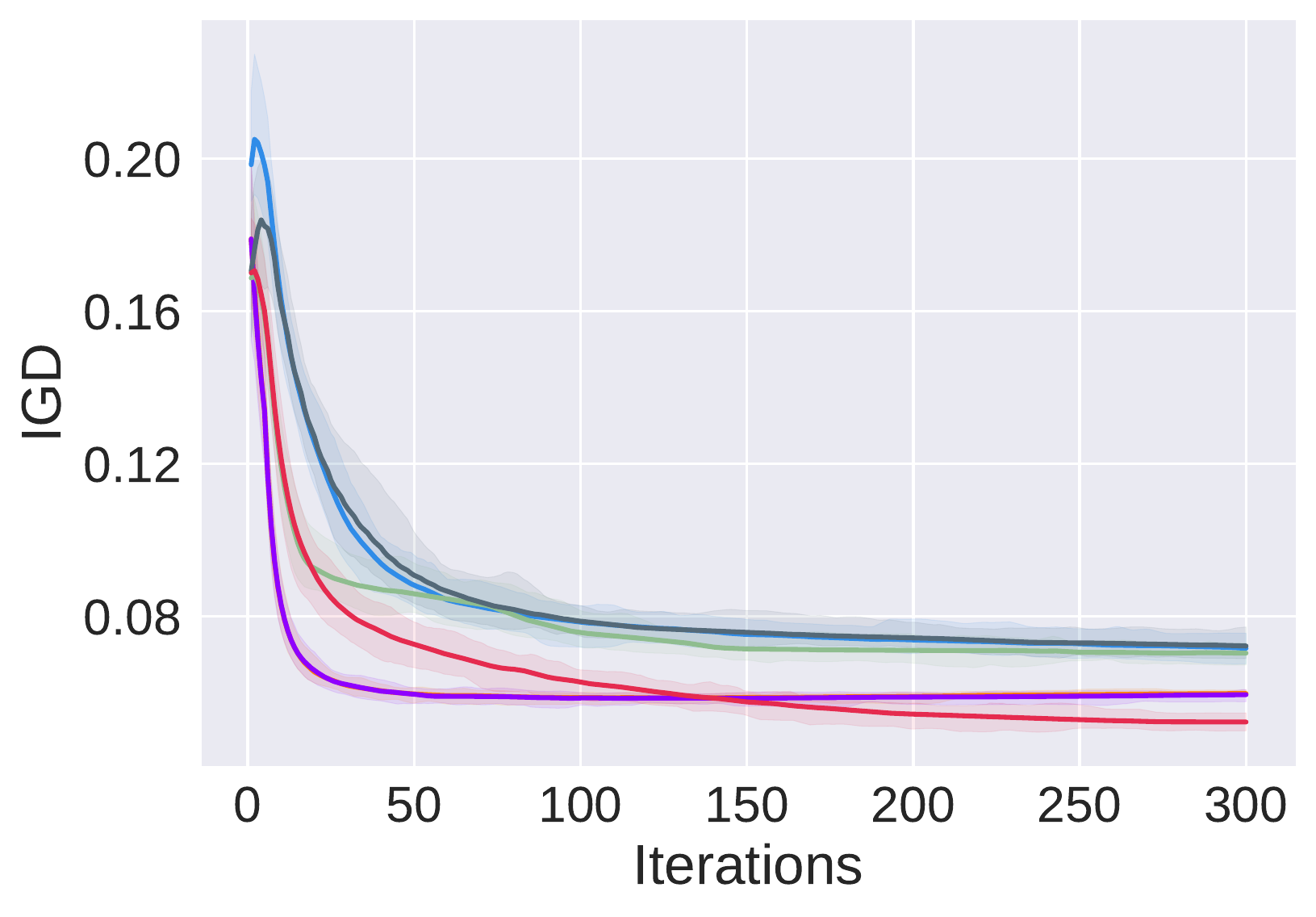}
    }
    \subfigure[WFG5\_3]{
        \includegraphics[width=0.23\textwidth]{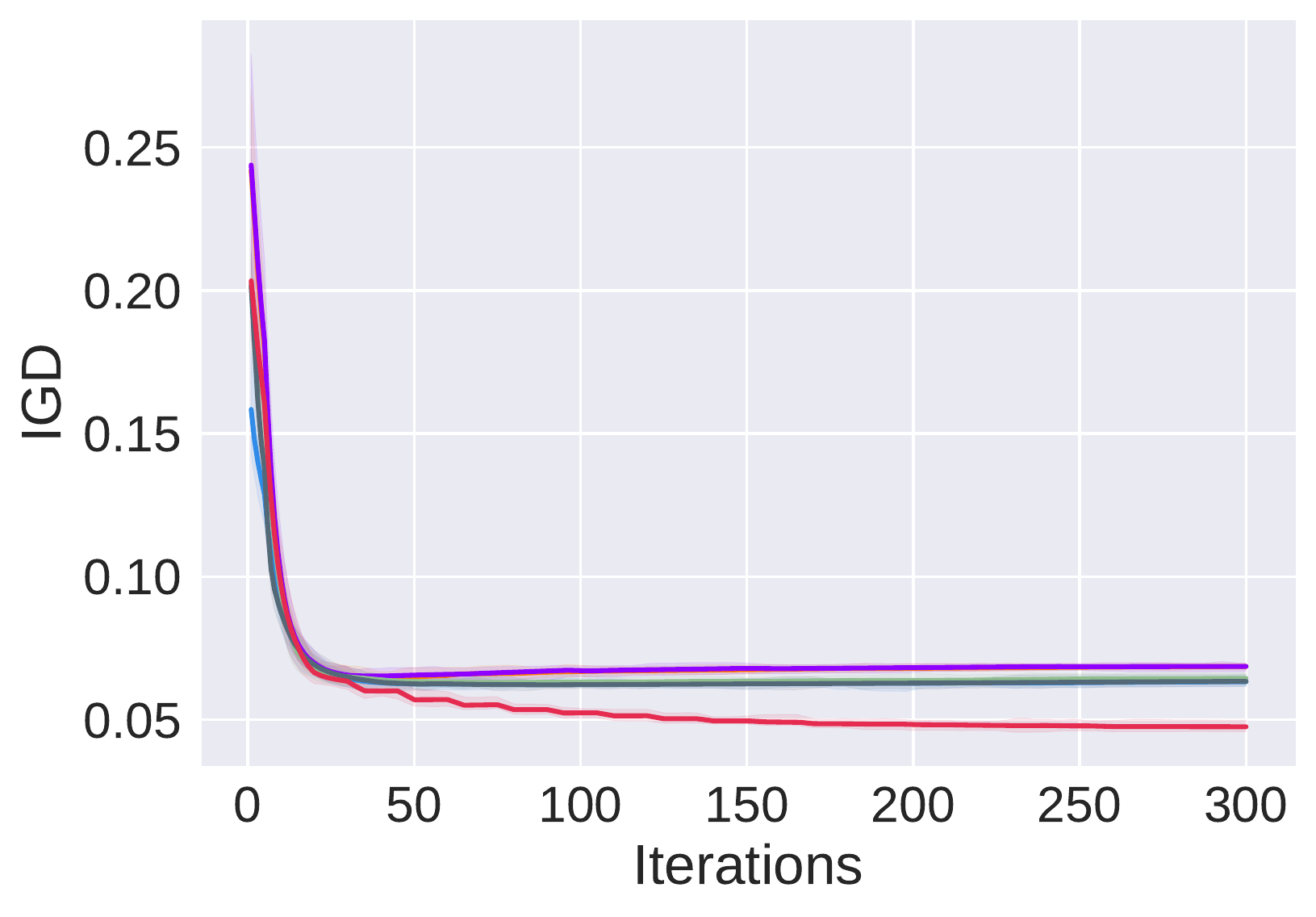}
    }
    \vspace{-1em}\\
    \subfigure[WFG6\_3]{
        \includegraphics[width=0.23\textwidth]{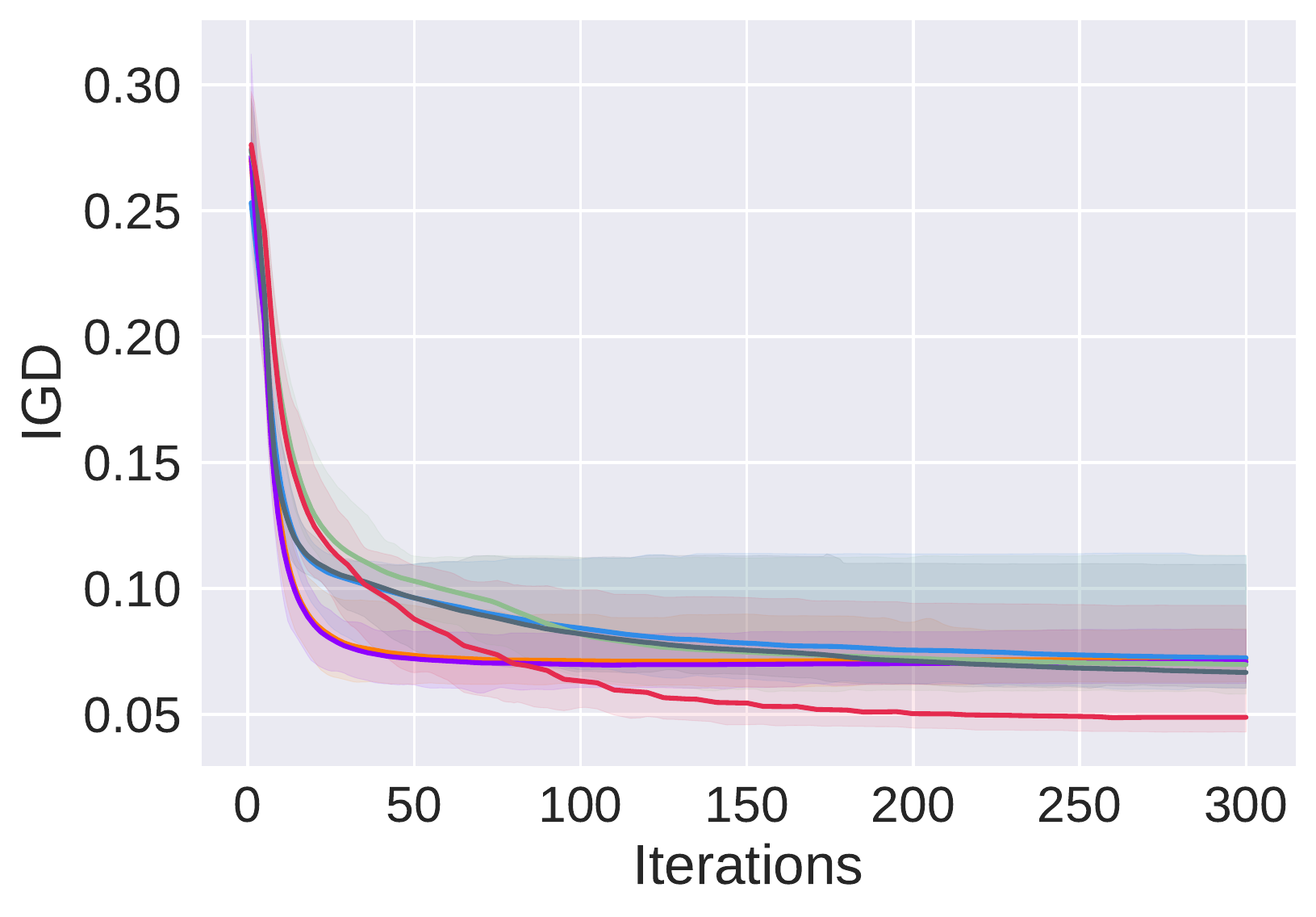}
    }
    \subfigure[WFG7\_3]{
        \includegraphics[width=0.23\textwidth]{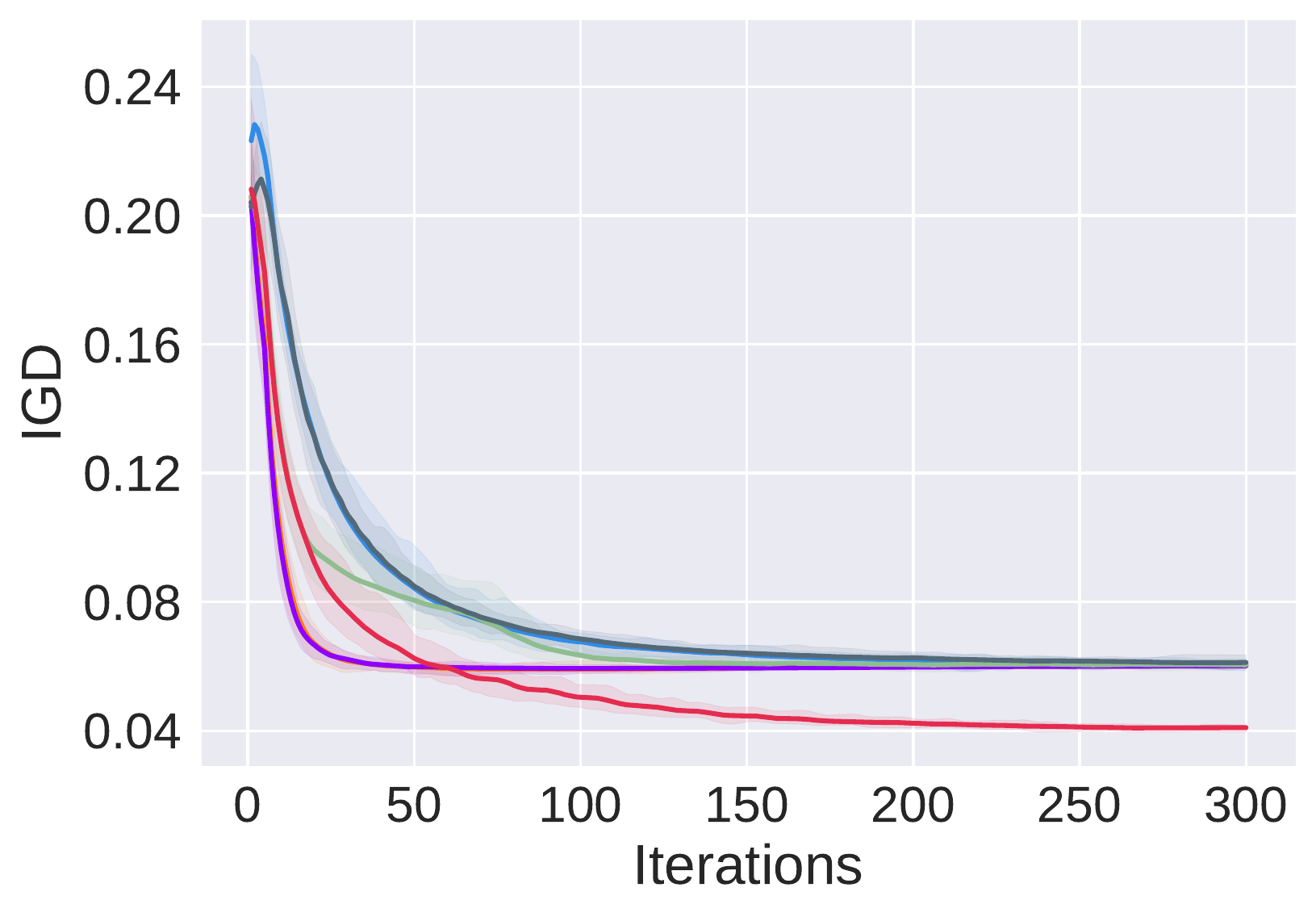}
    }
    \subfigure[WFG8\_3]{
        \includegraphics[width=0.23\textwidth]{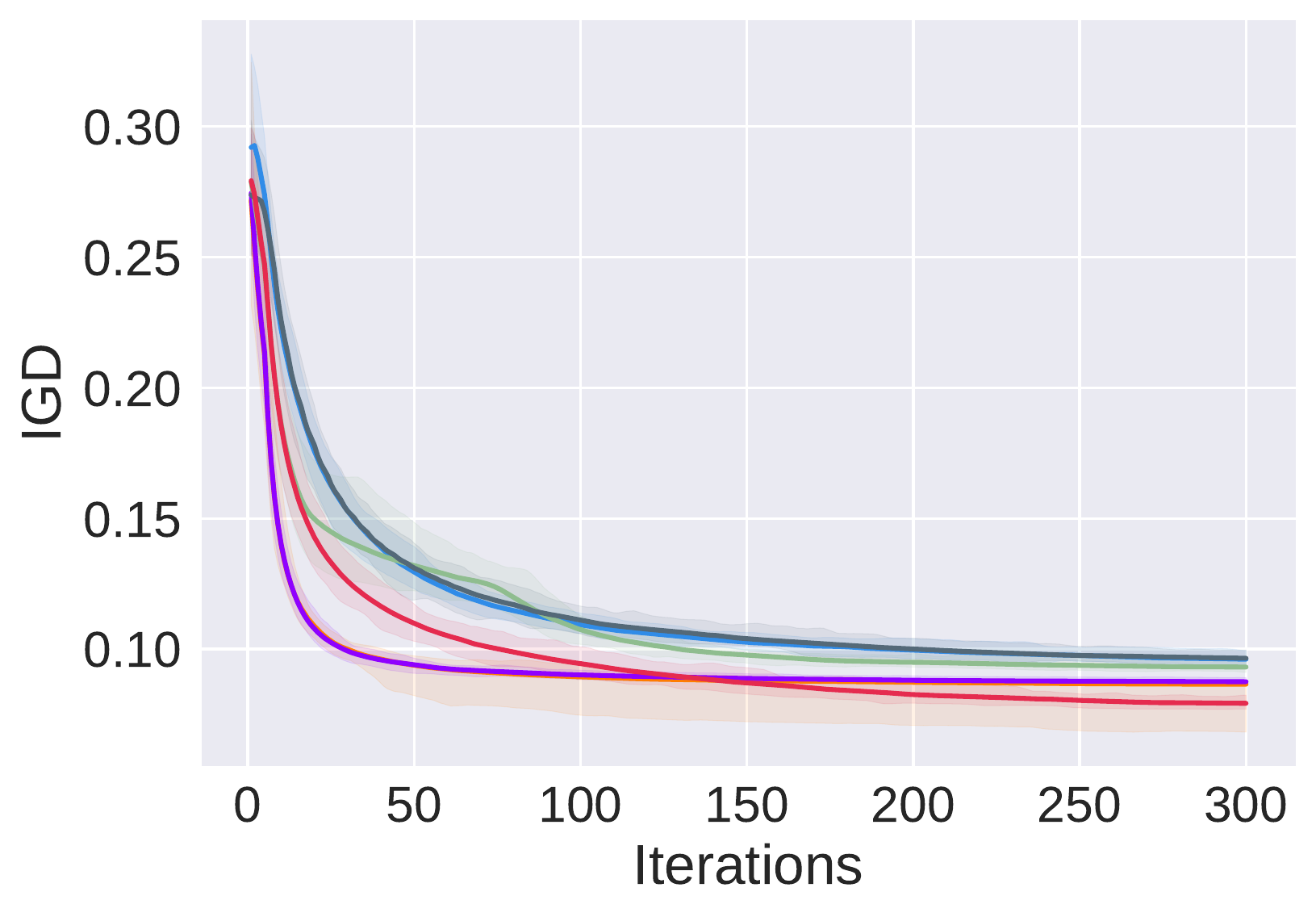}
    }
    \subfigure[WFG9\_3]{
        \includegraphics[width=0.23\textwidth]{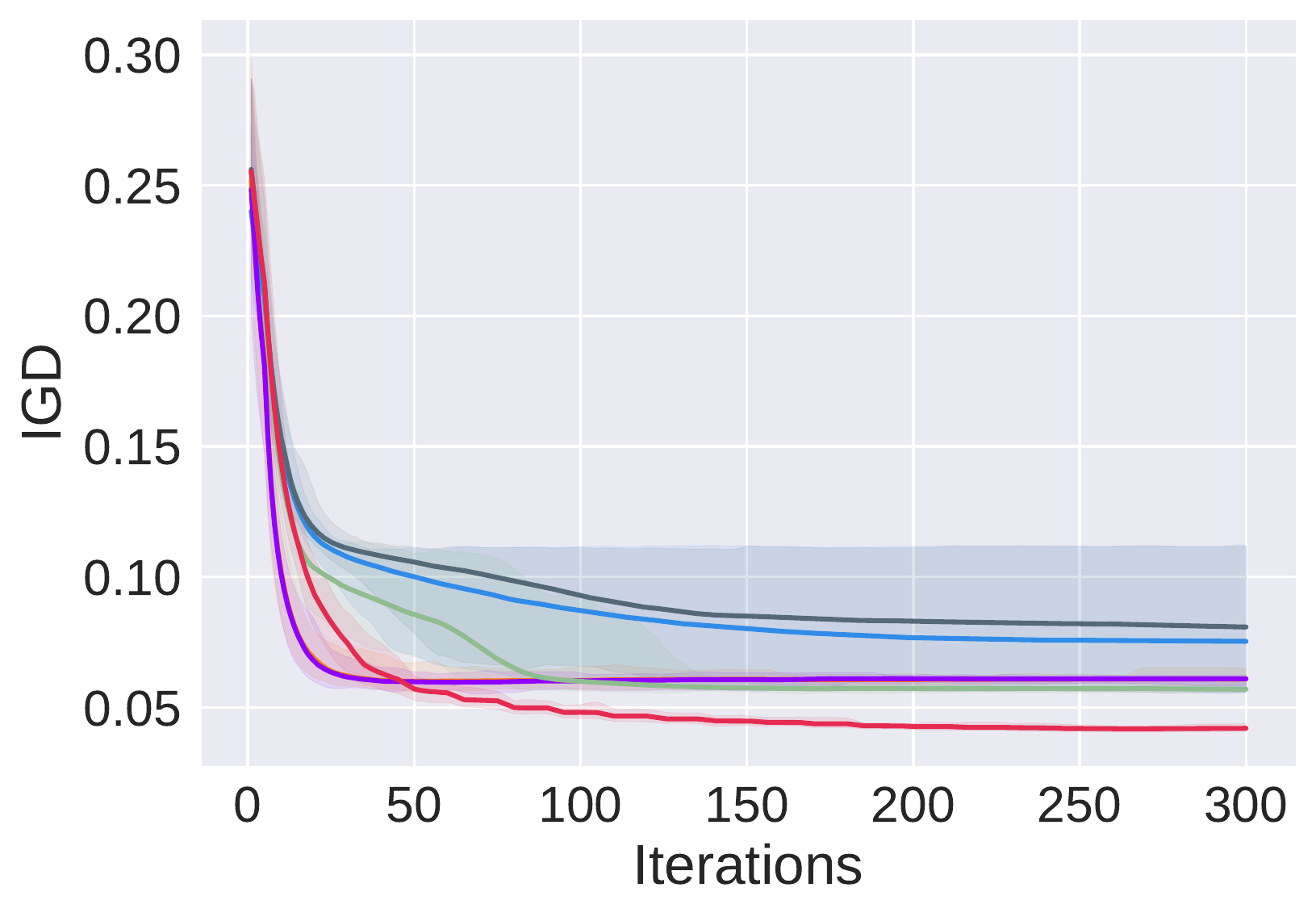}
    }
    \caption{Curves of IGD value obtained by the compared methods on the 3-objective problems.}
    \label{fig:3objs}
\end{figure*}

\begin{figure*}[htbp!]
    \centering
    \centering
    \subfigure[DTLZ2\_5]{
	    \includegraphics[width=0.23\textwidth]{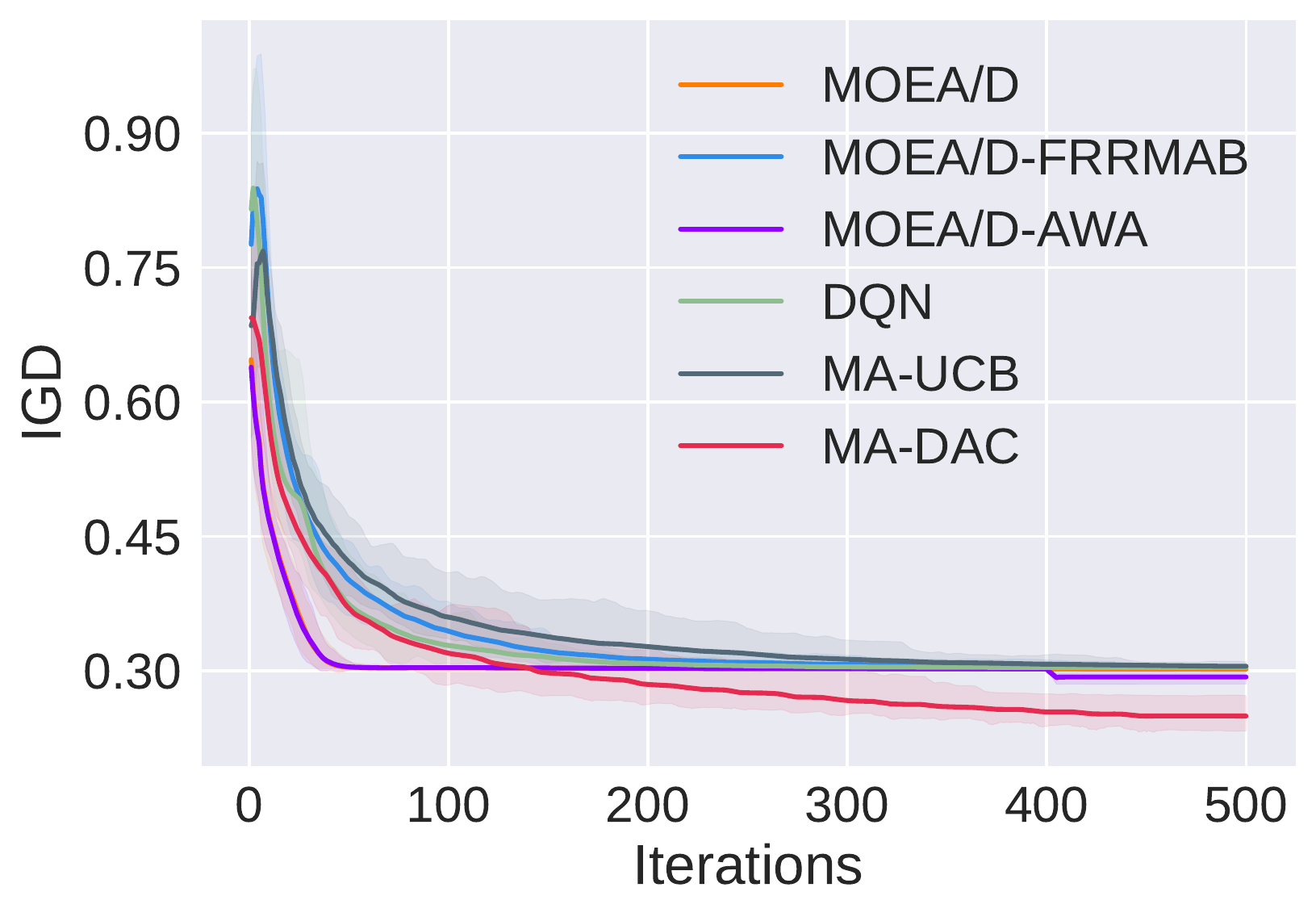}
    }
    \subfigure[DTLZ4\_5]{
        \includegraphics[width=0.23\textwidth]{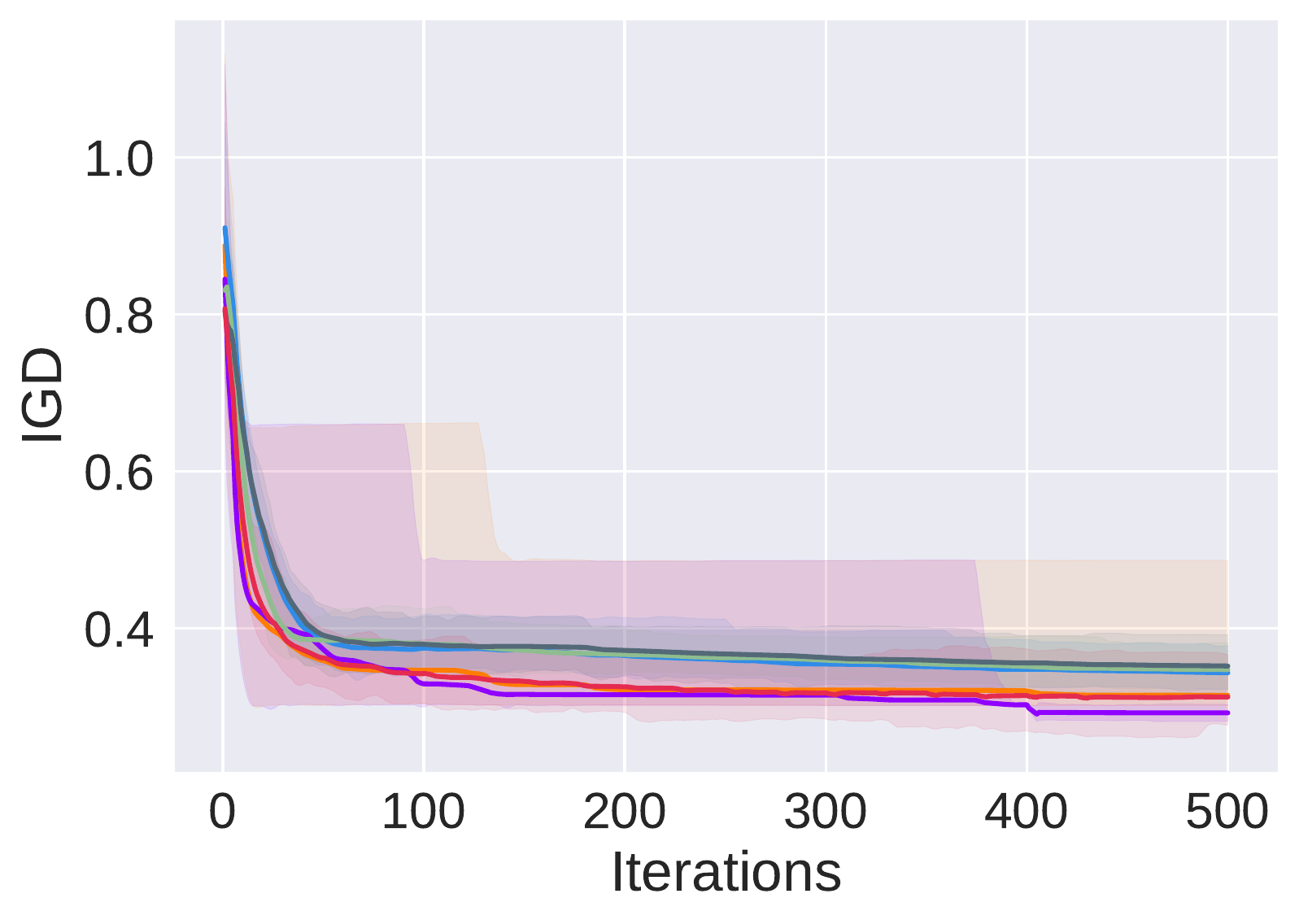}
    }
    \subfigure[WFG4\_5]{
        \includegraphics[width=0.23\textwidth]{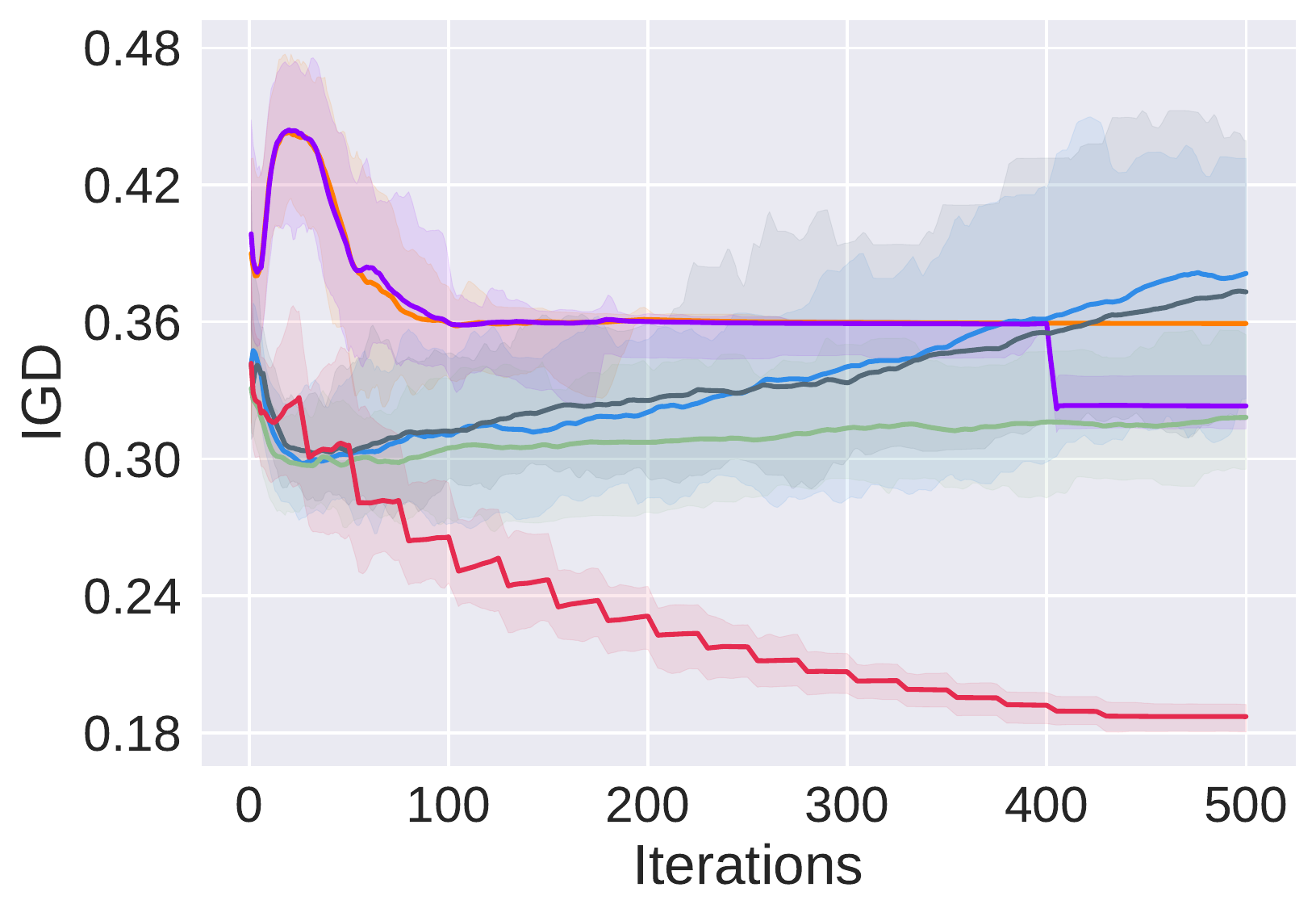}
    }
    \subfigure[WFG5\_5]{
        \includegraphics[width=0.23\textwidth]{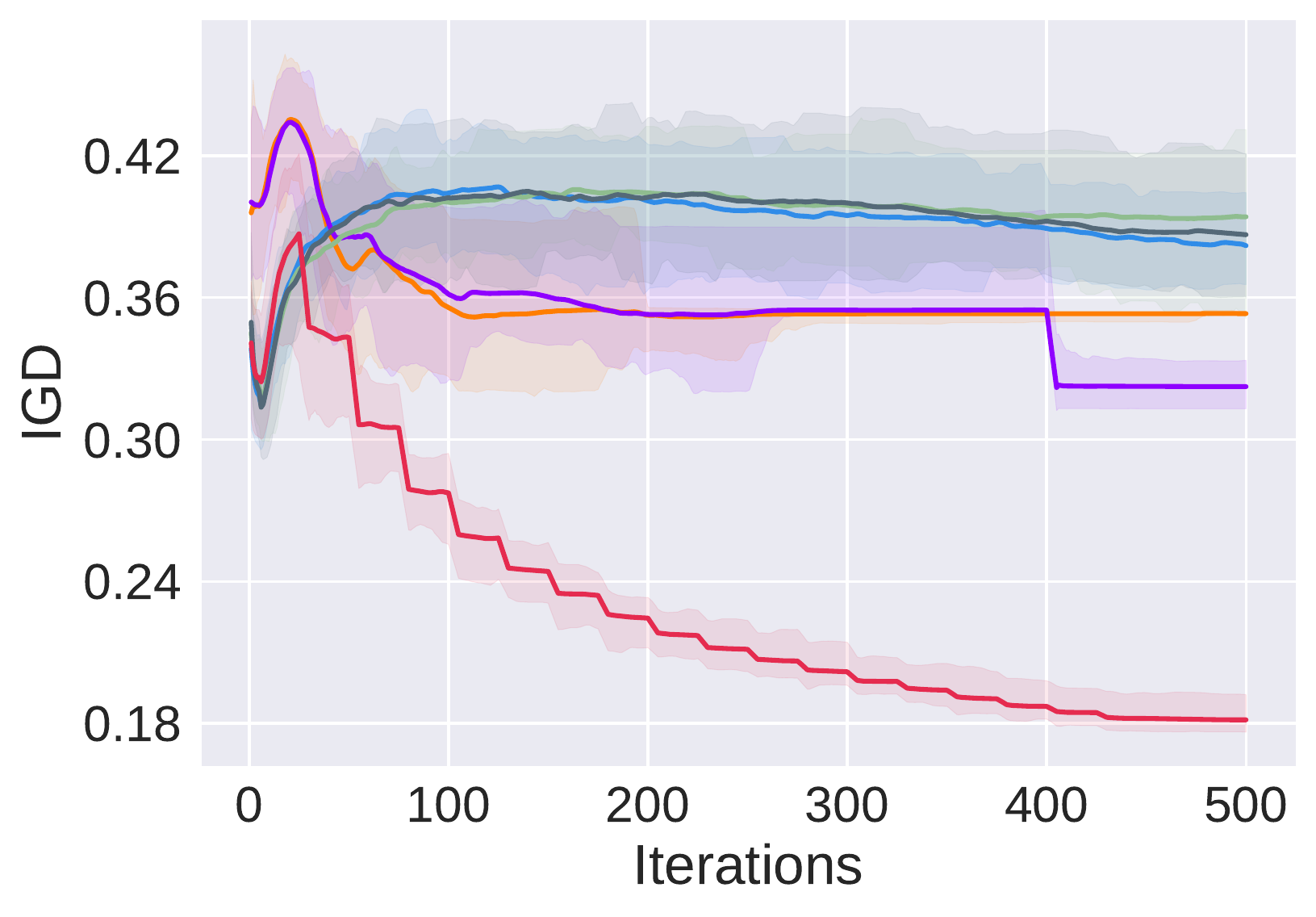}
    }
\\
    \subfigure[WFG6\_5]{
        \includegraphics[width=0.23\textwidth]{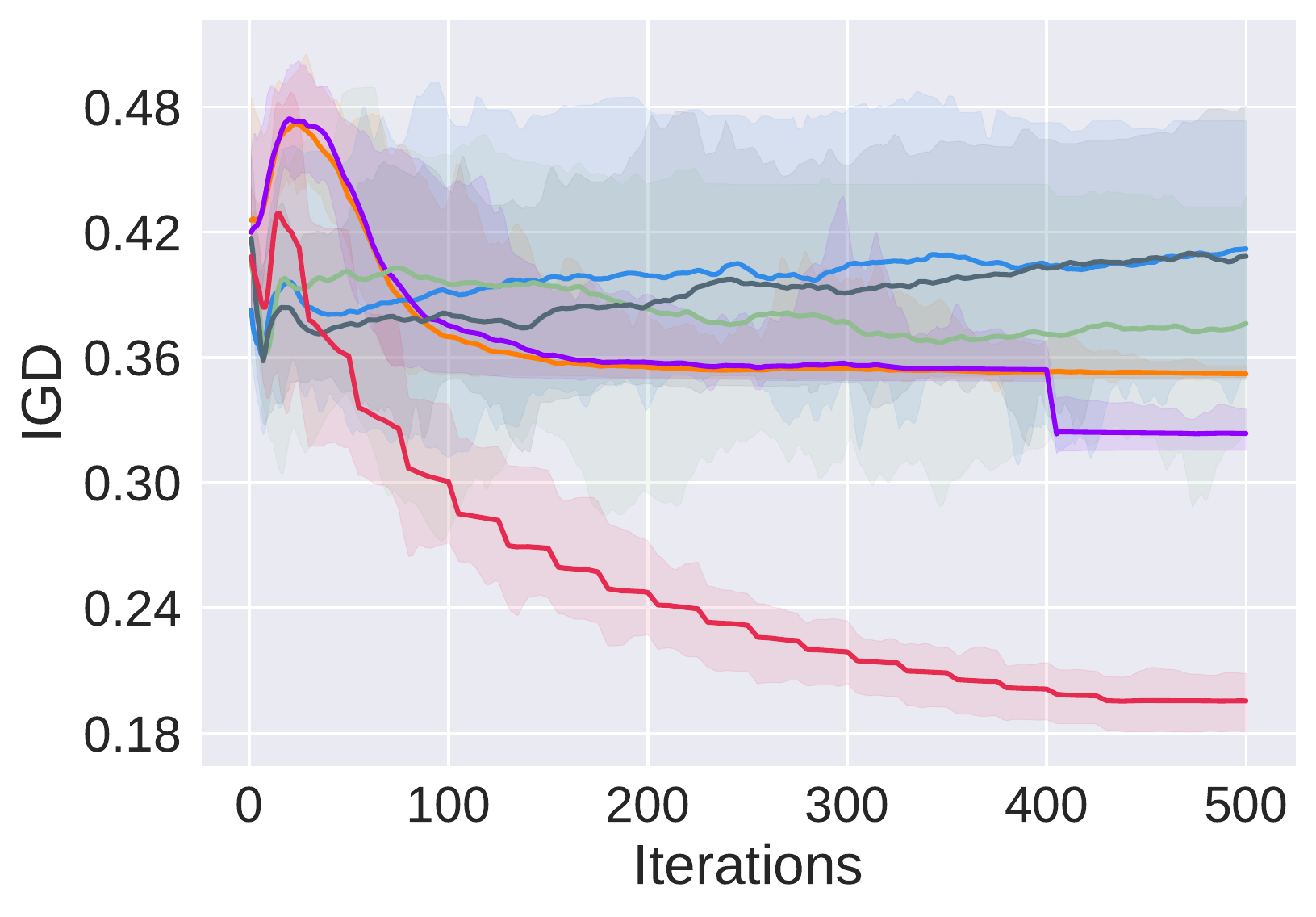}
    }
    \subfigure[WFG7\_5]{
        \includegraphics[width=0.23\textwidth]{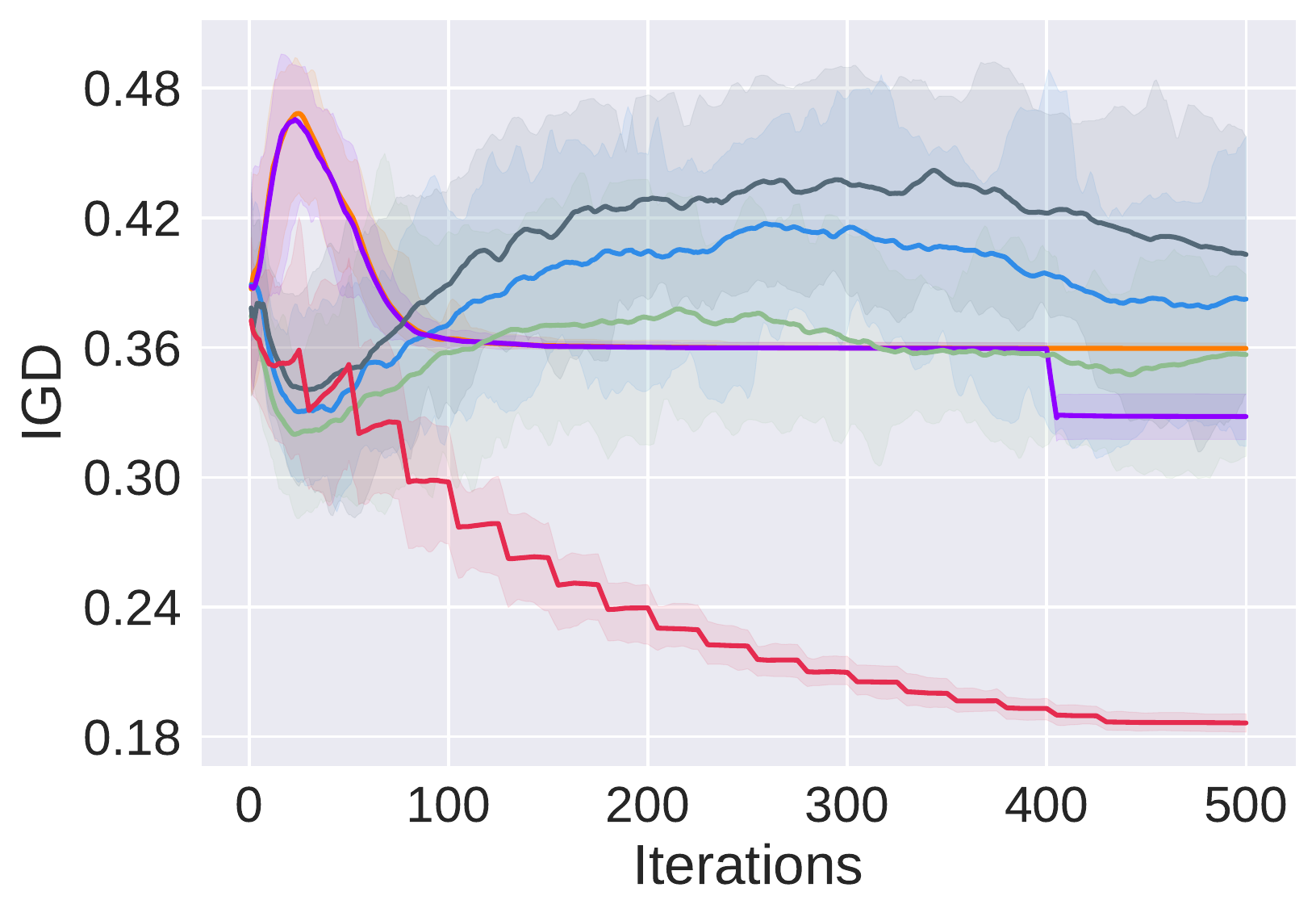}
    }
    \subfigure[WFG8\_5]{
        \includegraphics[width=0.23\textwidth]{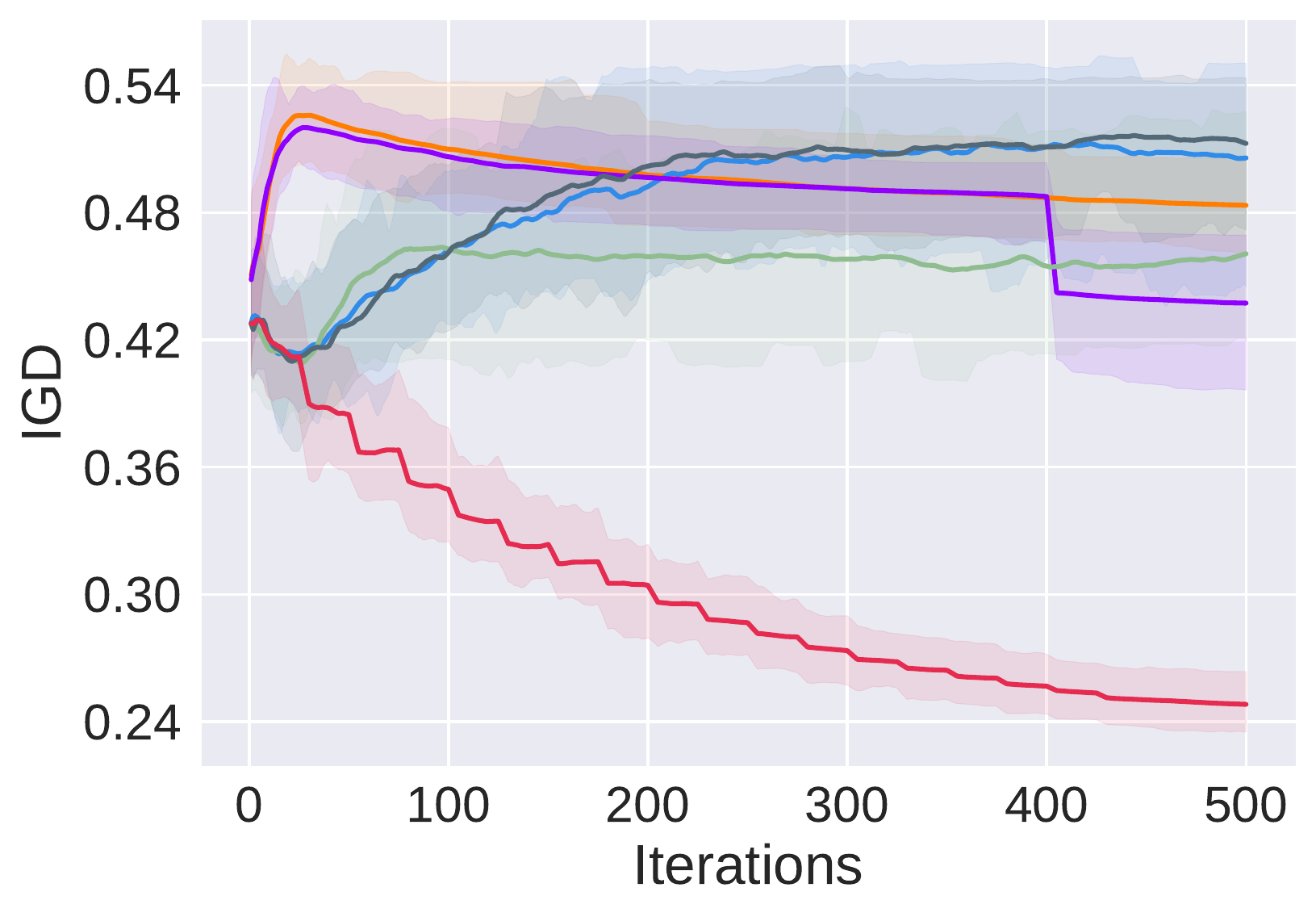}
    }
    \subfigure[WFG9\_5]{
        \includegraphics[width=0.23\textwidth]{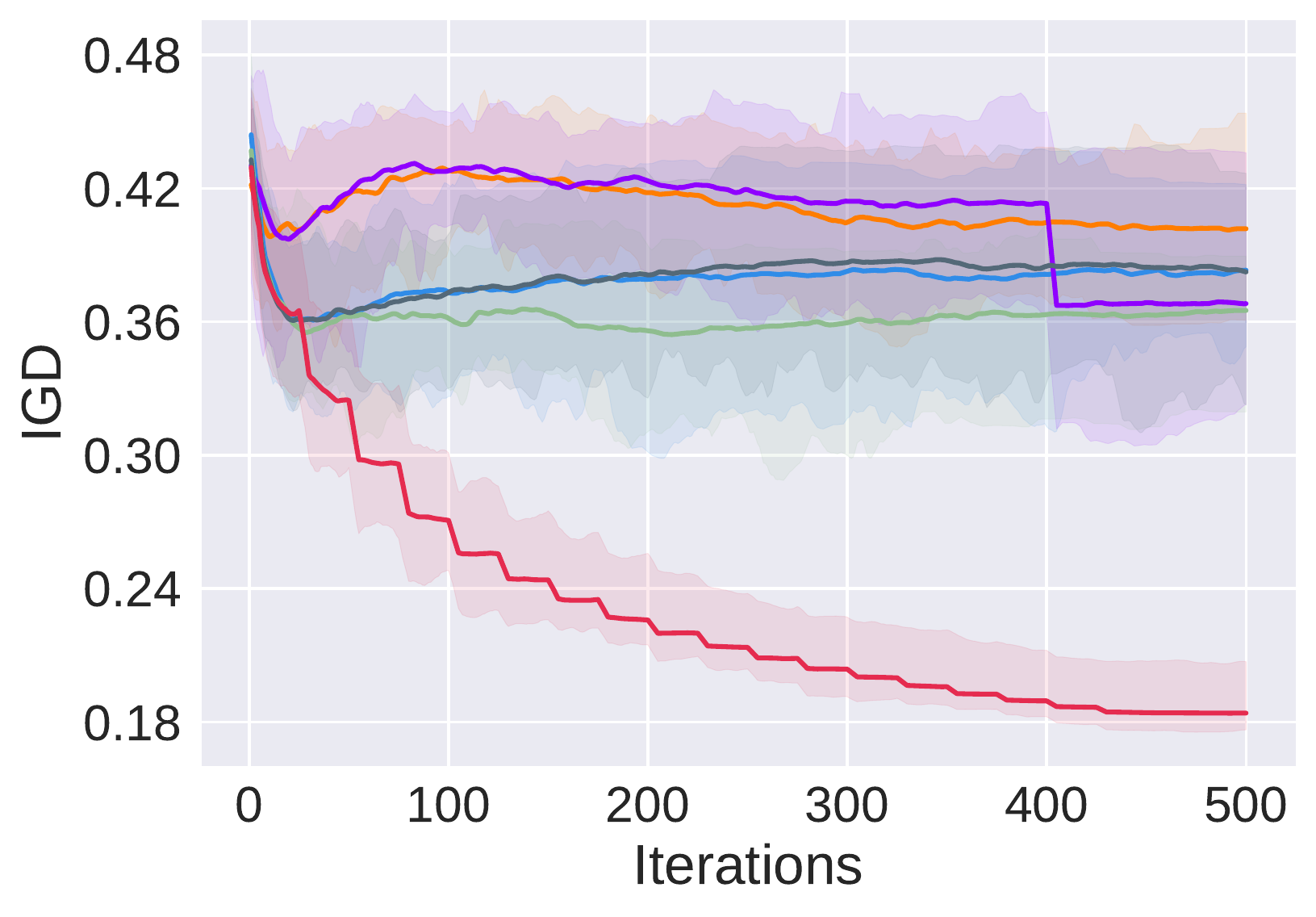}
    }
    \caption{Curves of IGD value obtained by the compared methods on the 5-objective problems.}
    \label{fig:5objs}
\end{figure*}

\begin{figure*}[htbp!]
    \centering
    \vspace{-1em}
    \centering
    \subfigure[DTLZ2\_7]{
	    \includegraphics[width=0.23\textwidth]{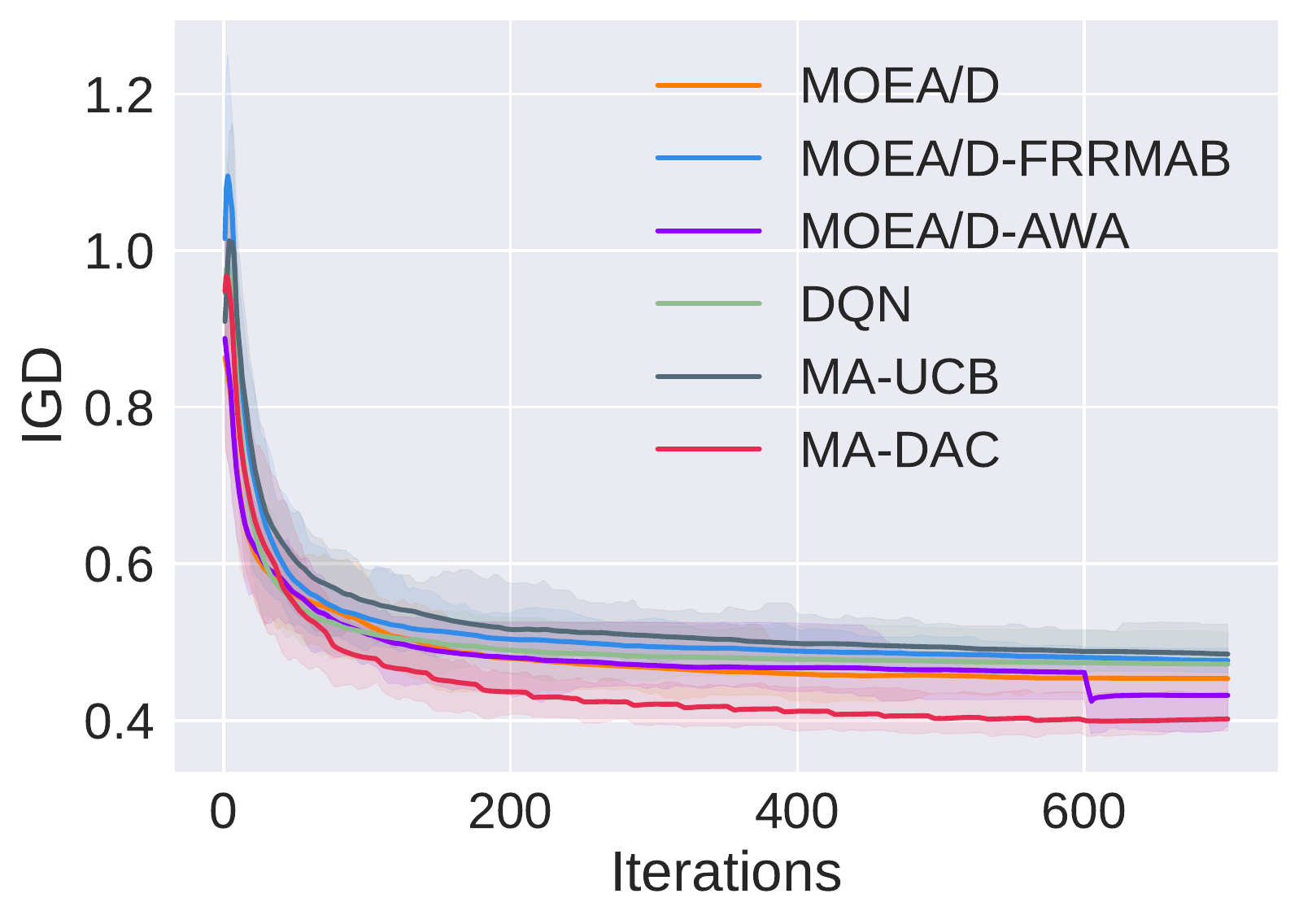}
    }
    \subfigure[DTLZ4\_7]{
        \includegraphics[width=0.23\textwidth]{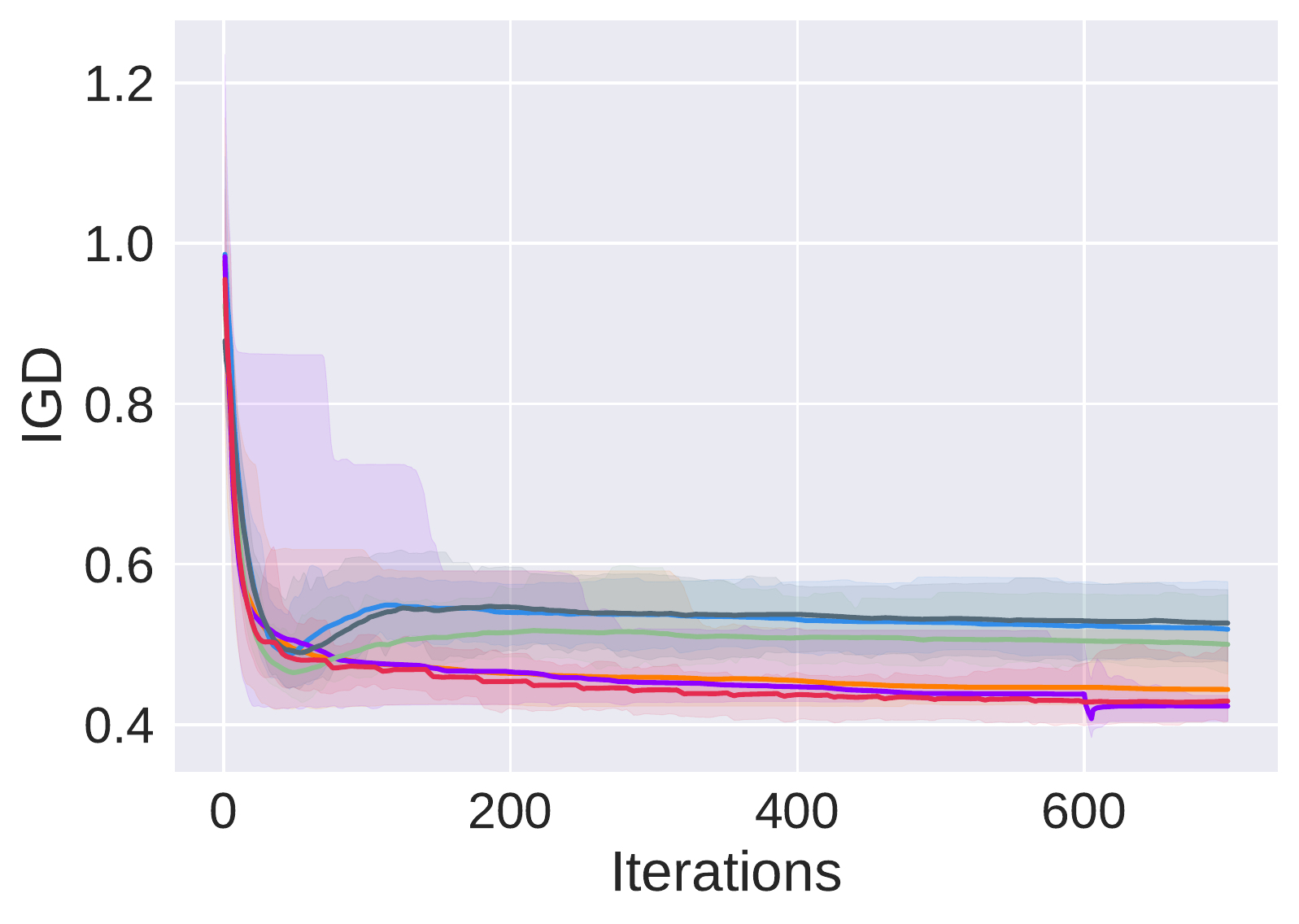}
    }
    \subfigure[WFG4\_7]{
        \includegraphics[width=0.23\textwidth]{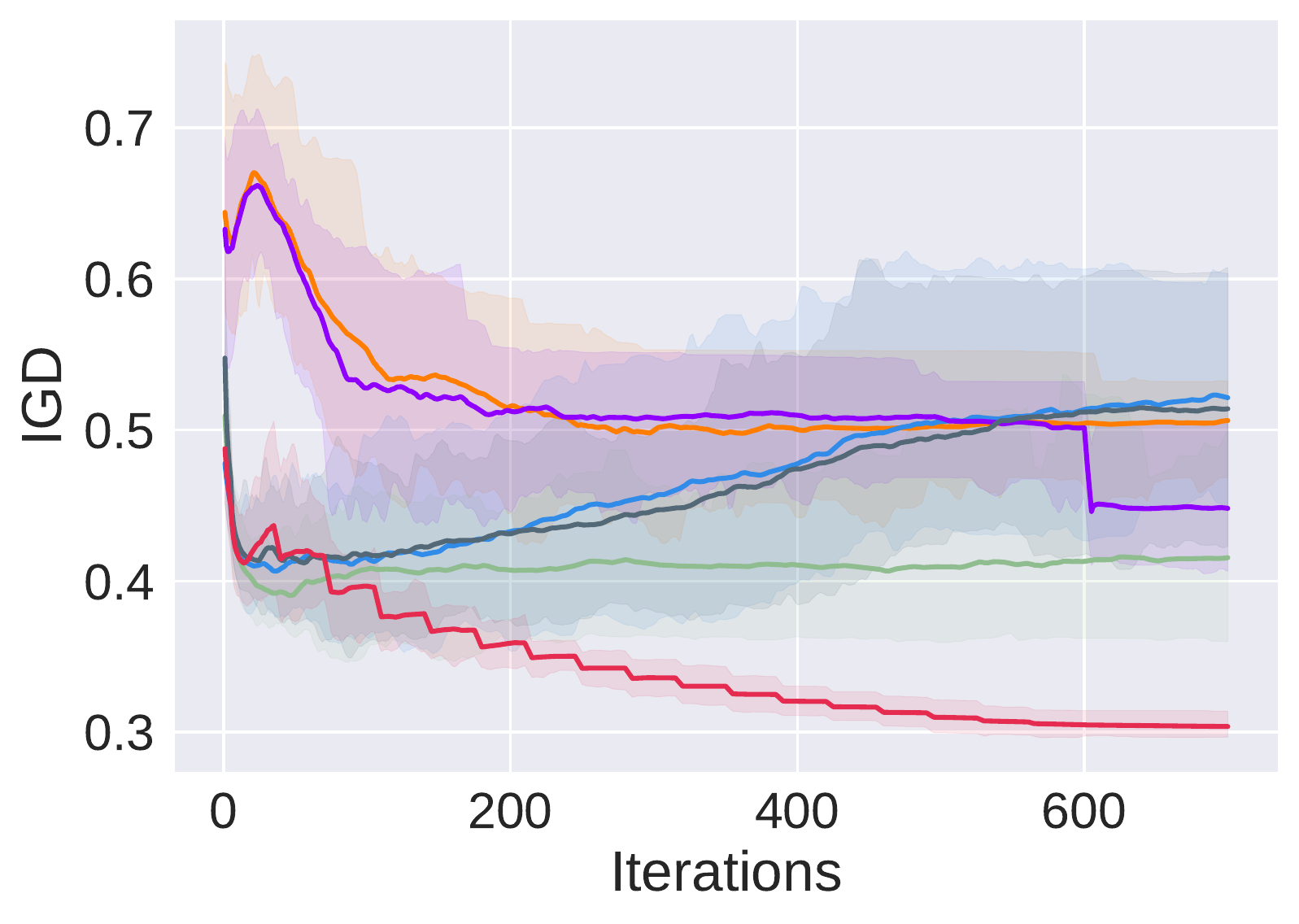}
    }
    \subfigure[WFG5\_7]{
        \includegraphics[width=0.23\textwidth]{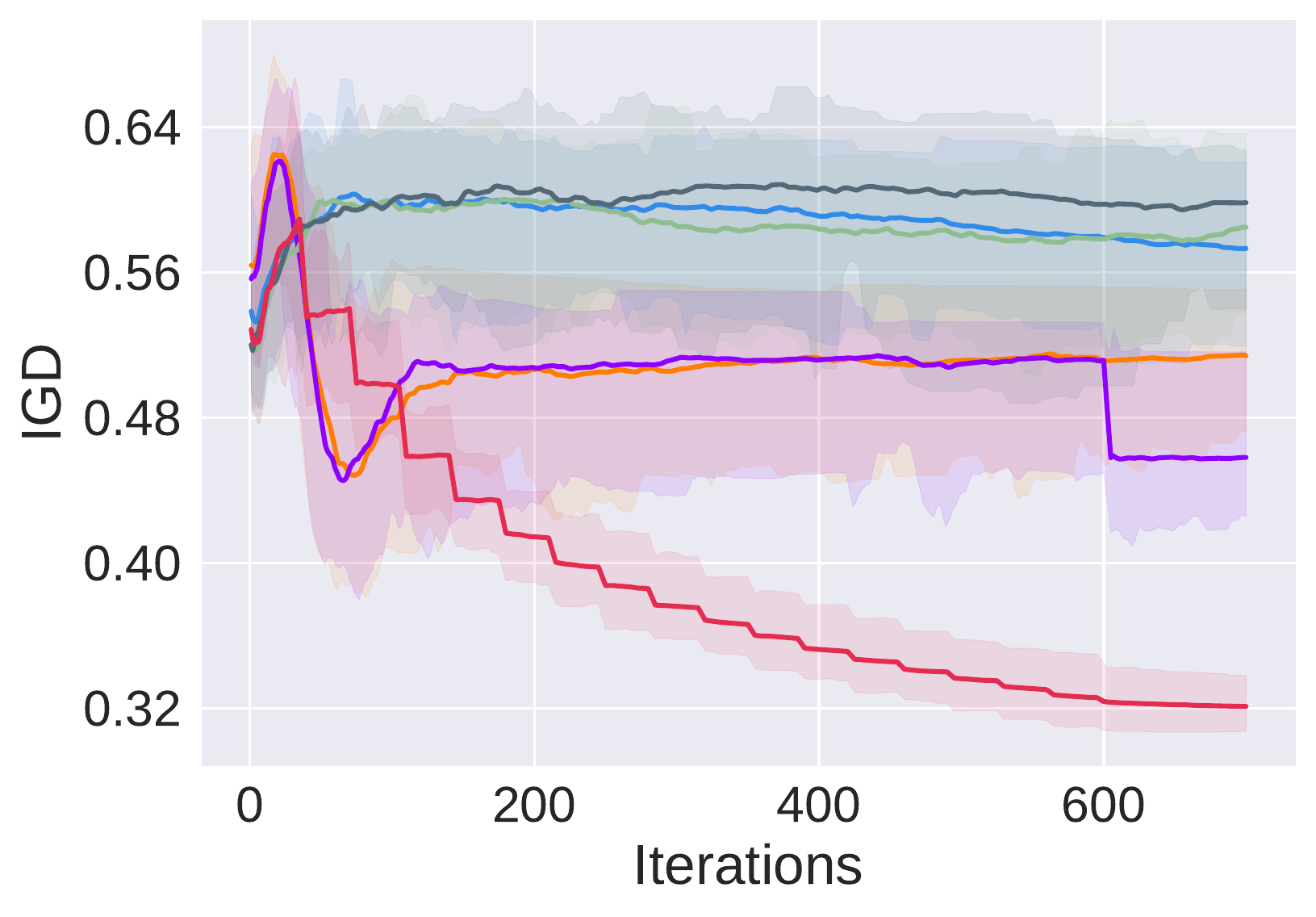}
    }
    \vspace{-1em}\\
    \subfigure[WFG6\_7]{
        \includegraphics[width=0.23\textwidth]{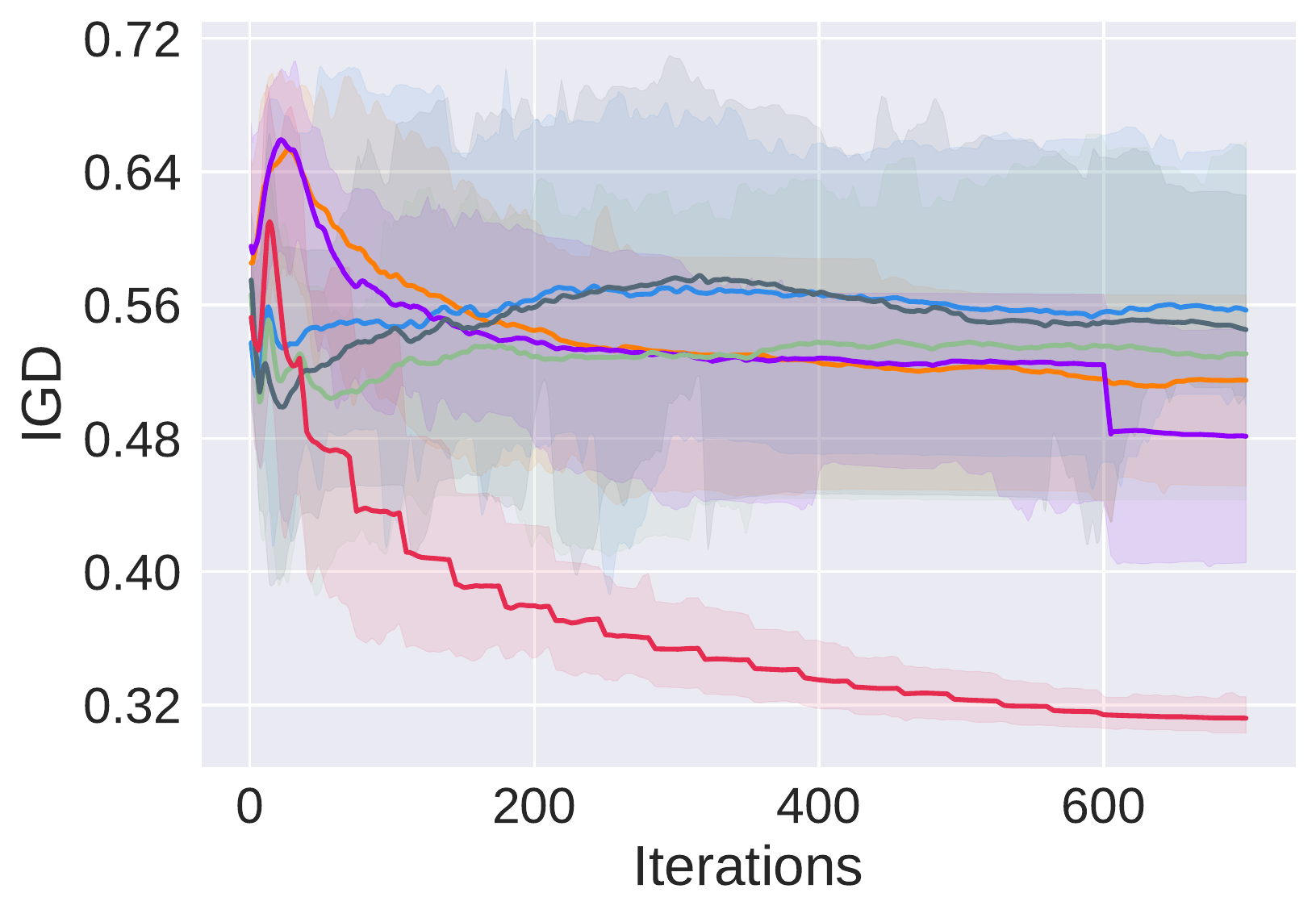}
    }
    \subfigure[WFG7\_7]{
        \includegraphics[width=0.23\textwidth]{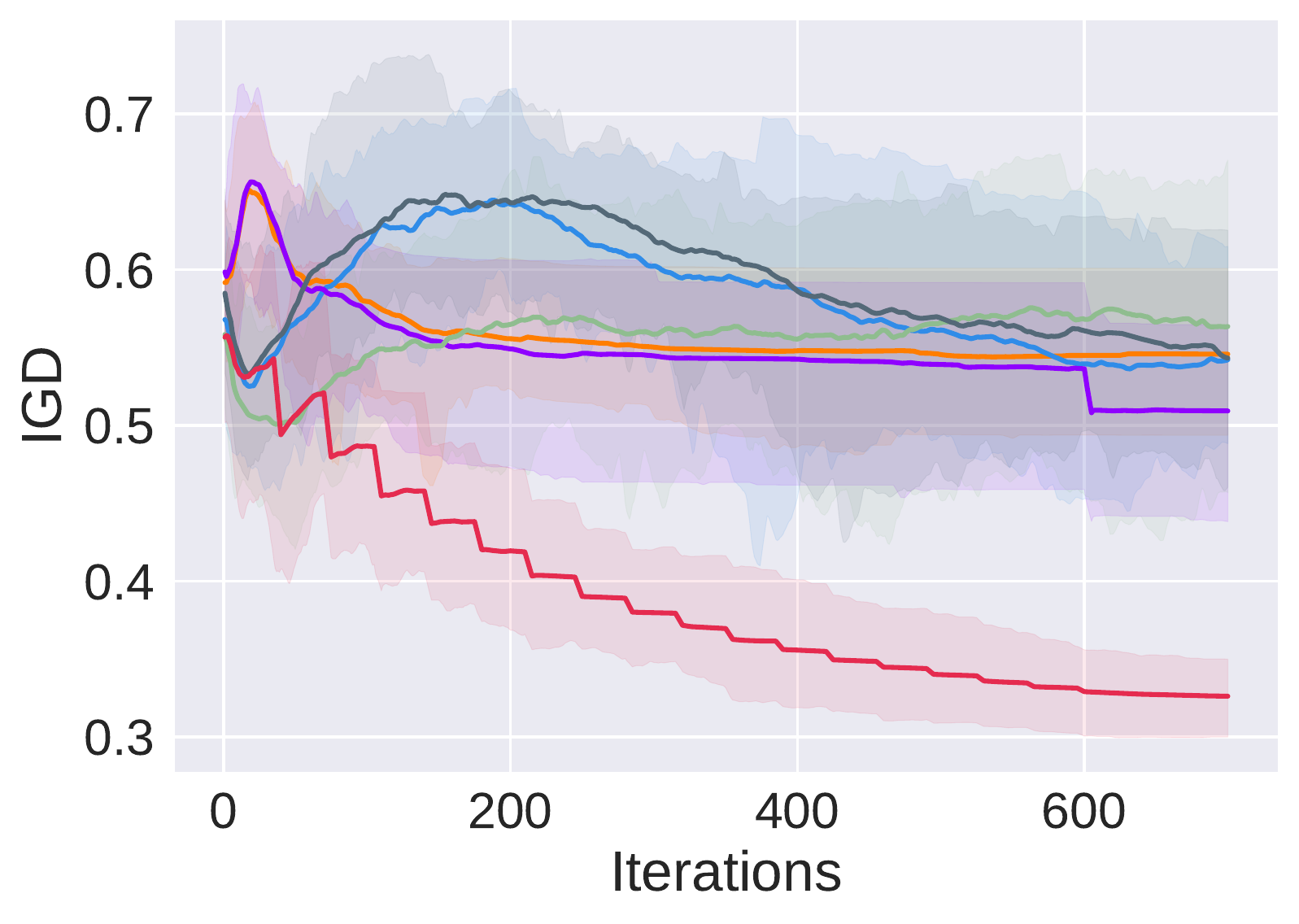}
    }
    \subfigure[WFG8\_7]{
        \includegraphics[width=0.23\textwidth]{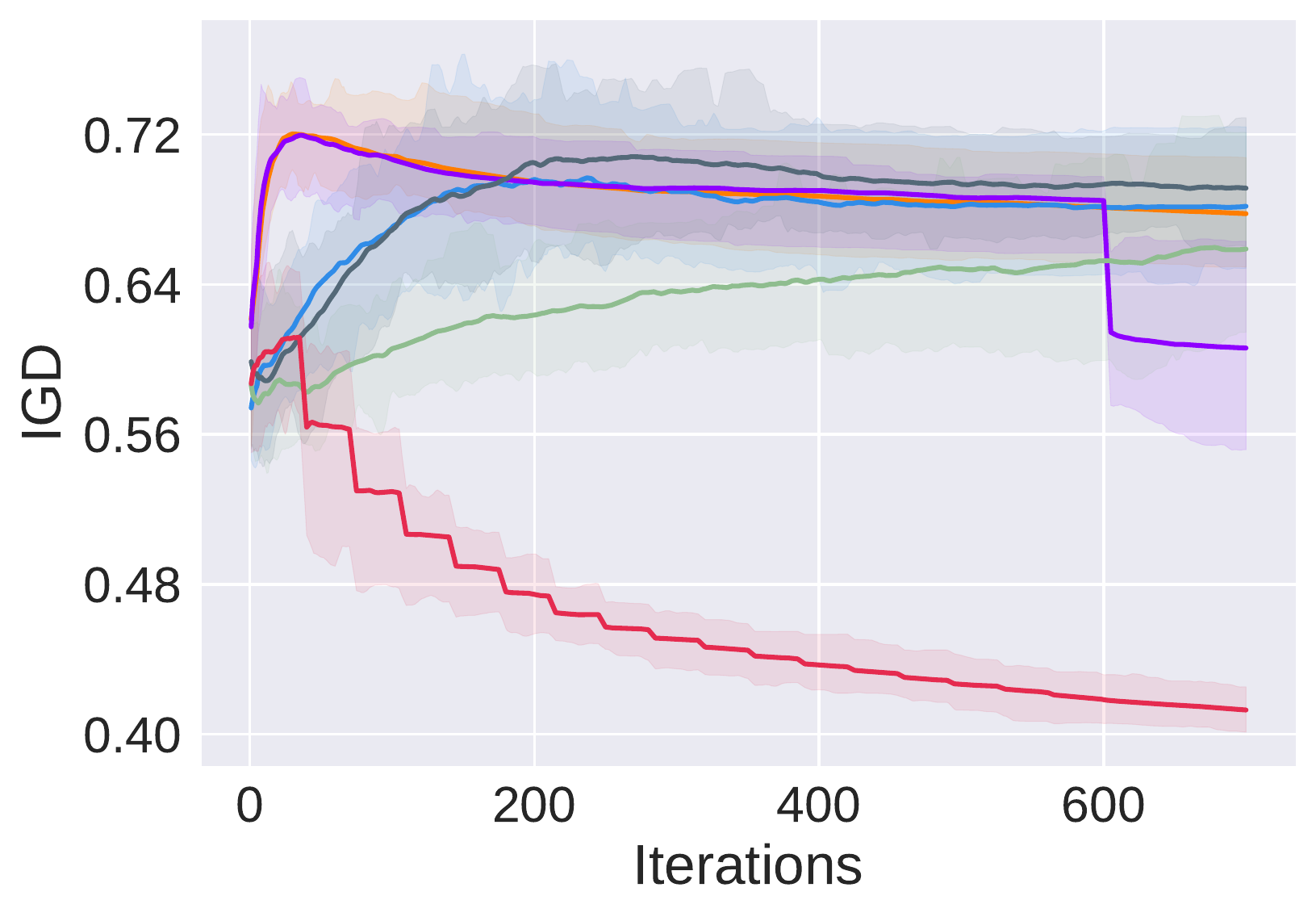}
    }
    \subfigure[WFG9\_7]{
        \includegraphics[width=0.23\textwidth]{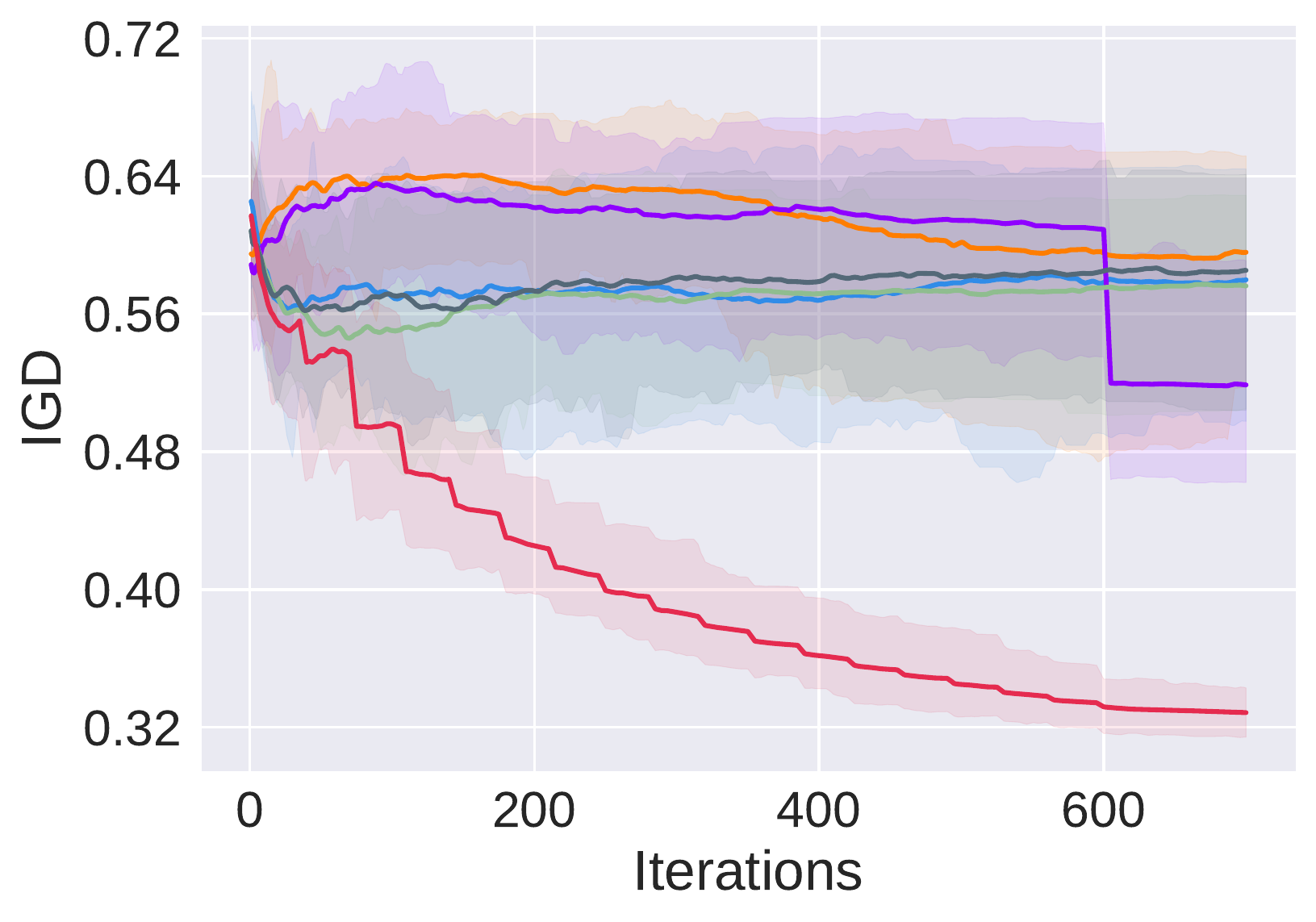}
    }
    \caption{Curves of IGD value obtained by the compared methods on the 7-objective problems.}
    \label{fig:7objs}
\end{figure*}

We plot the curves of IGD value of all the compared methods (i.e., MOEA/D, MOEA/D-FRRMAB, MOEA/D-AWA, DQN, MA-UCB and MA-DAC) on the problems with 3, 5 and 7 objectives of 30 runs, as shown in Figures~\ref{fig:3objs},~\ref{fig:5objs} and~\ref{fig:7objs}, respectively. We can observe that MA-DAC performs the best in general, and the superiority is more clear on the problems with 5 and 7 objectives. As the number of objectives increases, the problems become more difficult, thus requiring a powerful policy of adjusting the configuration hyperparameters. This also implies the applicability of MA-DAC in solving difficult problems.

\subsection{Comparison with different DQN variants}
\label{app: Comparison with different DQN variants}
DQN has been shown the limited ability to handle large-scale action spaces~\citep{dac, theoryInspired}. To improve the compared baseline, we use DQN to dynamically adjust each type of hyperparameter of MOEA/D, resulting in four DQN variants, denoted as DQN-1, DQN-2, DQN-3 and DQN-4. They represent using DQN to adjust weights, neighborhood size, types of reproduction operators, and parameters of reproduction operators, respectively. Other fixed hyperparameters' default settings are the same as those in Appendix~\ref{app: details exp settings}. We compare them with DQN (which dynamically adjusts all the hyperparameters of MOEA/D) on all the functions with 3, 5 and 7 objectives. Training functions remain the same. That is, DTLZ2, WFG4 and WFG6.

\begin{table}[htbp]\scriptsize
\centering
\caption{IGD values obtained by DQN-1, DQN-2, DQN-3 and DQN-4 on different problems. Each result consists of the mean and standard deviation of 30 runs. The best mean value of each problem is highlighted in \textbf{bold}. The symbols `$+$', `$-$' and `$\approx$' indicate that the result is significantly superior to, inferior to, and almost equivalent to DQN, respectively, according to the Wilcoxon rank-sum test with significance level 0.05.}\vspace{0.5em}
\resizebox{\linewidth}{!}{
\begin{tabular}{c c|c c c c c }
\toprule
     Problem     &   $M$  & DQN-1  & DQN-2 & DQN-3 & DQN-4 \\\midrule
\multirow{3}{*}{DTLZ2} & 3 & \textbf{4.100E-02 }(7.64E-04) $+$  & 4.283E-02 (5.50E-04) $+$  & 4.284E-02 (3.47E-04) $+$  & 4.244E-02 (5.46E-04) $+$ \\
 & 5 & \textbf{2.408E-01 }(1.03E-02) $+$  & 2.469E-01 (3.31E-03) $+$  & 2.531E-01 (3.43E-03) $+$  & 2.530E-01 (4.35E-03) $+$ \\
 & 7 & \textbf{4.064E-01 }(1.14E-02) $+$  & 4.164E-01 (8.81E-03) $+$  & 4.149E-01 (8.92E-03) $+$  & 4.132E-01 (8.89E-03) $+$ \\\midrule

\multirow{3}{*}{WFG4} & 3 & 6.136E-02 (1.80E-03) $+$  & 6.383E-02 (1.92E-03) $+$  & \textbf{5.885E-02 }(1.50E-03) $+$  & 6.031E-02 (1.24E-03) $+$ \\
 & 5 & \textbf{1.887E-01 }(2.15E-03) $+$  & 2.264E-01 (4.37E-03) $+$  & 2.351E-01 (5.49E-03) $+$  & 2.274E-01 (3.69E-03) $+$ \\
 & 7 & \textbf{3.018E-01 }(3.14E-03) $+$  & 3.421E-01 (9.24E-03) $+$  & 3.427E-01 (9.67E-03) $+$  & 3.419E-01 (9.62E-03) $+$ \\\midrule

\multirow{3}{*}{WFG6} & 3 & 5.149E-02 (1.17E-02) $+$  & \textbf{5.104E-02 }(8.75E-03) $+$  & 6.538E-02 (2.37E-02) $+$  & 6.120E-02 (2.10E-02) $+$ \\
 & 5 & \textbf{1.991E-01 }(9.10E-03) $+$  & 2.642E-01 (2.21E-02) $+$  & 2.665E-01 (1.73E-02) $+$  & 2.712E-01 (2.29E-02) $+$ \\
 & 7 & \textbf{3.204E-01 }(7.11E-03) $+$  & 3.992E-01 (2.64E-02) $+$  & 3.967E-01 (2.29E-02) $+$  & 3.824E-01 (1.88E-02) $+$ \\\midrule
 
 Train: $+$/$-$/$\approx$ &  & 9/0/0  & 9/0/0  & 9/0/0  & 9/0/0 \\\midrule
 
 \multirow{3}{*}{DTLZ4} & 3 & \textbf{4.697E-02 }(3.78E-03) $+$  & 5.249E-02 (1.14E-02) $+$  & 6.116E-02 (5.10E-02) $+$  & 5.469E-02 (3.65E-02) $+$ \\
 & 5 & \textbf{3.074E-01 }(1.24E-02) $+$  & 3.136E-01 (1.19E-02) $+$  & 3.207E-01 (1.26E-02) $+$  & 3.243E-01 (1.75E-02) $+$ \\
 & 7 & 4.263E-01 (1.28E-02) $+$  & \textbf{4.179E-01 }(1.44E-02) $+$  & 4.330E-01 (2.83E-02) $+$  & 4.211E-01 (1.21E-02) $+$ \\\midrule
 
 \multirow{3}{*}{WFG5} & 3 & \textbf{4.815E-02 }(6.96E-04) $+$  & 5.173E-02 (7.46E-04) $+$  & 5.188E-02 (5.99E-04) $+$  & 5.175E-02 (7.38E-04) $+$ \\
 & 5 & \textbf{1.833E-01 }(3.37E-03) $+$  & 2.532E-01 (5.66E-03) $+$  & 2.537E-01 (5.17E-03) $+$  & 2.560E-01 (6.80E-03) $+$ \\
 & 7 & \textbf{3.294E-01 }(8.51E-03) $+$  & 4.053E-01 (1.80E-02) $+$  & 4.090E-01 (1.11E-02) $+$  & 4.088E-01 (1.34E-02) $+$ \\\midrule
 
\multirow{3}{*}{WFG7} & 3 & \textbf{4.624E-02 }(6.74E-04) $+$  & 4.841E-02 (1.20E-03) $+$  & 4.638E-02 (8.69E-04) $+$  & 4.690E-02 (8.85E-04) $+$ \\
 & 5 & \textbf{1.835E-01 }(3.02E-03) $+$  & 2.476E-01 (7.11E-03) $+$  & 2.492E-01 (5.16E-03) $+$  & 2.445E-01 (9.30E-03) $+$ \\
 & 7 & \textbf{3.274E-01 }(1.15E-02) $+$  & 3.956E-01 (1.96E-02) $+$  & 3.951E-01 (1.88E-02) $+$  & 3.843E-01 (1.81E-02) $+$ \\\midrule
\multirow{3}{*}{WFG8} & 3 & 8.658E-02 (1.30E-03) $+$  & 8.850E-02 (1.40E-03) $+$  & \textbf{8.490E-02 }(2.33E-03) $+$  & 8.555E-02 (1.59E-03) $+$ \\
 & 5 & \textbf{2.519E-01 }(8.57E-03) $+$  & 3.216E-01 (1.01E-02) $+$  & 3.485E-01 (1.65E-02) $+$  & 3.328E-01 (1.29E-02) $+$ \\
 & 7 & \textbf{4.163E-01 }(7.50E-03) $+$  & 5.153E-01 (1.61E-02) $+$  & 5.172E-01 (1.73E-02) $+$  & 4.970E-01 (1.12E-02) $+$ \\\midrule
\multirow{3}{*}{WFG9} & 3 & 7.747E-02 (2.46E-02) $-$  & 7.320E-02 (2.62E-02) $\approx$  & \textbf{4.969E-02 }(1.34E-02) $+$  & 6.138E-02 (2.42E-02) $+$ \\
 & 5 & \textbf{2.041E-01 }(5.21E-03) $+$  & 2.542E-01 (1.03E-02) $+$  & 2.493E-01 (9.82E-03) $+$  & 2.486E-01 (1.18E-02) $+$ \\
 & 7 & \textbf{3.418E-01 }(1.18E-02) $+$  & 4.088E-01 (1.71E-02) $+$  & 4.179E-01 (1.89E-02) $+$  & 4.069E-01 (1.73E-02) $+$ \\\midrule

Test: $+$/$-$/$\approx$ &  & 14/1/0  & 14/0/1  & 15/0/0  & 15/0/0 \\
\bottomrule
\end{tabular}}
\label{table:8 dqn_ablation}
\end{table}

The results are shown in Table~\ref{table:8 dqn_ablation}. As expected, all the four DQN variants outperform the DQN version that adjusts all four hyperparameters, demonstrating the ineffectiveness of using a single agent to solve complex DAC problems with large action spaces. Furthermore, it can also be observed that DQN-1 performs the best among the four DQN variants, which is consistent with the observation in Table~\ref{table:3 ablation}, i.e., adjusting weights is generally more important.

\subsection{Comparison with different MARL algorithms}
\label{app: different marl algorithms}
To further demonstrate that the effectiveness of MA-DAC is agnostic to specific MARL algorithms, we use two more MARL algorithms as the implementations of policy networks, i.e., Independent Q Learning (IQL)~\citep{iql} and QMIX~\citep{qmix}.
IQL models each hyperparameter as an individual agent and trains each agent independently, where each agent learns an individual value function that ignores the possible effect of other agents. QMIX and VDN are value decomposition algorithms that decompose the joint value function into individual value functions. Compared to the linear value decomposition of VDN, QMIX employs a non-linear mixing network to decompose the total reward, enabling the learning of a rich joint value function.

As shown in Table~\ref{table:9 marl algorithms}, all the three MA-DAC policies have clear advantage over the competitive baseline DQN-1, demonstrating the effectiveness of MA-DAC. We can also see that no MARL algorithm can take a lead on all the instances, although QMIX is theoretically a better algorithm~\citep{qmix}. One possible reason is that the tasks' strong randomness and heterogeneity introduce new challenges. We will further investigate the MARL algorithm in this regard, and meanwhile we hope that the MaMo benchmark can attract the attention of the MARL community to this type of problems.

\begin{table}[htbp]\scriptsize
\centering
\caption{IGD values obtained by DQN-1, VDN, IQL and QMIX on different problems. Each result consists of the mean and standard deviation of 30 runs. The best mean value of each problem is highlighted in \textbf{bold}. The symbols `$+$', `$-$' and `$\approx$' indicate that the result is significantly superior to, inferior to, and almost equivalent to DQN-1 in Table~\ref{table:8 dqn_ablation}, respectively, according to the Wilcoxon rank-sum test with significance level 0.05.}\vspace{0.5em}
\resizebox{\linewidth}{!}{
\begin{tabular}{c c|c c c c c}
\toprule
     Problem     &   $M$  & DQN-1 & VDN & IQL & QMIX \\\midrule

\multirow{3}{*}{DTLZ2} & 3 & 4.100E-02 (7.64E-04)   & \textbf{3.807E-02 }(5.05E-04) $+$  & 3.933E-02 (4.29E-04) $+$  & 3.916E-02 (7.21E-04) $+$ \\
 & 5 & 2.408E-01 (1.03E-02)   & 2.442E-01 (1.26E-02) $\approx$  & \textbf{2.310E-01 }(9.29E-03) $+$  & 2.419E-01 (1.37E-02) $\approx$ \\
 & 7 & 4.064E-01 (1.14E-02)   & \textbf{3.944E-01 }(1.17E-02) $+$  & 4.138E-01 (9.79E-03) $-$  & 4.162E-01 (1.59E-02) $-$ \\\midrule
\multirow{3}{*}{WFG4} & 3 & 6.136E-02 (1.80E-03)   & \textbf{5.200E-02 }(1.19E-03) $+$  & 5.438E-02 (8.83E-04) $+$  & 5.206E-02 (1.16E-03) $+$ \\
 & 5 & 1.887E-01 (2.15E-03)   & 1.868E-01 (2.81E-03) $+$  & 1.879E-01 (2.78E-03) $\approx$  & \textbf{1.859E-01 }(2.25E-03) $+$ \\
 & 7 & 3.018E-01 (3.14E-03)   & 3.033E-01 (3.66E-03) $\approx$  & 3.046E-01 (3.55E-03) $-$  & \textbf{2.998E-01 }(4.21E-03) $+$ \\\midrule
\multirow{3}{*}{WFG6} & 3 & 5.149E-02 (1.17E-02)   & 4.831E-02 (8.95E-03) $+$  & 5.592E-02 (1.57E-02) $\approx$  & \textbf{4.542E-02 }(3.02E-03) $+$ \\
 & 5 & 1.991E-01 (9.10E-03)   & \textbf{1.942E-01 }(6.90E-03) $+$  & 1.981E-01 (6.76E-03) $\approx$  & 1.975E-01 (6.98E-03) $\approx$ \\
 & 7 & 3.204E-01 (7.11E-03)   & \textbf{3.112E-01 }(4.93E-03) $+$  & 3.148E-01 (3.34E-03) $+$  & 3.128E-01 (7.39E-03) $+$ \\\midrule
Train: $+$/$-$/$\approx$ &  &    & 7/0/2  & 4/2/3  & 6/1/2 \\\midrule

\multirow{3}{*}{DTLZ4} & 3 & \textbf{4.697E-02 }(3.78E-03)   & 6.700E-02 (6.14E-02) $\approx$  & 6.328E-02 (4.48E-02) $-$  & 5.094E-02 (2.38E-03) $-$ \\
 & 5 & 3.074E-01 (1.24E-02)   & \textbf{2.995E-01 }(2.10E-02) $+$  & 3.021E-01 (1.62E-02) $\approx$  & 3.013E-01 (1.80E-02) $+$ \\
 & 7 & 4.263E-01 (1.28E-02)   & \textbf{4.182E-01 }(1.21E-02) $+$  & 4.323E-01 (1.43E-02) $\approx$  & 4.303E-01 (1.95E-02) $\approx$ \\\midrule
\multirow{3}{*}{WFG5} & 3 & 4.815E-02 (6.96E-04)   & \textbf{4.730E-02 }(7.89E-04) $+$  & 4.818E-02 (6.24E-04) $\approx$  & 4.736E-02 (7.49E-04) $+$ \\
 & 5 & 1.833E-01 (3.37E-03)   & \textbf{1.811E-01 }(3.02E-03) $+$  & 1.812E-01 (2.21E-03) $+$  & 1.813E-01 (2.54E-03) $+$ \\
 & 7 & 3.294E-01 (8.51E-03)   & 3.206E-01 (8.04E-03) $+$  & \textbf{3.173E-01 }(6.91E-03) $+$  & 3.175E-01 (7.19E-03) $+$ \\\midrule
\multirow{3}{*}{WFG7} & 3 & 4.624E-02 (6.74E-04)   & \textbf{4.066E-02 }(5.31E-04) $+$  & 4.313E-02 (7.79E-04) $+$  & 4.077E-02 (4.94E-04) $+$ \\
 & 5 & 1.835E-01 (3.02E-03)   & 1.858E-01 (2.12E-03) $-$  & 1.832E-01 (2.34E-03) $\approx$  & \textbf{1.825E-01 }(3.16E-03) $\approx$ \\
 & 7 & 3.274E-01 (1.15E-02)   & 3.258E-01 (1.25E-02) $\approx$  & \textbf{3.219E-01 }(1.09E-02) $+$  & 3.226E-01 (1.12E-02) $+$ \\\midrule
\multirow{3}{*}{WFG8} & 3 & 8.658E-02 (1.30E-03)   & \textbf{7.901E-02 }(1.19E-03) $+$  & 8.652E-02 (2.75E-03) $\approx$  & 7.909E-02 (1.60E-03) $+$ \\
 & 5 & 2.519E-01 (8.57E-03)   & \textbf{2.479E-01 }(7.20E-03) $+$  & 2.544E-01 (8.10E-03) $\approx$  & 2.496E-01 (9.83E-03) $\approx$ \\
 & 7 & 4.163E-01 (7.50E-03)   & 4.127E-01 (5.93E-03) $+$  & 4.175E-01 (7.32E-03) $\approx$  & \textbf{4.006E-01 }(9.42E-03) $+$ \\\midrule
\multirow{3}{*}{WFG9} & 3 & 7.747E-02 (2.46E-02)   & \textbf{4.159E-02 }(6.10E-04) $+$  & 4.423E-02 (7.08E-04) $+$  & 4.167E-02 (5.92E-04) $+$ \\
 & 5 & 2.041E-01 (5.21E-03)   & \textbf{1.832E-01 }(7.10E-03) $+$  & 1.915E-01 (8.87E-03) $+$  & 1.921E-01 (6.43E-03) $+$ \\
 & 7 & 3.418E-01 (1.18E-02)   & \textbf{3.278E-01 }(7.21E-03) $+$  & 3.322E-01 (8.41E-03) $+$  & 3.298E-01 (8.46E-03) $+$ \\\midrule
Test: $+$/$-$/$\approx$ &  &    & 12/1/2  & 7/1/7  & 11/1/3 \\
\bottomrule
\end{tabular}}
\label{table:9 marl algorithms}
\end{table}

\section{Experiments on DACBench}
\label{app: exp on dacbench}

DACBench~\citep{dacbench} is a benchmark library for DAC which collects and standardizes existing DAC benchmarks from different domains. Among these benchmarks, Sigmoid~\citep{dac} is an ideal one for DAC developers to test the performance of the learned policy. Sigmoid can generate instance sets with a wide range of difficulties, e.g., by increasing the number of parameters and their configuration space dimension, or extending the heterogeneity between instances.


To formulate the instance set of Sigmoid, one needs to specify the number $N_p$ of parameters, the size $D_j$ of the configuration space for each parameter $j$, and the way to generate the distribution of the instances. For each parameter $j$ in a given instance $i$, we need to choose its configuration to approximate an independent Sigmoid function  $$\operatorname{sig}(t,s_{i,j},p_{i,j})=(1+e^{-s_{i,j}\cdot (t-p_{i,j})})^{-1},$$ 
where $s_{i,j}$ is a scaling factor, and $p_{i,j}$ is an infection point; they are the instance information for each parameter $j$ sampled from predefined distributions. To simulate interaction effects of the individual parameters, the total reward is computed as 
$$
r_t=\prod_{j=1}^{N_p} \left(1-\operatorname{abs}\left(\operatorname{sig}\left(t, s_{i, j}, p_{i, j}\right)-a_{j, t}\right)\right),
$$
where $a_{j,t}$ is the configuration value for parameter $j$ at step $t$. 


Firstly, we conduct experiments on the $5D$-Sigmoid (i.e., there are $N_p=5$ agents in MA-DAC) with action space size $D_j=3$ (i.e., each agent has a $3$-dimensional discrete action space).
The instance information for each parameter is re-sampled from a predefined normal distribution at the beginning of an episode. This is a very difficult problem, and none of the learned policies in~\citep{dac} can achieve satisfactory performance due to the large action space.

We choose the DQN in~\citep{dac}, and the best static policy achieved by SMAC~\citep{SMAC} as our baselines. All the settings of experiments and baselines are consistent with~\citep{dac}\footnote{\url{https://github.com/automl/DAC}}.
For MA-DAC, we use three popular MARL algorithms, i.e., VDN, IQL and QMIX, as the internal implementation (see Appendix~\ref{app: different marl algorithms} for the details of these algorithms). We implement VDN, IQL and QMIX policy networks based on the \texttt{EPyMARL}\footnote{\url{https://github.com/uoe-agents/epymarl}}~\citep{epymarl} framework and use the default hyperparameters, except for changing the hidden layer size from $64$ to $32$, because of the smaller state space. The state and reward of Sigmoid for each agent in MA-DAC are same as the DQN baseline, where the only difference is that MA-DAC models each parameter as an individual agent.

\begin{figure}

    \centering
    \subfigure[$5D$-Sigmoid]{
        \includegraphics[width=0.45\textwidth]{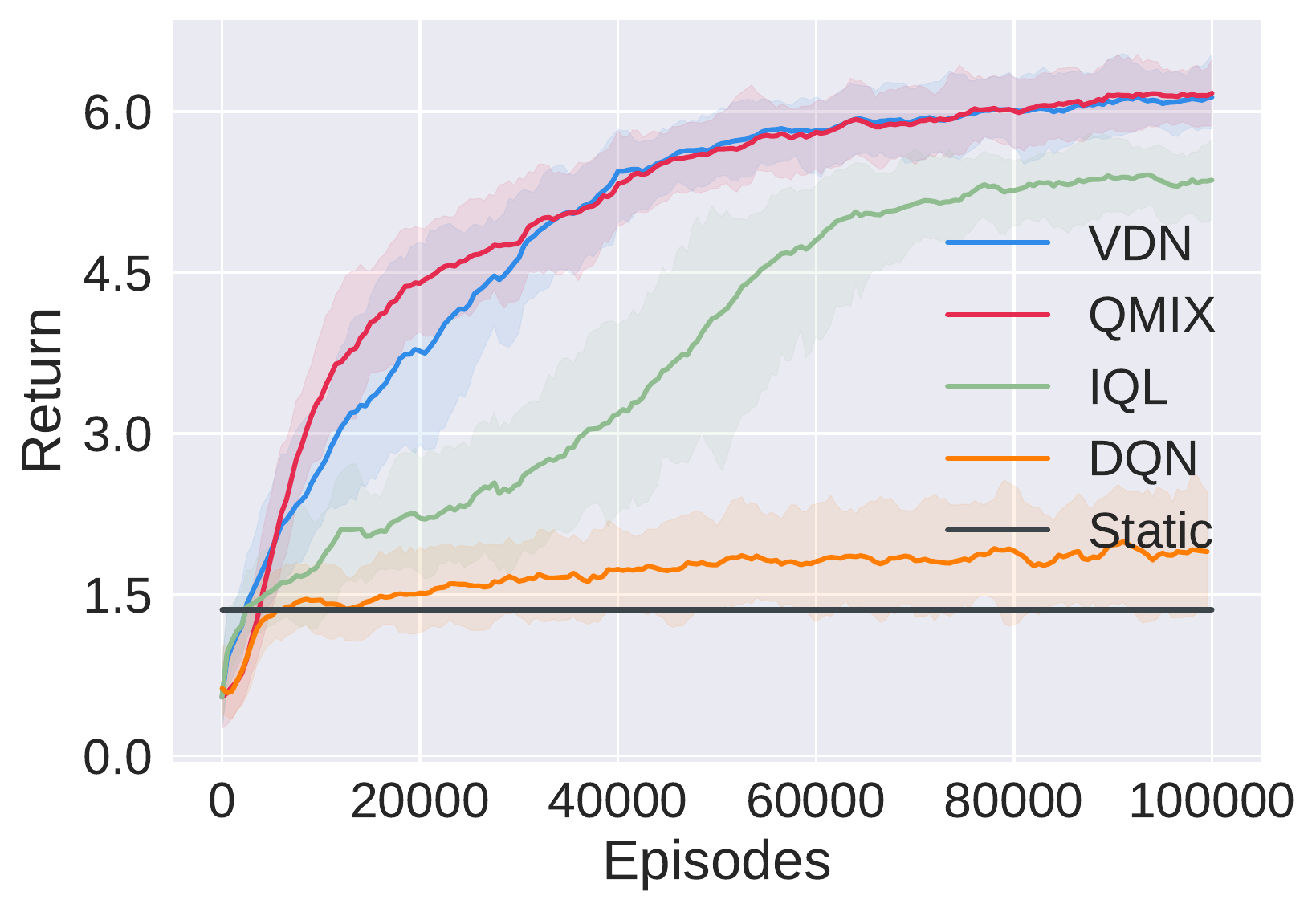}
        \label{fig:5DM3}
    }
    \subfigure[$10D$-Sigmoid]{
        \includegraphics[width=0.45\textwidth]{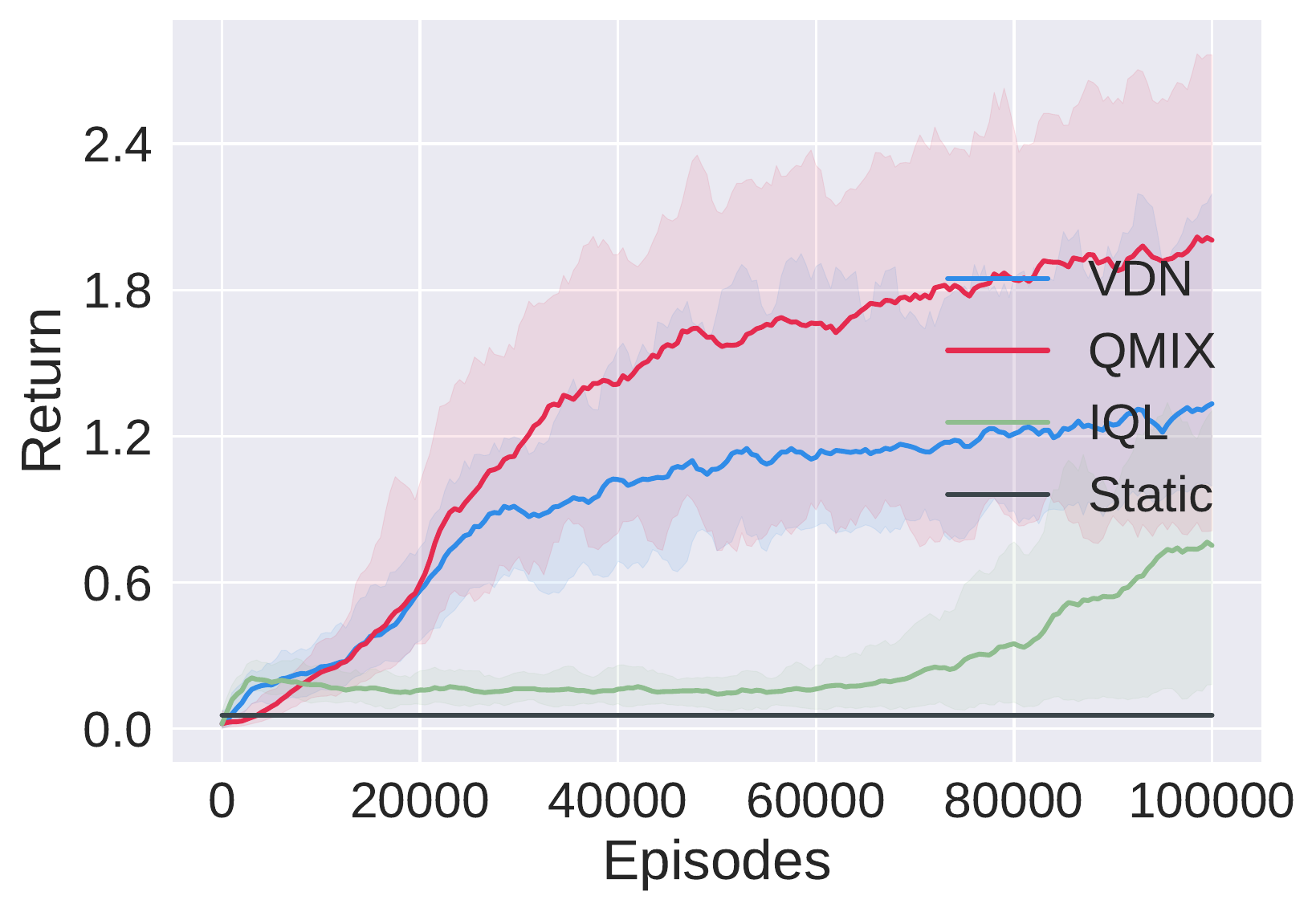}
        \label{fig:10DM3}
    }
    \caption{Curves of return value obtained by the compared methods on two Sigmoid tasks of 10 runs.}
    \label{fig:sigmoid}
\end{figure}

As shown in Figure~\ref{fig:5DM3}, all MA-DAC methods perform significantly better than DQN and static policies. The action space of the DQN agent increases exponentially with the number of parameters, making it hard to find a good policy and thus fail on this task~\citep{reviewOfCoop,selectiveOverview}. In contrast, MA-DAC can solve this task well by decomposing the action space so that each agent searches in a much smaller action space. 
Meanwhile, the independent learning method (i.e., IQL) is inferior to centralized training methods (i.e., VDN and QMIX). This is because the environment becomes non-stationary due to the concurrent learning of multiple learners~\citep{iql-inferior}.

Next, to test the scalability of MA-DAC, we create another task named $10D$-Sigmoid, whose only difference from $5D$-Sigmoid is that it has $10$ parameters to be configured. The DQN implementation of the DAC method cannot handle this task's ten thousand-dimensional action space due to its exponential growth in action space with the number of configured parameters. So we only take the best static policy discussed above as the baseline. As shown in Figure~\ref{fig:10DM3}, all MA-DAC versions can still find higher return than static policy, demonstrating the scalability of MA-DAC. QMIX outperforms VDN due to its more accurate value decomposition from the non-linear mixing network.
\end{document}